\documentclass[journal]{IEEEtran}

\IEEEoverridecommandlockouts 

\usepackage[pagebackref=true,colorlinks,citecolor=brown]{hyperref}
\usepackage{booktabs}
\usepackage{amsmath}
\usepackage{macros}
\usepackage{graphicx}
\usepackage{subcaption}
\usepackage{makecell}
\usepackage{enumitem}
\usepackage[font=small]{caption}
\usepackage{float}
\usepackage{dblfloatfix}
\usepackage{hyperref}
\usepackage{adjustbox}
\usepackage{amsfonts}

\usepackage{tikz}
\usetikzlibrary{positioning, arrows.meta, shapes.geometric}
\usepackage{wrapfig}
\usepackage{xspace}
\usepackage{todonotes}
\definecolor{Diffuser}{HTML}{F8DA7A} 
\definecolor{Hybrid}{HTML}{A02B93} 
\definecolor{Diffuser_joint}{HTML}{F4F0CD}
\definecolor{Diffuser_separate}{HTML}{80BEDA}

\renewcommand{\paragraph}[1]{\vspace{.1em}\noindent\textbf{#1}.}

\usepackage{algorithm}
\usepackage{algorithmic}

\makeatletter
\newcommand\fs@betterruled{%
  \def\@fs@cfont{\bfseries}\let\@fs@capt\floatc@ruled
  \def\@fs@pre{\vspace*{7pt}\hrule height.8pt depth0pt \kern2pt}%
  \def\@fs@post{\kern2pt\hrule\relax}%
  \def\@fs@mid{\kern2pt\hrule\kern2pt}%
  \let\@fs@iftopcapt\iftrue}
\floatstyle{betterruled}
\restylefloat{algorithm}
\makeatother

\usepackage{pgfplots}
\usepackage{xcolor}
\pgfplotsset{compat=newest}

\begin{document}

\title{\LARGE \bf Hybrid Diffusion for Simultaneous Symbolic and Continuous Planning}

\author{Sigmund H. H{\o}eg$^{1}$, Aksel Vaaler$^{1}$, Chaoqi Liu$^{2}$, Olav Egeland$^{1}$, and Yilun Du$^{3}$
\thanks{$^{1}$Department of Mechanical and Industrial Engineering,
        Norwegian University of Science and Technology (NTNU)
        {\tt\small sigmund.hoeg@ntnu.no}}%
\thanks{$^{2}$ University of Illinois Urbana-Champaign}%
\thanks{$^{3}$ Harvard University}%
\thanks{© 2026 IEEE. Personal use of this material is permitted.  Permission from IEEE must be obtained for all other uses, in any current or future media, including reprinting/republishing this material for advertising or promotional purposes, creating new collective works, for resale or redistribution to servers or lists, or reuse of any copyrighted component of this work in other works. IEEE Explore: https://ieeexplore.ieee.org/document/11395608}
\thanks{Please cite as: S. H. Høeg, A. Vaaler, C. Liu, O. Egeland, and Y. Du, "Hybrid Diffusion for Simultaneous Symbolic and Continuous Planning," in IEEE Robotics and Automation Letters, vol. 11, no. 4, pp. 4489-4496, April 2026, doi: 10.1109/LRA.2026.3664616.}
}

\maketitle

\newcommand{\imwidth}{0.32\textwidth}
\begin{figure*}[t]
    \vspace{10pt}
    \centering
    \begin{subfigure}[t]{0.45\textwidth}
        \centering
        \vspace{-85pt}
        \includegraphics[width=0.8\linewidth]{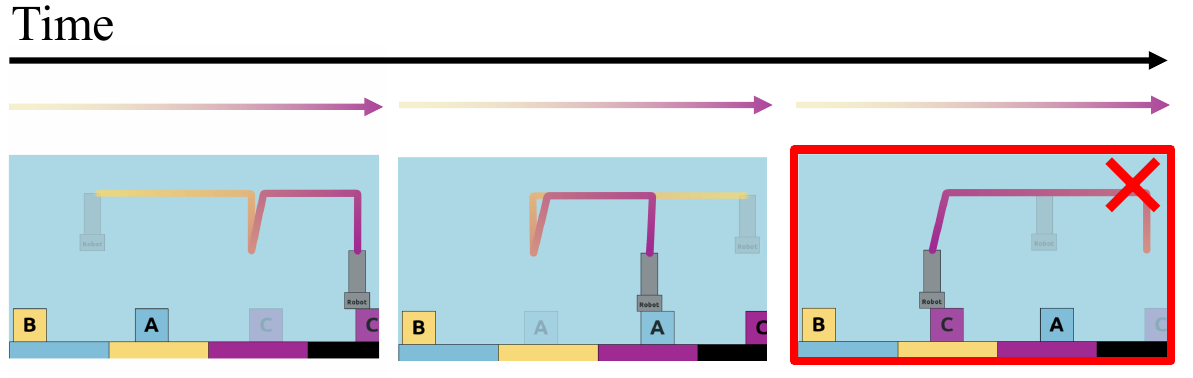}
        
        \vspace{37.5pt}
        
        \begin{tabular}{c@{\hspace{2pt}}c@{\hspace{2pt}}c@{\hspace{2pt}}c}
            \adjustbox{trim={0.2\width} {0} {0.2\width} {0}, clip}{\includegraphics[width=\imwidth]{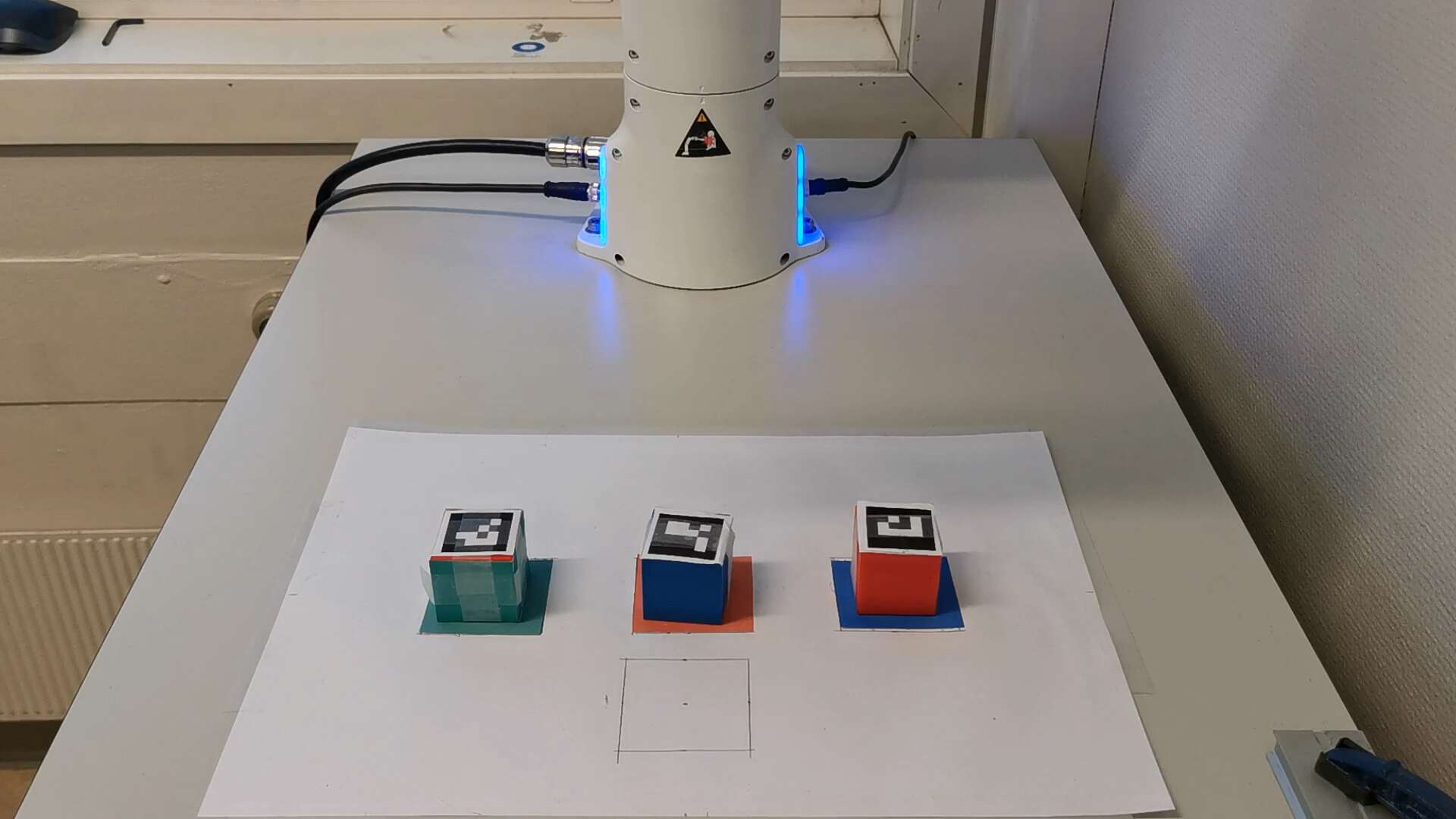}} &
            \adjustbox{trim={0.2\width} {0} {0.2\width} {0}, clip}{\includegraphics[width=\imwidth]{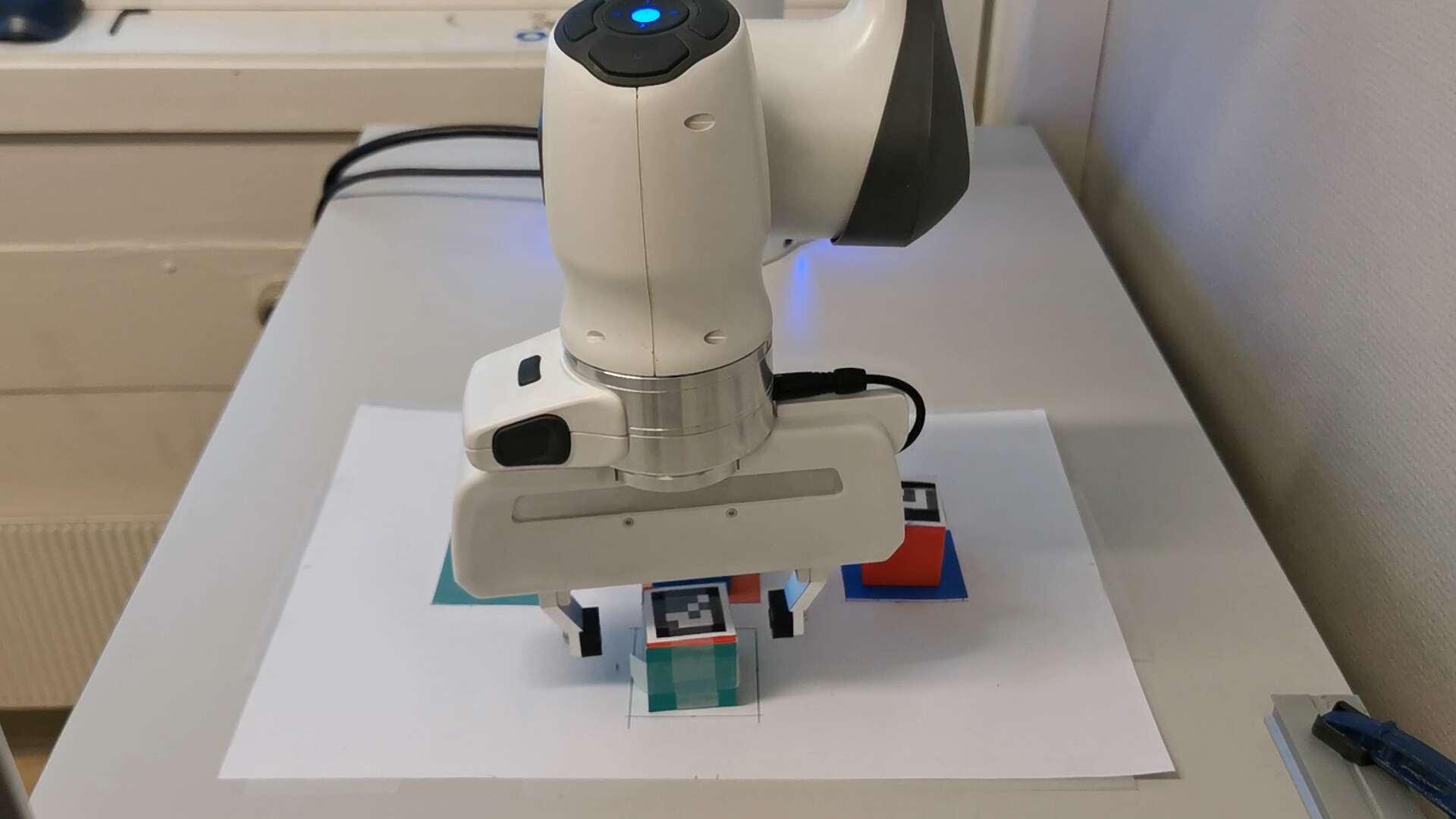}} &
            \adjustbox{trim={0.2\width} {0} {0.2\width} {0}, clip}{\includegraphics[width=\imwidth]{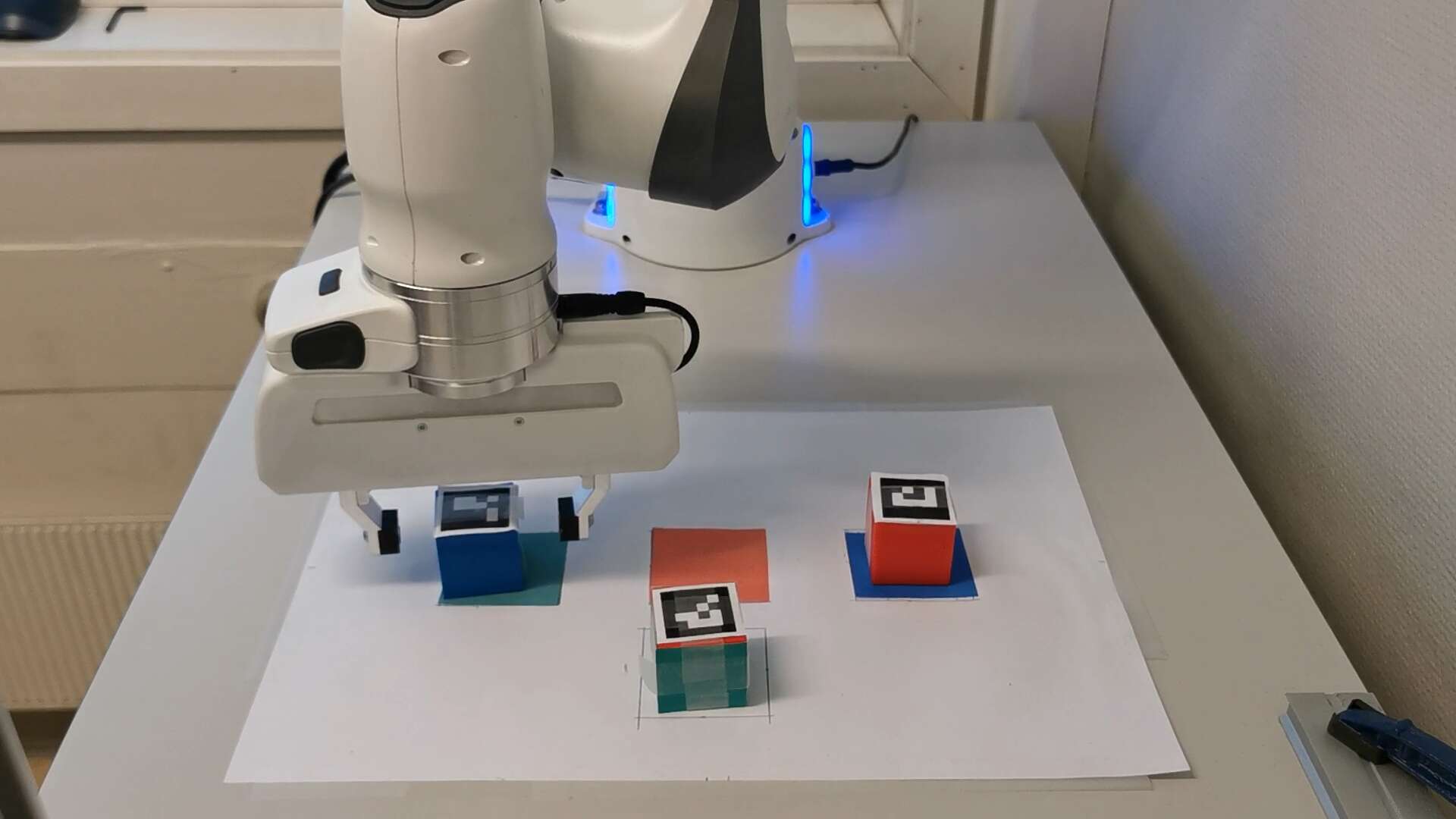}} &
            \adjustbox{trim={0.2\width} {0} {0.2\width} {0}, clip}{\includegraphics[width=\imwidth]{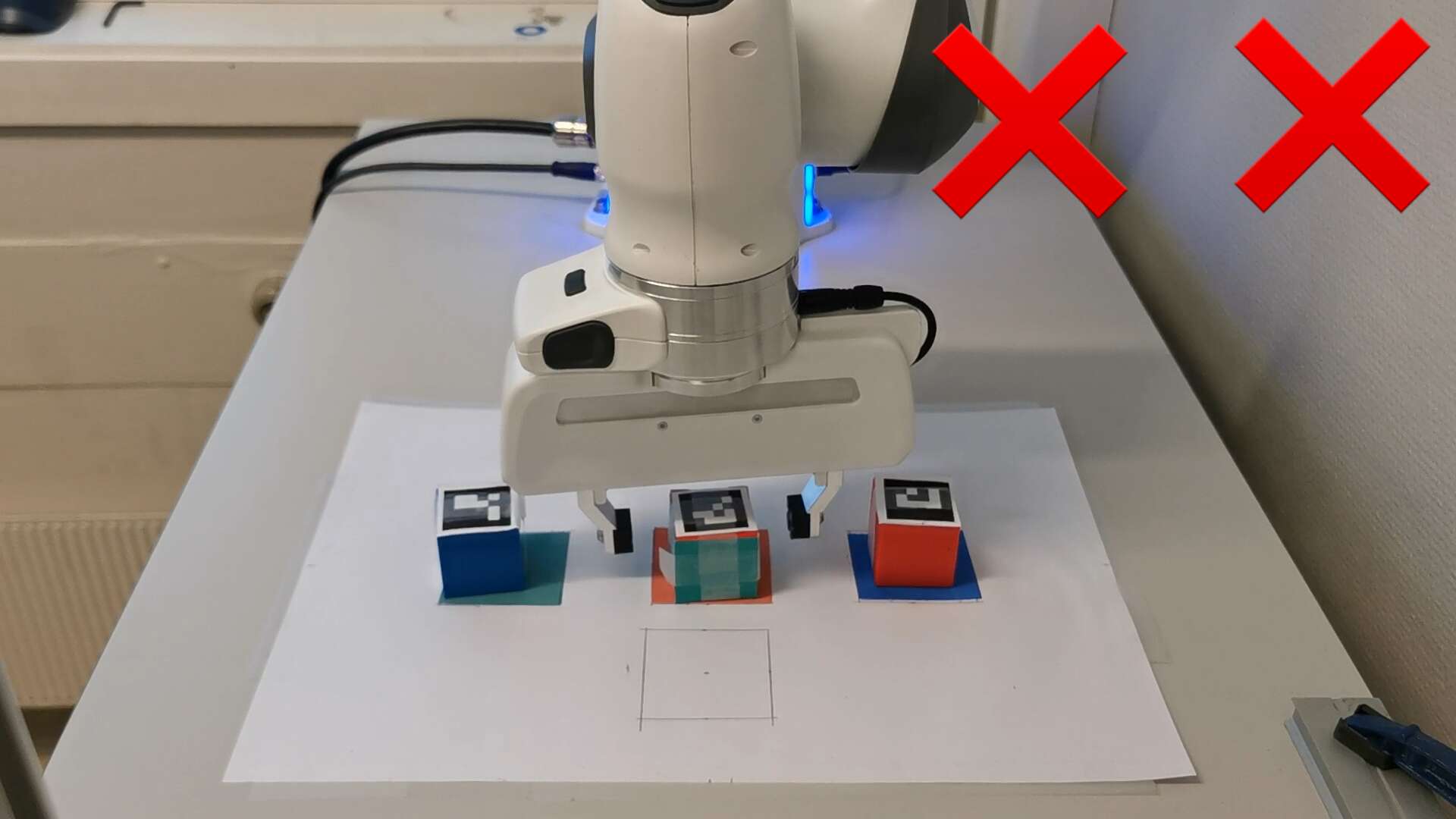}}
        \end{tabular}
        
        \label{fig:teaser_diffuser}
    \end{subfigure}
    \hspace{20pt}
    \begin{subfigure}[t]{0.45\textwidth}
        \centering
        \hspace{-47pt}
        \includegraphics[width=1.0\linewidth]{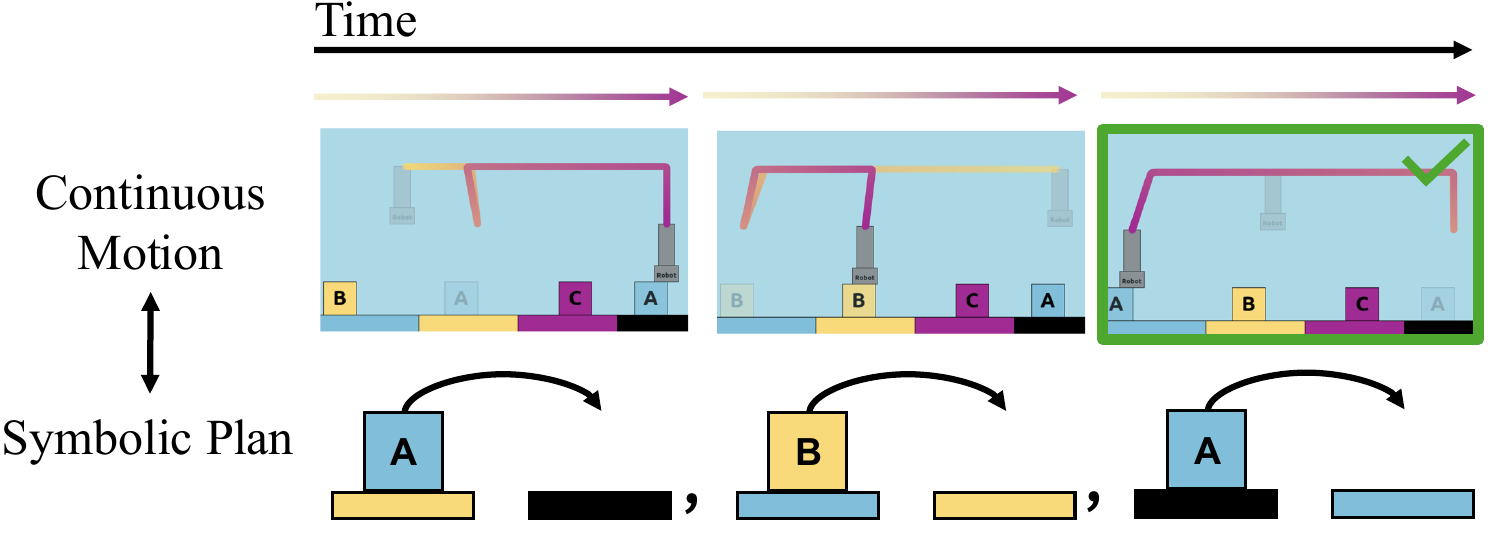}

        \vspace{10pt}
        
        \begin{tabular}{c@{\hspace{2pt}}c@{\hspace{2pt}}c@{\hspace{2pt}}c}
            \adjustbox{trim={0.2\width} {0} {0.2\width} {0}, clip}{\includegraphics[width=\imwidth]{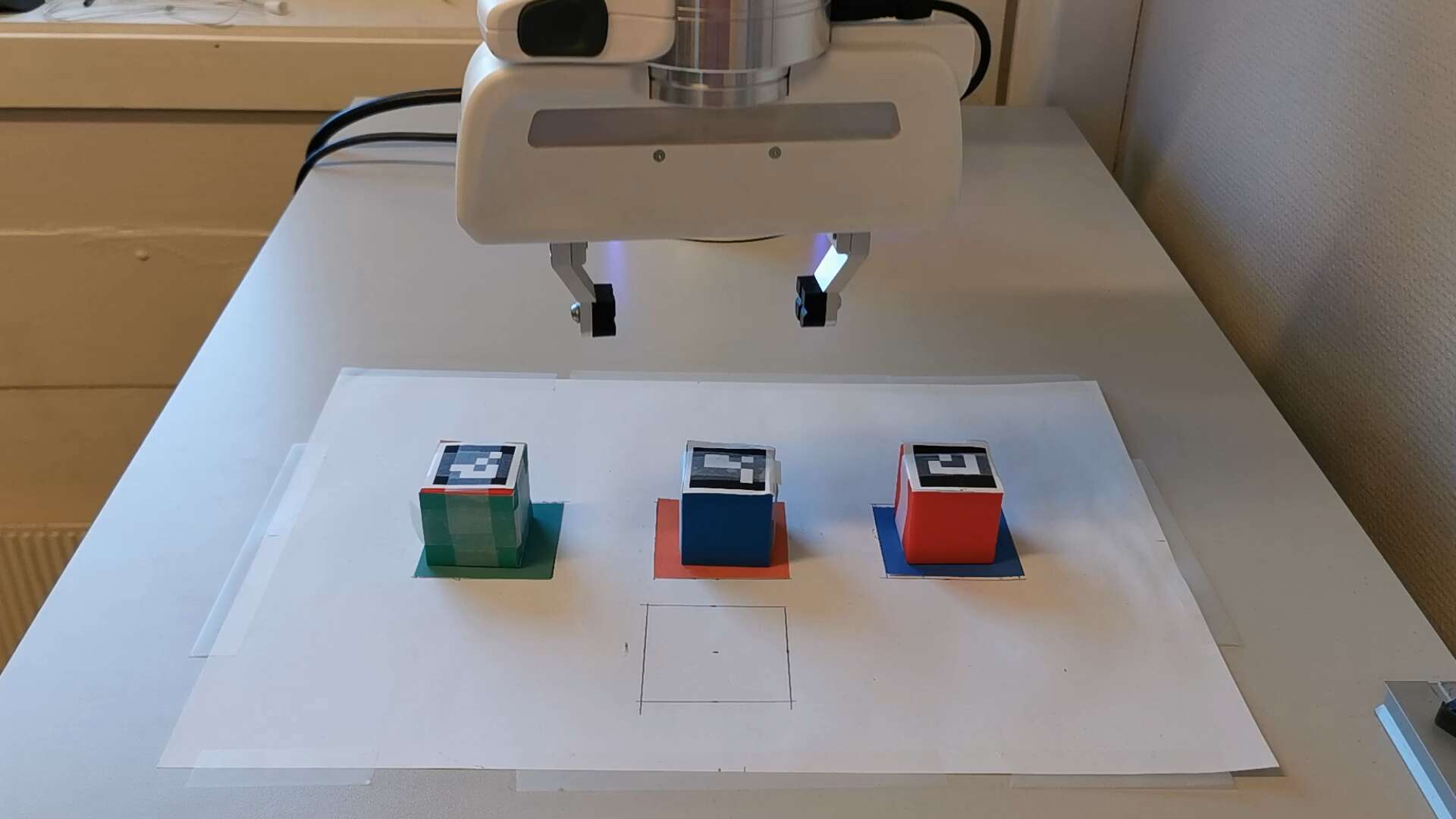}} &
            \adjustbox{trim={0.2\width} {0} {0.2\width} {0}, clip}{\includegraphics[width=\imwidth]{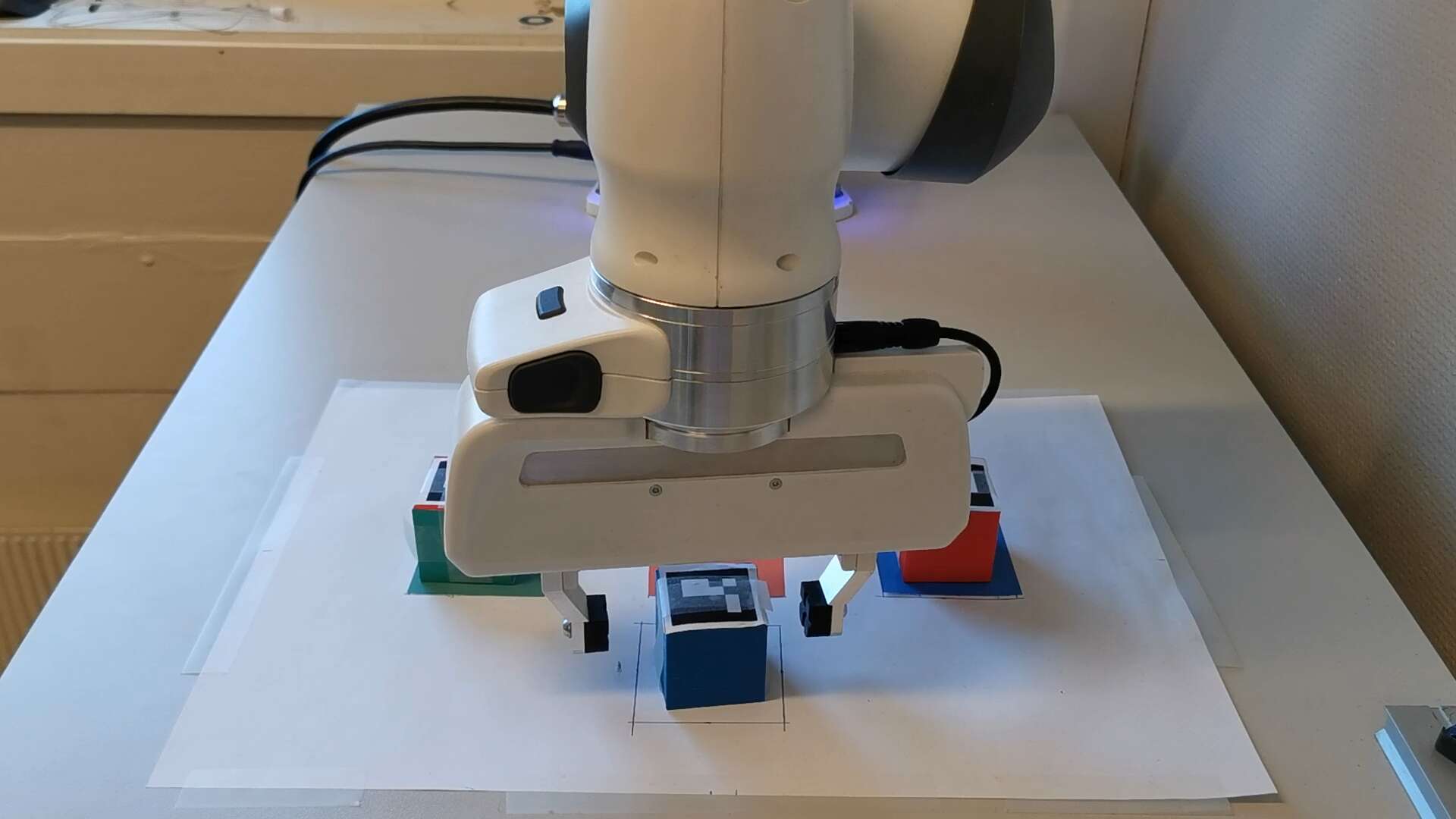}} &
            \adjustbox{trim={0.2\width} {0} {0.2\width} {0}, clip}{\includegraphics[width=\imwidth]{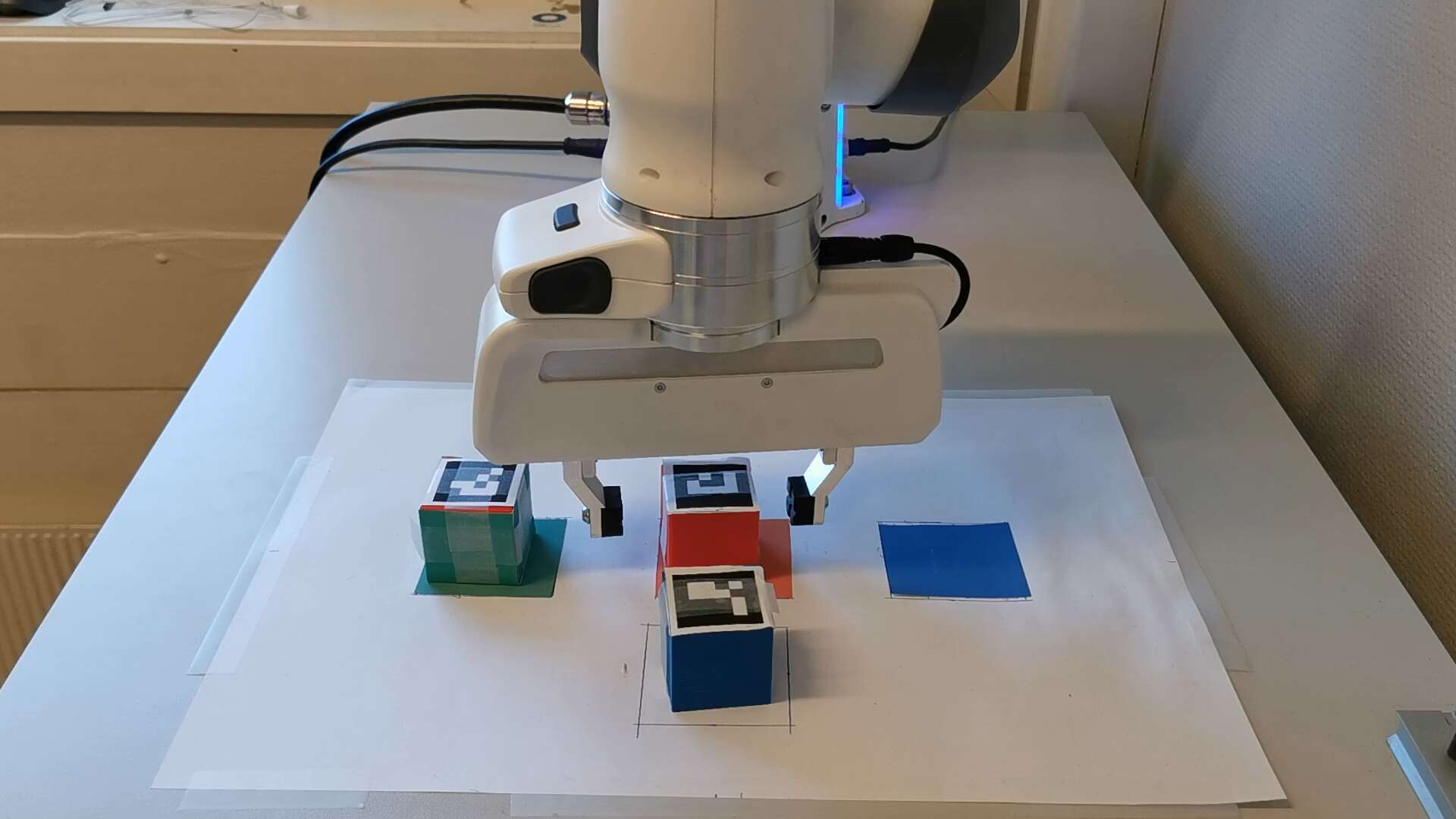}} &
            \adjustbox{trim={0.2\width} {0} {0.2\width} {0}, clip}{\includegraphics[width=\imwidth]{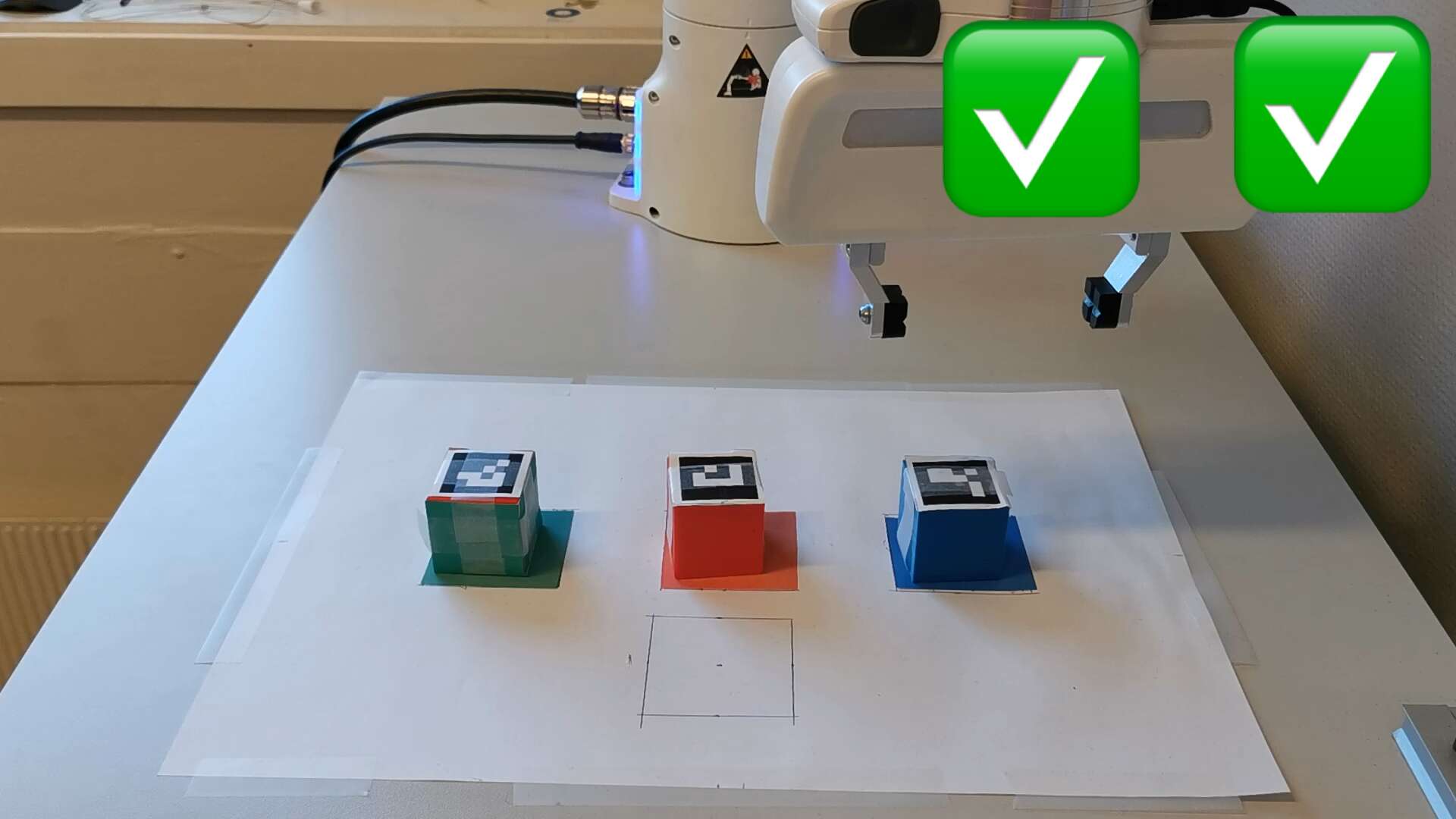}}
        \end{tabular}
        
        \label{fig:teaser_ours}
    \end{subfigure}
    
    \caption{\textit{Left:} \textbf{Diffusion-based planning fails long-horizon decision-making tasks.} In our simulated and real experiments, Diffuser~\cite{jannerPlanningDiffusionFlexible2022a} fails the task of sorting three blocks. Despite all trajectories in the dataset solving the task, sampled trajectories do not. Comparison of diffusion-based planning approaches for long-horizon robotic tasks. \textit{Right:} \textbf{Hybrid Diffusion Planning.} Our method, \modellong, jointly constructs symbolic and discrete plans, enabling more robust performance over long-horizon tasks.} 
    \label{fig:teaser}

\end{figure*}

\begin{abstract}
Constructing robots to accomplish long-horizon tasks is a long-standing challenge within artificial intelligence. Approaches using generative methods, particularly Diffusion Models, have gained attention due to their ability to model continuous robotic trajectories for planning and control. However, we show that these models struggle with long-horizon tasks that involve complex decision-making and, in general, are prone to confusing different modes of behavior, leading to failure. To remedy this, we propose to augment continuous trajectory generation by simultaneously generating a high-level symbolic plan. We show that this requires a novel mix of discrete variable diffusion and continuous diffusion, which dramatically outperforms the baselines. In addition, we illustrate how this hybrid diffusion process enables flexible trajectory synthesis, allowing us to condition synthesized actions on partial and complete symbolic conditions. Project website: \href{https://sigmundhh.com/hybrid_diffusion/}{sigmundhh.com/hybrid\_diffusion}.
\end{abstract}

\IEEEpeerreviewmaketitle

\pagestyle{empty} 
\thispagestyle{empty}

\section{Introduction}
\label{sec:intro}

In the quest for general-purpose robotics, learning from demonstrations has proven a widely applicable paradigm. The primary task of imitation learning is to absorb a large number of demonstrations involving diverse behaviors. A performant and widely used technique for this task is to apply diffusion models~\cite{hoDenoisingDiffusionProbabilistic2020a} for modeling robotic behavior. In addition to handling multimodal behavior~\cite{chiDiffusionPolicyVisuomotor2023b}, they are stable to train, and allow for flexible guidance through conditioning and composition~\cite{jannerPlanningDiffusionFlexible2022a, ajayConditionalGenerativeModeling2023, luo_potential_2024,carvalhoMotionPlanningDiffusion2023}. They are, as a result, ubiquitous in a number of robotic systems, such as open-loop trajectory modelling~\cite{jannerPlanningDiffusionFlexible2022a, dong_diffuserlite_2024, luo_potential_2024}, closed-loop action inference using image-conditioning~\cite{chiDiffusionPolicyVisuomotor2023b, reussGoalConditionedImitationLearning2023a}, or as modules in composite systems~\cite{xian_chaineddiffuser_2023, mishra_generative_2023, luo_generative_2025}.

However, diffusion models often struggle to form long-horizon, non-smooth plans, which restricts them to only modeling relatively short and dense trajectories in Cartesian space \cite{ajayConditionalGenerativeModeling2023}. This limits their ability to do long-horizon decision-making tasks. A motivating example is shown on the left of Figure \ref{fig:teaser}, where a trajectory-level diffusion model is tasked with sorting three blocks. Despite the demonstrations always terminating in a sorted state, sampled trajectories from the planner fail to sort the blocks. This is exacerbated when task complexity is increased, as the diffusion model struggles to account for interdependencies over long time horizons. Indeed, our experiments (Sec. \ref{sec:result}) demonstrate that when tasked with sorting an increasing number of blocks, pure diffusion models quickly fail.

A popular paradigm for allowing robots to perform long-horizon decision-making tasks is Task-and-Motion Planning (TAMP). TAMP methods typically exploit the connection between symbolic and continuous motion plans to simplify and reduce the overall size of the search space~\cite{garrett_integrated_2020}. For example, symbolic planners can construct symbolic abstracted plans that transfer the system to the goal state, while continuous motion planning can find motion plans that correspond to this symbolic plan~\cite{garrett_pddlstream_2020}. The inclusion of planning in symbolic space not only increases the planner's efficiency and performance, but it also allows for transparency, unlike that of pure continuous planners. Upon generation, the symbolic plan provides a clear indication of the high-level steps involved, offering clarity. Additionally, their connection between continuous and symbolic plans allows for direct control of the robot by modifying parts of the symbolic plan and having the continuous plan respect these restrictions. For example, when a robot is tasked with moving three boxes from one location to another, we may want to specify at test time that a particular box should be moved first. It would be of interest to combine these techniques with planning using diffusion models to make them more transparent by providing a symbolic description of the robot plan and allowing for guidance and conditioning at a symbolic level.

As a response, we present the \modellong (\model), a performant method for simultaneously generating both continuous and symbolic plans, as shown on the right of Fig. \ref{fig:teaser}. Its connection to symbolic plans enables unprecedented transparency and guidance functionality. Surprisingly, incorporating modeling of symbolic information with \model improves the long-horizon planning performance drastically compared to motion-only diffusion.  Through our experiments, we show that our formulation of a joint objective consisting of masked diffusion~\cite{shi_simplified_2024} and continuous diffusion~\cite{hoDenoisingDiffusionProbabilistic2020a} is crucial to the success of the method. In addition to the performance gain, \model enables flexible conditional sampling at inference. By fixing a partial or complete symbolic plan through inpainting, \model can generate a continuous plan that satisfies the specified constraints. Such flexible conditioning allows \model to be easily controlled and used for diverse tasks outside of explicit plan generation.

To further highlight the challenges of long-horizon planning and show the benefits of \model, we present a novel task suite of simulated and real robotic tasks focused on long-horizon complex planning. Previous IL benchmarks either focus on single-task performance~\cite{chiDiffusionPolicyVisuomotor2023b} or, when considering long-horizon operations, the subtasks are specified to the policy by an external oracle~\cite{meesCALVINBenchmarkLanguageConditioned2022, gupta_relay_2019}. In contrast, we focus on long-horizon robotic manipulation tasks where the \textit{agent is tasked with determining the sequence of subtasks itself}. We find that traditional diffusion-based planning from demonstrations performs poorly in this scenario.

Overall, our contributions are threefold: \textbf{(1)} We introduce a novel Imitation Learning task suite, exhibiting complex and long-horizon planning for robotic manipulation, and show that widely used diffusion-model planning struggles in the face of long-horizon, multimodal demonstrations. \textbf{(2)} We introduce the \modellong, a novel diffusion-based planner that uses a coupled discrete and continuous diffusion process for generating both symbolic and continuous motion plans. \textbf{(3)} Lastly, we empirically demonstrate \model's flexible conditioning capabilities.
 
\vspace{-5pt}
\section{Related Work}

\paragraph{Planning with Diffusion Models} Several works have shown that Diffusion Models~\cite{hoDenoisingDiffusionProbabilistic2020a} excel at modelling distributions over trajectories \cite{jannerPlanningDiffusionFlexible2022a, pearceImitatingHumanBehaviour2023}, with Diffuser by Janner et al. \cite{jannerPlanningDiffusionFlexible2022a} as a seminal work showing their application to planning. Beneficial for planning is diffusion models' flexibility during sampling, such as inpainting~\cite{jannerPlanningDiffusionFlexible2022a}, composition with auxiliary cost functions~\cite{carvalhoMotionPlanningDiffusion2023, saha_edmp_2023, luo_potential_2024}, and classifier-free guidance~\cite{ajayConditionalGenerativeModeling2023}, and they have, as such, been widely applied in robotic systems~\cite{ubukata_diffusion_2024}. However, as we show, when task complexity and time horizon increase, diffusion models for planning struggle to form valid plans. Related to our approach are \textit{hierarchical diffusion models} \cite{pmlr-v202-li23ad, chen2024simple, hao2025chd}, which aim to improve long-horizon trajectory planning by predicting coarse, subsampled trajectories using a global high-level planner, and using a local low-level planner to predict dense trajectories in a segment-wise manner. In contrast, our approach jointly models symbolic and continuous plans without explicit plan decomposition.

\paragraph{Long-horizon Planning} TAMP-methods~\cite {garrett_integrated_2020} solve long-horizon robotic tasks efficiently. However, current methods pose assumptions that require significant engineering effort when applied to new tasks. Specifically, methods relying on PDDLStream~\cite{garrett_pddlstream_2020} require upfront specification of the transition model, action primitives with preconditions and effects, procedures for determining predicates, streams for sampling, and the types of objects in the scene~\cite{garrett_online_2020, fang_dimsam_2024}. While several methods have been proposed to lower the barrier of application to new environments, they still require specification of action schemas~\cite{mandlekar_human---loop_2023}, predicates~\cite{mandlekar_human---loop_2023, zhou_spire_2024}, an action skeleton~\cite{mishra_generative_2023}, the transition model~\cite{silver_predicate_2025}, or the sets of objects~\cite{shah_real_2025}, resulting in a need for hand-engineering for new tasks or environments. Our focus is, however, on methods that learn purely from demonstrations, and \model learns long-horizon behavior directly from data, allowing for straightforward application to new tasks. Works such as Transporter Networks~\cite{zengTransporterNetworksRearranging2022} can perform long-horizon tasks from demonstrations by training affordance functions for robot manipulators, for instance, for tasks like picking and placing. We aim to improve the capabilities of long-horizon diffusion models to solve similar tasks successfully and still allow for flexible conditioning and guidance.

\section{Hybrid Diffusion Planning}
\label{sec:method}
\vspace{-3pt}

We are interested in solving the task of long-horizon planning of robotic motion given demonstration data. Given an initial observation of the environment $\mathbf{O}$, the planner is tasked to predict a feasible continuous plan over a large number of time steps $T$. Specifically, our overall goal is to generate an action trajectory  $\mathbf{A}_c \in \R^{T \times D_a}$ that results in the robot completing the task, where $D_a$ is the action dimensionality. A novel aspect of our setting is that we also allow the model access to associated symbolic plans, $\mathbf{A}_d \in \R^{T_d \times D_d}$, represented by a string of tokens, each of dimension $D_d$. Notably, the plans do not need to be temporally aligned; i.e., the temporal lengths $T$ and $T_d$ do not have to match.

We present the \modellong (\model) for solving this task. When planning, \model simultaneously predicts both the continuous action trajectory $\mathbf{A}_c$ as well as the symbolic sequence of actions $\mathbf{A}_d$, by modeling the joint distribution over continuous and discrete plans using two coupled diffusion processes. We first describe modeling the continuous plan and discrete plan separately with diffusion in Section~\ref{sect:continuous_diffusion} and \ref{sect:discrete_diffusion}. We then introduce our hybrid diffusion procedure, \model,  which jointly models both discrete and continuous planning simultaneously, enabling us to solve long-horizon planning tasks effectively.

\vspace{-3pt}
\subsection{Modeling Continuous Plans with Continuous Diffusion}
\label{sect:continuous_diffusion}

For modeling continuous motion plans, \model uses a continuous variable diffusion process. Specifically, we apply DDPM~\cite{hoDenoisingDiffusionProbabilistic2020a} for trajectory generation, following Diffuser~\cite{jannerPlanningDiffusionFlexible2022a}. During training, a continuous demonstration trajectory $\mathbf{A}_c$ is added noise with a magnitude proportional to a uniformly sampled diffusion step $k$. Specifically, the noise-corrupted trajectory is sampled from the forward diffusion process
\begin{equation}
q_\text{DDPM}(\mathbf{A}_c^k|\mathbf{A}_c^0) = \mathcal{N}(\mathbf{A}_c^k; \sqrt{\bar{\alpha}_k}\mathbf{A}_c^0, (1-\bar{\alpha}_k)\mathbf{I}),
\label{eqn:continuous_noise}
\end{equation}
where $\alpha_k$ is given by the noise schedule and determines the signal-to-noise ratio for a given diffusion step. Then, given the corrupted sequence, the model $\epsilon_\theta$ is tasked to predict the noise component $\epsilon$ given a noisy sample 
\begin{equation}
\hat{\epsilon}=\epsilon_{\theta, \text{DDPM}}(\mathbf{A}_c^k,k), \quad \mathcal{L}_\text{DDPM} = MSE(\epsilon, \hat{\epsilon}).
\label{eqn:loss_continuous}
\end{equation}
Sampling is initiated from a pure-noise sample $\mathbf{A}_c^{K}$ and is iteratively updated. At each iteration, the trained model outputs the noise component $\hat{\epsilon}$, and $\mathbf{A}_c^{k-1}$ is sampled from 
\begin{equation}
    p_\text{DDPM}(\mathbf{A}_c^{k-1}|\mathbf{A}_c^k, \hat{\epsilon}) = \mathcal{N}(\mathbf{A}_c^{k-1};\mu(\mathbf{A}_c^k, k, \hat{\epsilon}), \sigma_k^2 \mathbf{I}),
    \label{eqn:continuous_sample}
\end{equation}
where $\mu(\mathbf{A}_c^k,k, \hat{\epsilon})=\frac{1}{\sqrt{\alpha_k}}\left(\mathbf{A}_c^k-\xi_k \hat{\epsilon}\right)$ is the predicted mean. The coefficients $\xi_k, \sigma_k$ are given by the diffusion schedule.

\subsection{Modeling the Symbolic Plans with Masked Diffusion}
\label{sect:discrete_diffusion}

\looseness=-1 For modeling the discrete action plan, \model employs MD4~\cite{shi_simplified_2024}, which provides a simple and performant framework for the diffusion of discrete variables. This is done through \textit{masked diffusion}, where the forward diffusion process is defined such that each token is masked with an increasing probability when moving along the diffusion axis. This is done by representing each action $\mathbf{A}_{d,i}$ in the discrete action sequence as one-hot encoded variables with $m+1$ possible states, where $\{e_0 \dots e_{m-1} \}$ correspond to the $m$ symbols in the vocabulary and the last state to a masked state $e_m$. The transition matrix $\bar{Q}(k)$ transfers tokens to this absorbing masked state with a probability given by $k$: 
\begin{equation}
q_\text{MD4}(\mathbf{A}_d^k|\mathbf{A}_d^0) = \text{Cat}(\mathbf{A}_d^k; \bar{Q}(k)^\top \mathbf{A}_d^0).
\label{eqn:discrete_noise}
\end{equation}
Here, $\text{Cat}(x;p)$ denotes a categorical distribution where $p$ is the vector of probabilities and the transition matrix is $\bar{Q}(k)= \alpha_k I + (1-\alpha_k)\mathbf{1} e_m^\top$, which places increasing weight on the masked state at higher diffusion steps.  As in continuous variable diffusion, $\alpha_k$ is given by the diffusion schedule.

The learned reverse process is parameterized with a network predicting logits over all possible symbols in the vocabulary, $\hat{\mu} = \mu_\theta(\mathbf{A}_{d}^k,k) \in \R^{m+1}$, where the probability of the masked state is set to zero. During training, the model is trained with a cross-entropy loss over the masked tokens in a partially-masked sequence
\begin{equation}
\mathcal{L}_\text{MD4} = \sum_{i:\mathbf{A}_{d,i}^{k}=m} w_k \mathcal{L}_\text{cross-entropy}(\hat{\mu}, \mathbf{A}_{d,i}^0),
\label{eqn:loss_discrete}
\end{equation}
where $w_k$ is a weighting term in the diffusion schedule. 

\noindent At sampling time, the sequence is instantiated as a fully masked sequence, and the tokens will be sampled from the predicted categorical distribution over all tokens in the vocabulary 
\begin{equation}\label{eqn:sample_discrete}
\begin{split}
  p_\text{MD4}(\mathbf{A}_d^{k-1}|\mathbf{A}_d^k, \hat{\mu}) &=\text{Cat}(\mathbf{A}_d^{k-1}; \bar{R}^\top \mathbf{A}_d^k), \\ \text{where } \bar{R} &=I+\gamma_k e_m(\hat{\mu} - e_m), 
\end{split}
\end{equation}
and $\gamma_k$ is given by the diffusion schedule. Intuitively, a sample will, with a given probability, transfer from the masked class to a sample from the predicted categorical distribution, at each reverse step. The diffusion schedule is designed to reveal tokens with increasing probability when moving backward along the diffusion axis~\cite{shi_simplified_2024}.

\subsection{Hybrid Diffusion Planning: Jointly Modeling Continuous and Discrete Plans}
\looseness=-1
The \modellong (\model) models two diffusion processes, over both the continuous plan $\mathbf{A}_c$ and the discrete plan $\mathbf{A}_d$. A unified model consumes both plan modalities, enabling \model to learn how they correspond to each other.
\looseness=-1
Given a corresponding pair of continuous and discrete plans $\mathbf{A}_c$ and $\mathbf{A}_d$, the forward diffusion process corrupts both plans independently by adding noise to the continuous plan (Eq.~\ref{eqn:continuous_noise}), and masking out tokens in the discrete plan (Eq.~\ref{eqn:discrete_noise})
\begin{multline}
q(\mathbf{A}_c^{k_c}, \mathbf{A}_d^{k_d} | \mathbf{A}_c^0, \mathbf{A}_d^0) \\ = 
\mathcal{N}(\mathbf{A}_c^{k_c}; \sqrt{\bar{\alpha}_{k_c}}\mathbf{A}_c^0, (1-\bar{\alpha}_{k_c})\mathbf{I}) \cdot 
\text{Cat}(\mathbf{A}_d^{k_d}; \mathbf{\bar{Q}}({k_d})^\top \mathbf{A}_d^0).
\raisetag{2\baselineskip}
\end{multline}
Given the independence of the forward process, we can split up the reverse process into two independent terms~\cite{furrutter2024quantum}
\begin{multline} \label{eqn:reverse_step}
    p(\mathbf{A}_c^{k_c-1}, \mathbf{A}_d^{k_d-1} | \mathbf{A}_c^{k_c}, \mathbf{A}_d^{k_d}, \mathbf{O}) \\ 
    = p(\mathbf{A}_c^{k_c-1}|\mathbf{A}_c^{k_c}, \mathbf{A}_d^{k_d},  \mathbf{O}) \cdot p(\mathbf{A}_d^{k_d-1},|\mathbf{A}_c^{k_c}, \mathbf{A}_d^{k_d}, \mathbf{O}) \\
    = \mathcal{N}(\mathbf{A}_c^{k_c-1}; \mu(\mathbf{A}_c^{k_c}, k_c, \hat{\epsilon}), \sigma_{k_c}^2 \mathbf{I}) \\
    \cdot \text{Cat}(\mathbf{A}_d^{k_d-1}; \mathbf{R}_{k_d}(\hat{\mu})^\top \mathbf{A}_d^{k_d}).
\end{multline}
This allows us to sample from the reverse process by sampling independently from Eq. \ref{eqn:continuous_sample} and Eq. \ref{eqn:sample_discrete}, given a prediction of $\hat{\epsilon}$ and $\hat{\mu}$. Importantly, these are obtained \textit{jointly} with a single denoising model
\begin{equation}
(\hat{\epsilon}, \hat{\mu}) = D_{\theta}(\mathbf{A}_c^{k_c}, \mathbf{A}_d^{k_d}, \mathbf{O}, k_c, k_d).
\end{equation}
By accepting the concatenation of both plan modalities, the above formulation enables the denoiser to leverage information across both plans to accurately denoise $\mathbf{A}_c^{k_c}$ and $\mathbf{A}_d^{k_d}$.
\begin{figure}[t]
    \centering
    \vspace{5pt}
    \includegraphics[width=\linewidth]{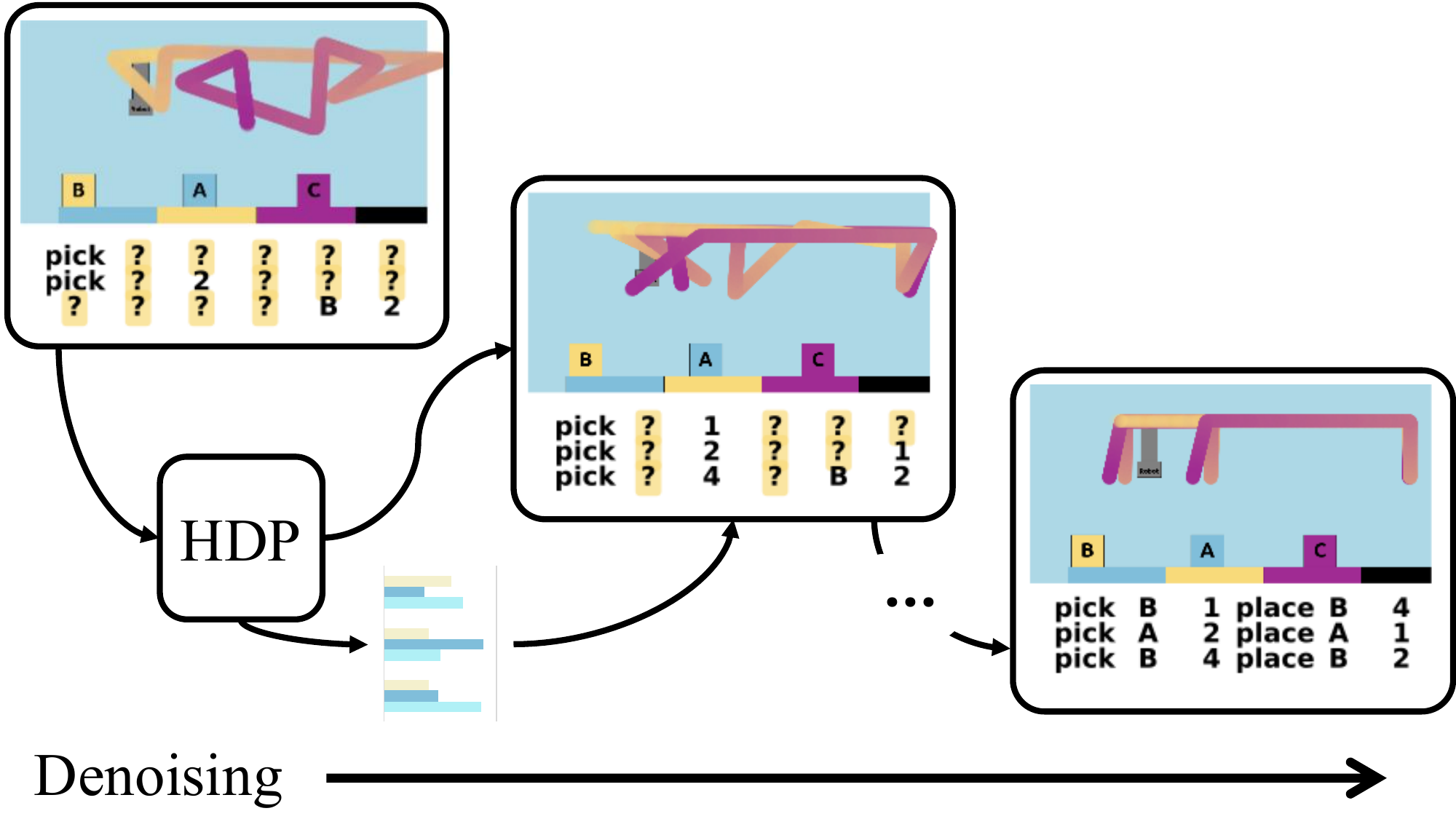}
    \caption{\small \textbf{Parallel denoising process} of symbolic and robot motion plans. \model consumes both the noisy continuous motion plan and the masked symbolic plan. At each sampling step, it partly denoises the continuous motion and partly unmasks the symbolic plan.}
    \label{fig:denoising_process}
\end{figure}
 \noindent \textbf{Training.} To train the denoising network, we use a weighted sum of the continuous denoising loss on $\mathbf{A}_c^{k_c}$ (Equation~\ref{eqn:loss_continuous}) and the discrete denoising loss on $\mathbf{A}_d^{k_d}$ (Equation~\ref{eqn:loss_discrete})
\begin{equation}
\mathcal{L} = \mathcal{L}_\text{DDPM} + \lambda \mathcal{L}_\text{MD4},
\end{equation}
where we set $\lambda =\frac{1}{30}$ to balance out the magnitude of the loss terms. A training overview is outlined in Algorithm~\ref{alg:hybrid_diffusion_training}. 

\looseness=-1
During training, both plans are corrupted using the forward process. Notably, the corruption levels are sampled independently, \(k_c,k_d \sim \mathcal{U}[0,K]\), allowing for different degrees of corruption for each plan modality. This leads the model to use information from one plan modality to uncorrupt the other. For example, if the discrete plan is fully unmasked ($k_d=0$) during a training iteration, the denoiser can learn to exploit the information in the discrete sequence when denoising the continuous trajectory. In contrast, when the discrete plan is fully masked ($k_d=K$), the denoiser learn to denoise the continuous trajectory unconditionally without relying on the masked information. This technique is related to Diffusion Forcing~\cite{chen2024diffusion}, which also learns correspondences through independent levels of corruption. 

\begin{figure}[h]
\vspace{-25pt}
\begin{minipage}{0.9\linewidth}
\begin{algorithm}[H] 
\caption{\textbf{Hybrid Diffusion Training}}
\label{alg:hybrid_diffusion_training}
\begin{algorithmic}[1]
\REQUIRE Dataset $\mathcal{D}$, Denoiser $D_\theta(\cdot)$
\WHILE{not converged}
    \STATE Sample $(\mathbf{A}_c, \mathbf{A}_d, \mathbf{O}) \sim \mathcal{D}$
    \STATE Sample diffusion steps $k_c, k_d \sim \mathcal{U}[0,K]$
    \STATE $\mathbf{A}_c^{k_c} \sim q_\text{DDPM}(\mathbf{A}_c^{k_c}|\mathbf{A}_c)$
    \STATE $\mathbf{A}_d^{k_d} \sim q_\text{MD4}(\mathbf{A}_d^{k_d}|\mathbf{A}_d)$
    \STATE $(\hat{\epsilon}, \hat{\mu}) \gets D_\theta(\mathbf{A}_c^{k_c}, \mathbf{A}_d^{k_d}, k_c, k_d, \mathbf{O})$
    \STATE $\mathcal{L} \gets \mathcal{L}_\text{DDPM} + \lambda \mathcal{L}_\text{MD4}$
    \STATE $\theta \gets \theta - \eta \nabla_\theta \mathcal{L}$
\ENDWHILE
\RETURN $D_\theta(\cdot)$
\end{algorithmic}
\end{algorithm}
\end{minipage}
\end{figure}

\paragraph{Sampling} The full reverse process can be written as
\begin{multline}
    p(\mathbf{A}^{0:K}_c, \mathbf{A}^{0:K}_d| \mathbf{O}) \\ = p(\mathbf{A}^{K}_c)p(\mathbf{A}^{K}_d) \prod_{k=1}^K p_\theta(\mathbf{A}^{k-1}_c, \mathbf{A}^{k-1}_d|\mathbf{A}^{k}_c, \mathbf{A}^{k}_d, \mathbf{O}).
\end{multline}Due to the Markov property, this sequence can be sampled by first sampling fully corrupted plans and then iteratively denoising them simultaneously through $K-1$ steps. This technique is outlined in Algorithm \ref{alg:hybrid_diffusion_planning} and visualized in Fig. \ref{fig:denoising_process}. This enables intermediate plan generation across both modalities to inform each other, allowing the symbolic plan to correspond to the continuous motion and vice versa.

\noindent \paragraph{Conditional sampling} In addition to joint sampling, the reverse step in Eq. \ref{eqn:reverse_step} allows for further sampling schemes. Of particular relevance to robotic planning, it allows for sampling from conditional distributions $p_\theta(\mathbf{A}_d|\mathbf{A}_c)$ given a specified continuous plan $\mathbf{A}_c$ or $p_\theta(\mathbf{A}_c|\mathbf{A}_d)$ given a specified discrete plan $\mathbf{A}_d$. This is achieved by passing the conditioned plan $\mathbf{A}_c$ or $\mathbf{A}_c$ with noise level $k_{\Box}=0, \Box \in \{d,c\}$ to the denoiser, and iteratively sampling the remaining plan modality. Furthermore, we can similarly condition on partially clean or unmasked plans. We showcase this technique and provide quantitative results in Section \ref{sec:cond_sampling}. We also note that extension to reward conditioning for either plan modality is possible, such as demonstrated by Ajay et al.~\cite{ajayConditionalGenerativeModeling2023}

\begin{figure}[t]
    \centering
    \vspace{5pt}
    \includegraphics[width=0.8\linewidth]{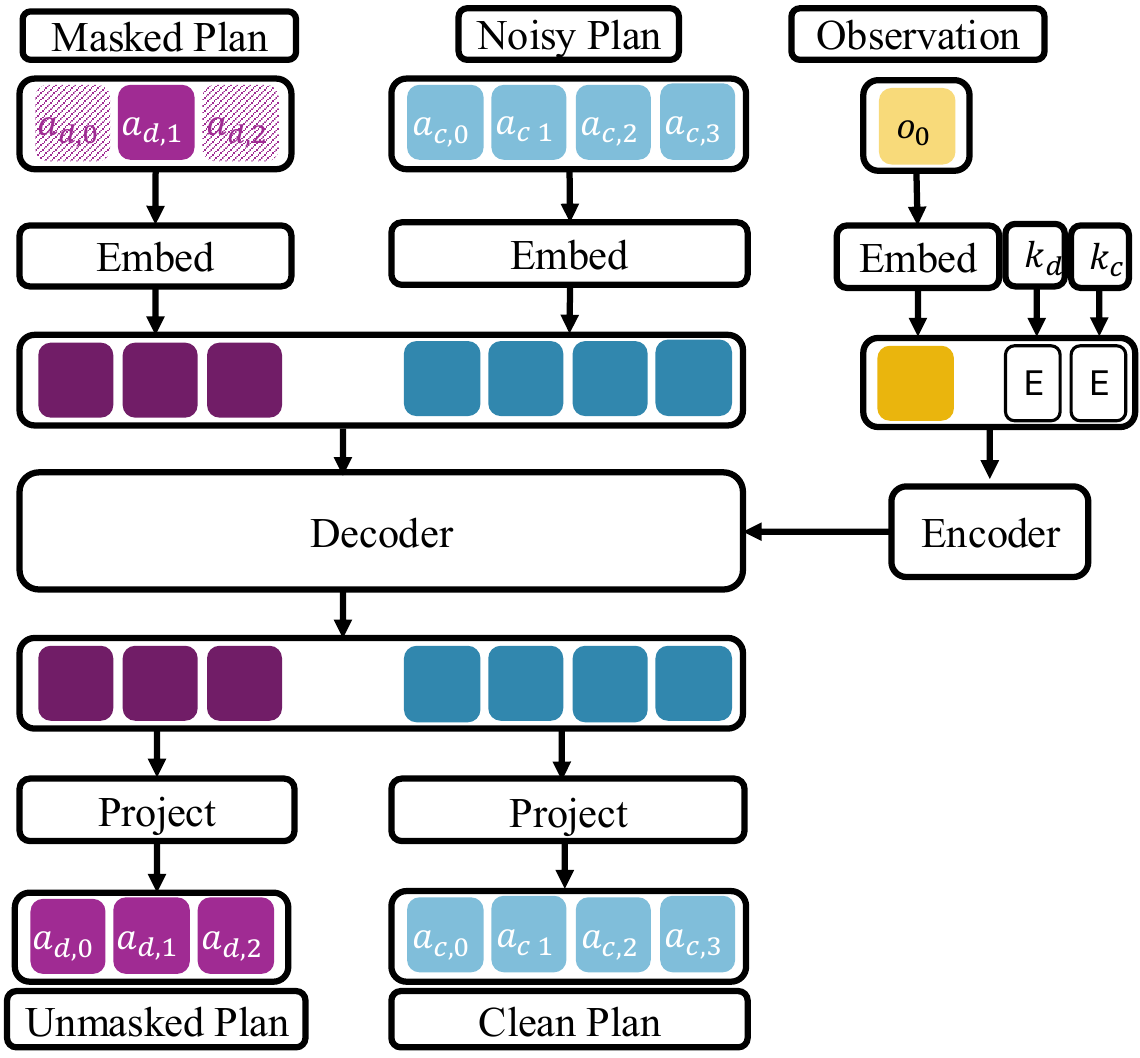}
    \caption{\small \textbf{Architecture of \model.} Allowing for hybrid denoising of the symbolic and continuous plan.}
    \label{fig:architecture}
\end{figure}

\begin{figure}[h]
\vspace{-25pt}
\begin{minipage}{0.9\linewidth}
\begin{algorithm}[H] 
\caption{\textbf{Hybrid Diffusion Planning}}
\label{alg:hybrid_diffusion_planning}
\begin{algorithmic}[1]
\REQUIRE Trained denoiser $D_\theta$, Observation $\mathbf{O}$, Planning horizons $h_c$, $h_d$
\STATE $\mathbf{A}_c \gets \mathcal{N}(\mathbf{0}_{h_c}, \mathbf{I}_{h_c})$
\STATE $\mathbf{A}_d \gets [e_m]_{h_d}$
\FOR{each diffusion step}
    \STATE $(\hat{\epsilon}, \hat{\mu}) \gets D_\theta(\mathbf{A}_c, \mathbf{A}_d, \mathbf{O}, \mathbf{k}, \mathbf{k})$
    \STATE $\mathbf{A}_c \sim p_\text{DDPM}(\mathbf{A}_c^{k-1} | \mathbf{A}_c^k, \hat{\epsilon})$
    \STATE $\mathbf{A}_d \sim p_\text{MD4}(\mathbf{A}_d^{k-1} | \mathbf{A}_d^k, \hat{\mu})$
\ENDFOR
\RETURN $\mathbf{A}_c, \mathbf{A}_d$
\end{algorithmic}
\end{algorithm}
\end{minipage}
\end{figure}

\newcommand{\numberedImg}[2]{%
  \begin{tikzpicture}[baseline, inner sep=0pt, outer sep=0pt]
    \node[anchor=south west, inner sep=0] (image) at (0,0) {#2};
    \node[anchor=south west, fill=white, fill opacity=0.7, text opacity=1, inner sep=2pt, font=\bfseries\sffamily\small] at (image.south west) {#1};
  \end{tikzpicture}%
}

\begin{figure*}[t]
    \vspace{10pt}
    \begin{subfigure}[t]{0.32\textwidth}
        \centering
        \setlength{\tabcolsep}{1pt}
        \renewcommand{\arraystretch}{0}
        \begin{tabular}{ccc}
            \numberedImg{1}{\includegraphics[width=0.4\linewidth,trim={200 200 200 200},clip]{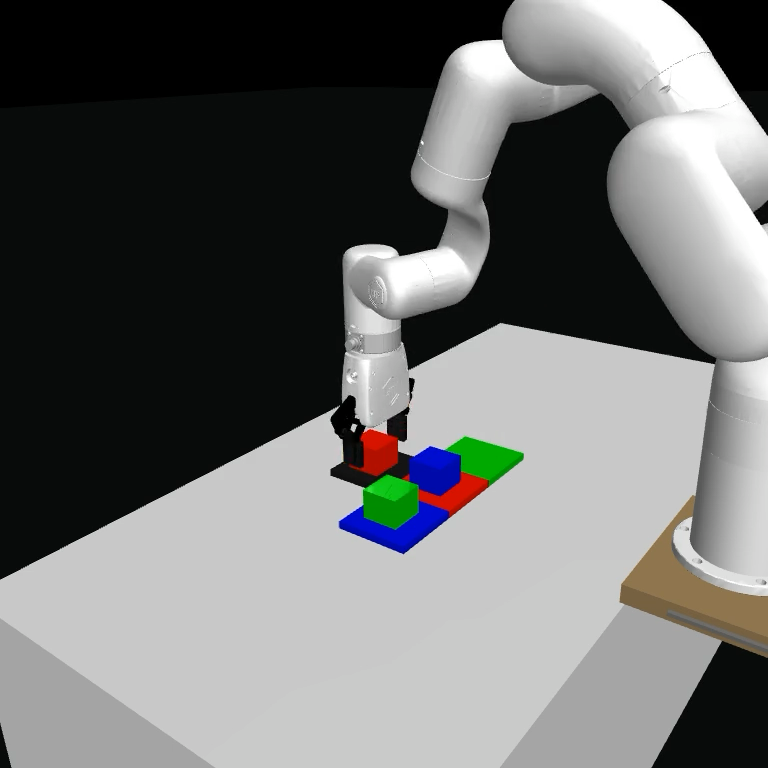}} &
            \numberedImg{2}{\includegraphics[width=0.4\linewidth,trim={200 200 200 200},clip]{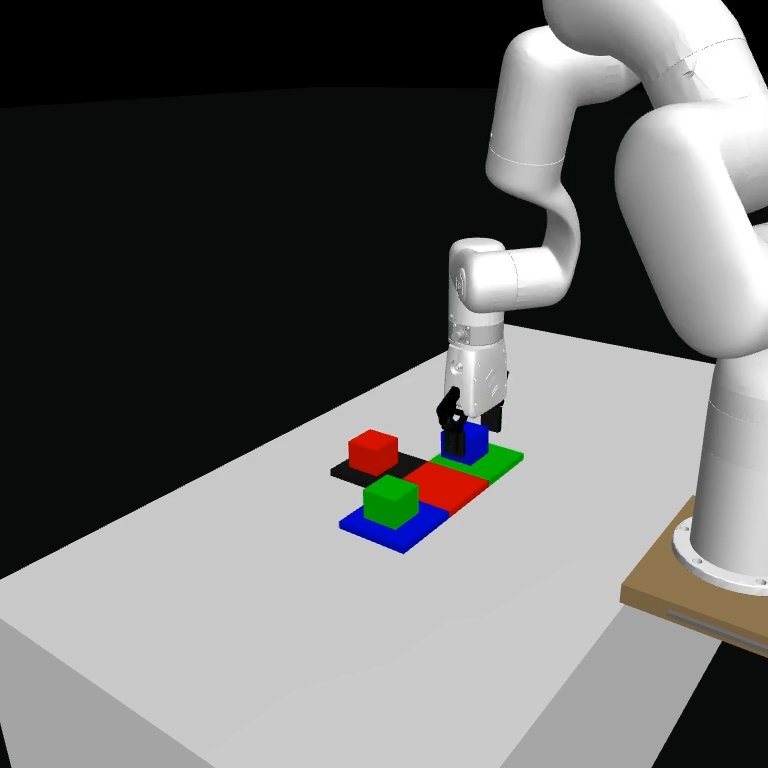}} &
            \numberedImg{3}{\includegraphics[width=0.4\linewidth,trim={200 200 200 200},clip]{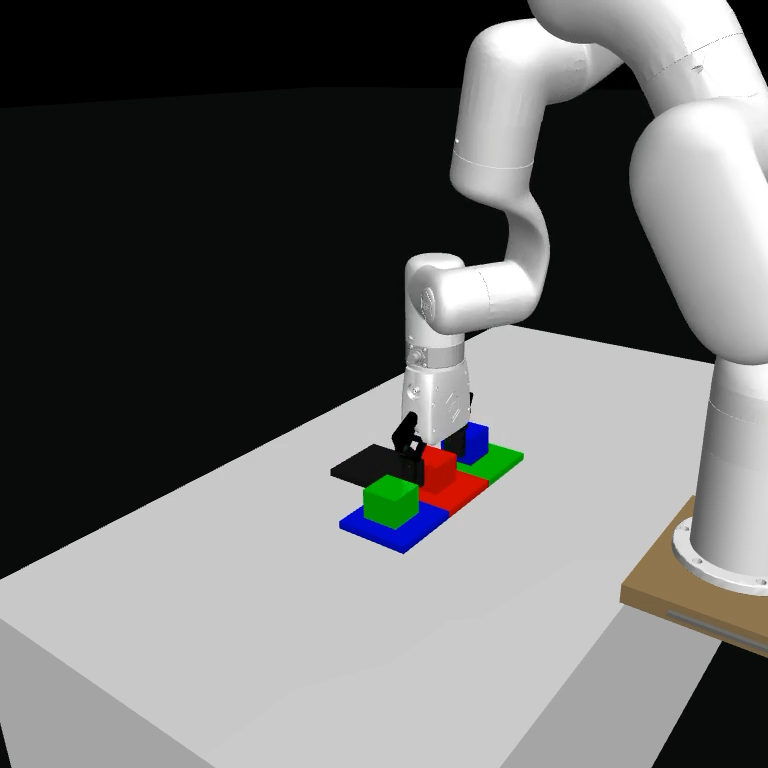}} \\
            \numberedImg{4}{\includegraphics[width=0.4\linewidth,trim={200 200 200 200},clip]{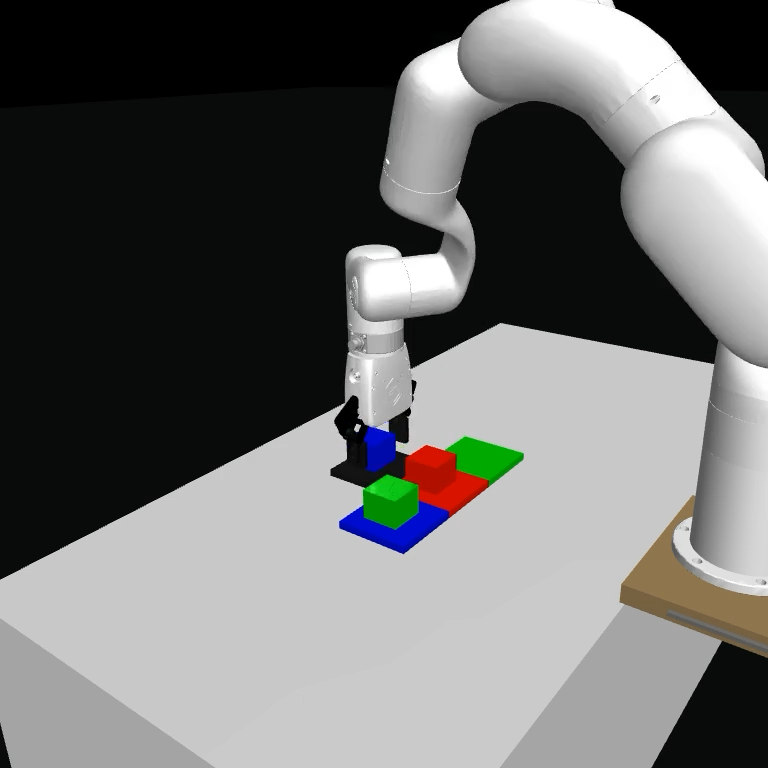}} &
            \numberedImg{5}{\includegraphics[width=0.4\linewidth,trim={200 200 200 200},clip]{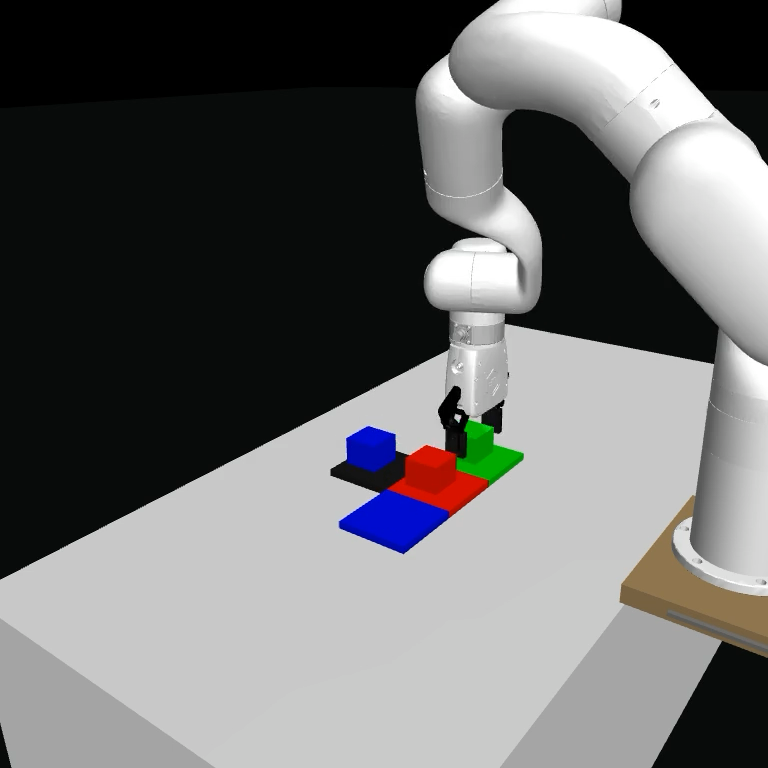}} &
            \numberedImg{6}{\includegraphics[width=0.4\linewidth,trim={200 200 200 200},clip]{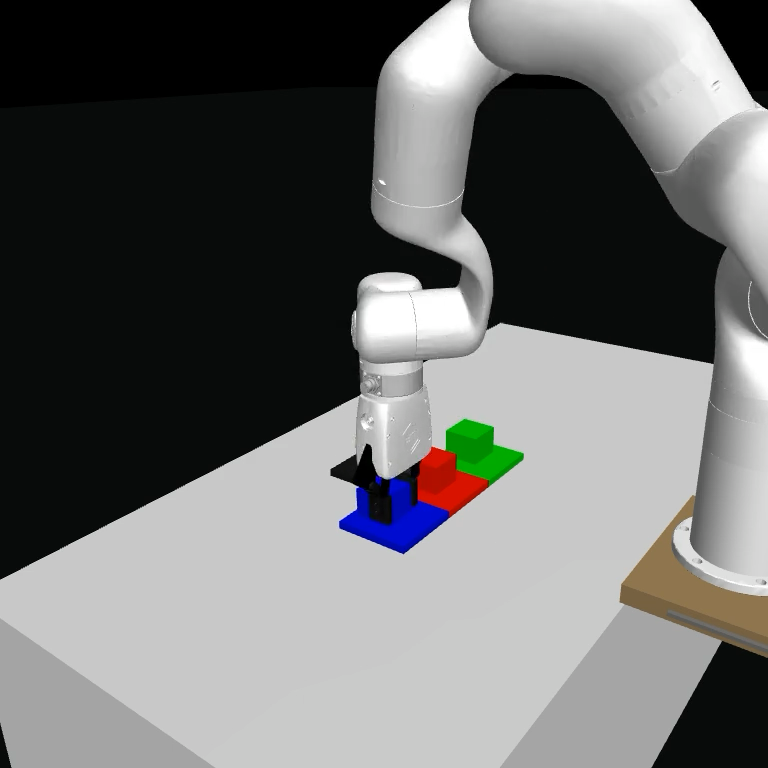}}
        \end{tabular}
        \caption{3 Block Sorting Task}
    \end{subfigure}
    \hspace{0em}
    \begin{subfigure}[t]{0.32\textwidth}
        \setlength{\tabcolsep}{1pt}
        \renewcommand{\arraystretch}{0}
        \centering
        \begin{tabular}{c}
            \numberedImg{1}{\includegraphics[width=0.38\linewidth]{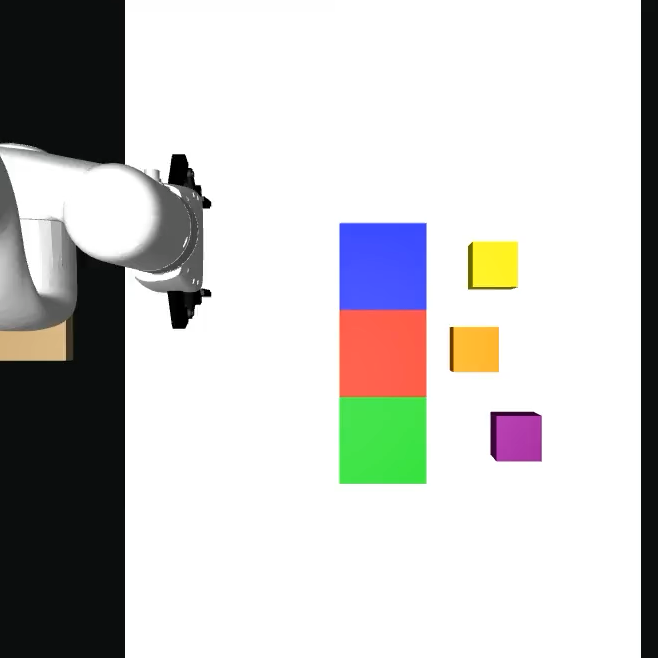}} \\
            \numberedImg{2}{\includegraphics[width=0.38\linewidth]{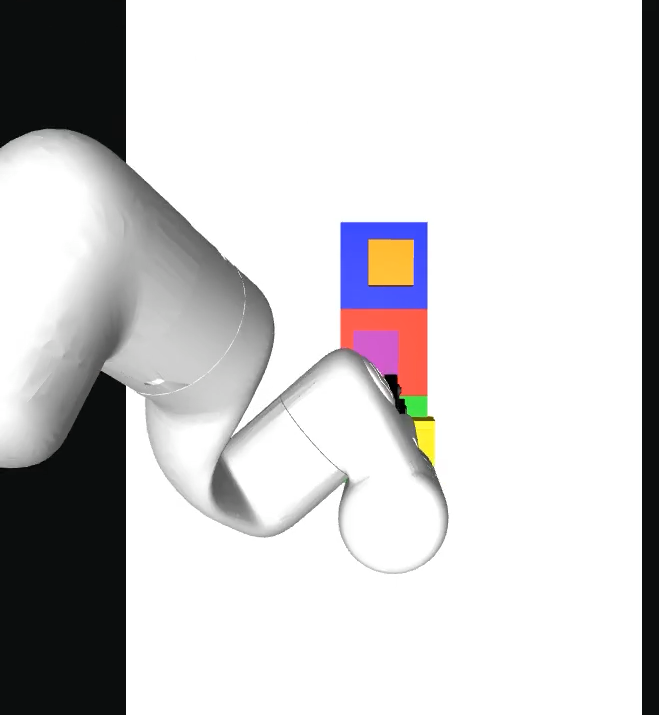}}
        \end{tabular}
        \caption{Arrange Block Task}
    \end{subfigure}
    \hspace{-4em}
    \begin{subfigure}[t]{0.32\textwidth}
        \centering
        \setlength{\tabcolsep}{1pt}
        \renewcommand{\arraystretch}{0}
        \begin{tabular}{ccc}
            \numberedImg{1}{\includegraphics[width=0.4\linewidth, trim=210 90 60 200, clip]{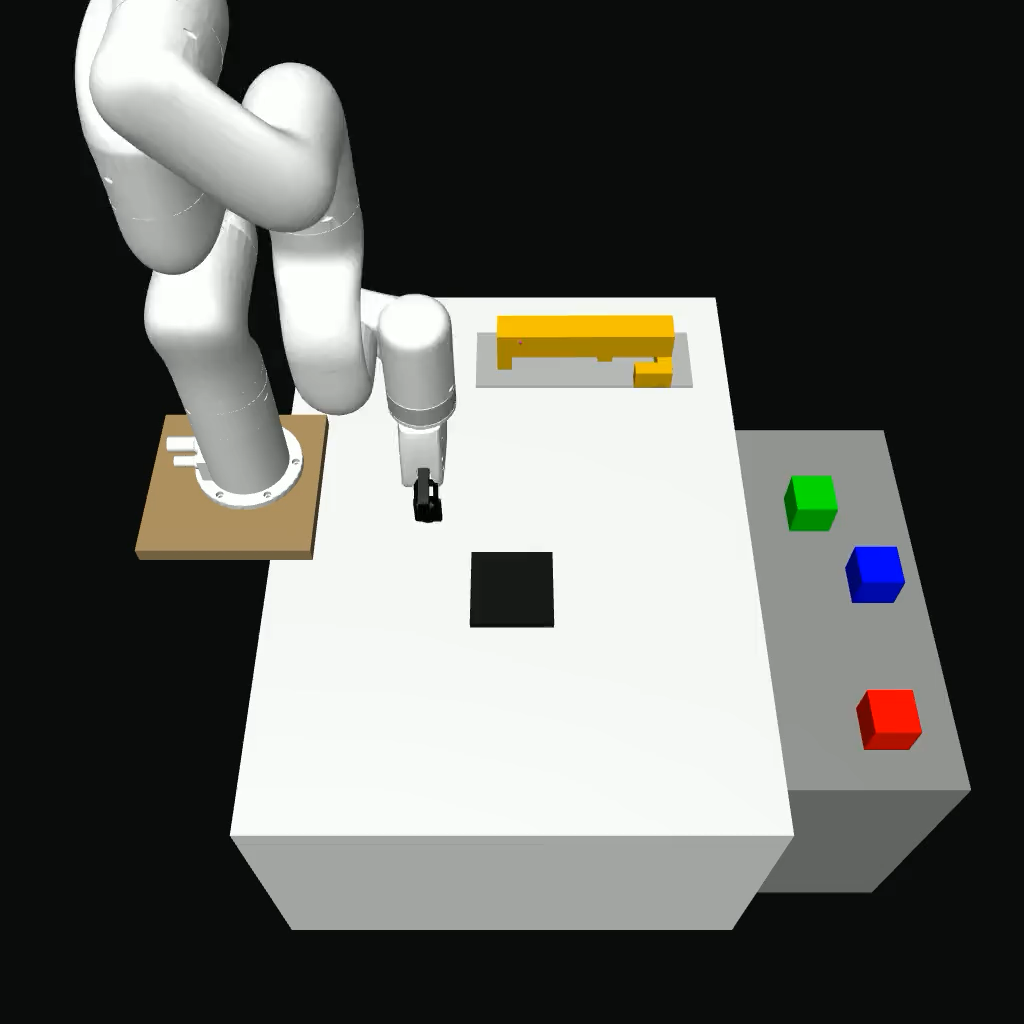}} &
            \numberedImg{2}{\includegraphics[width=0.4\linewidth, trim=210 90 60 200, clip]{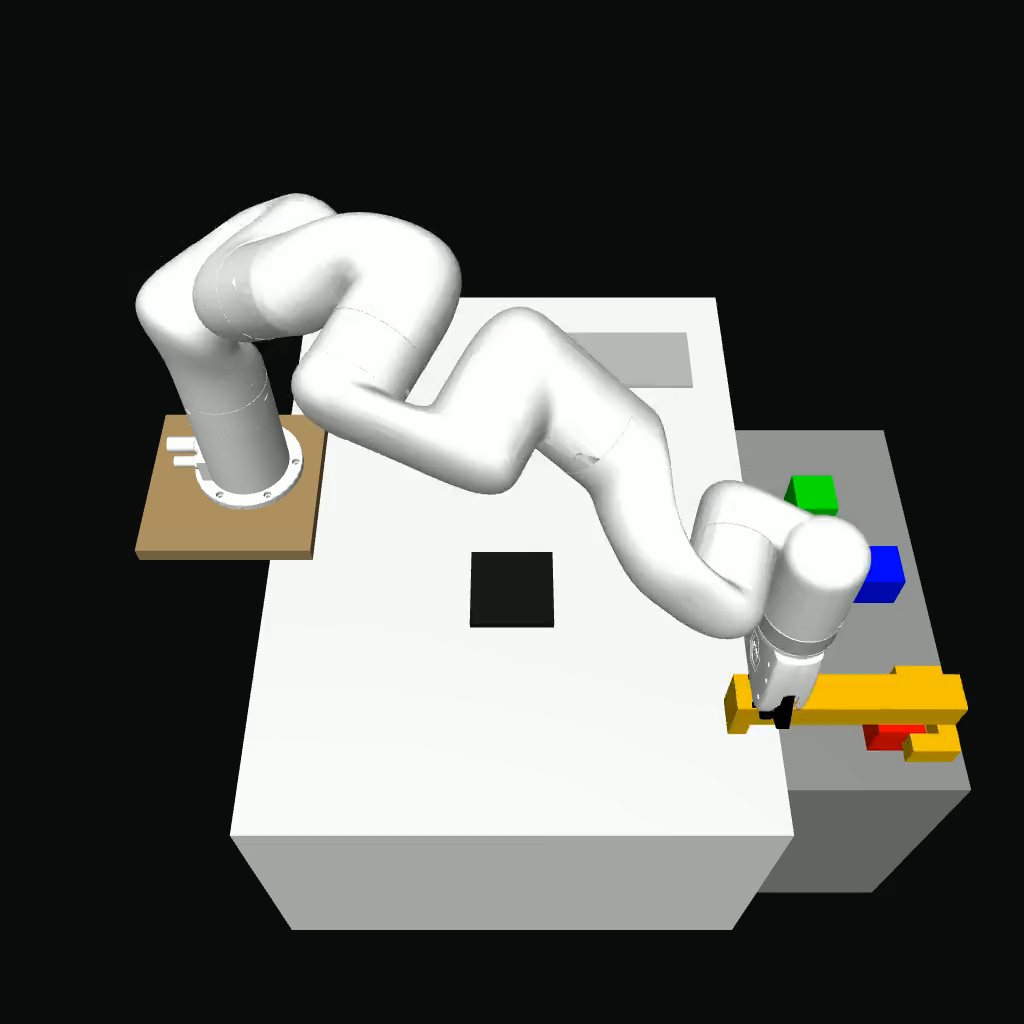}} &
            \numberedImg{3}{\includegraphics[width=0.4\linewidth, trim=210 90 60 200, clip]{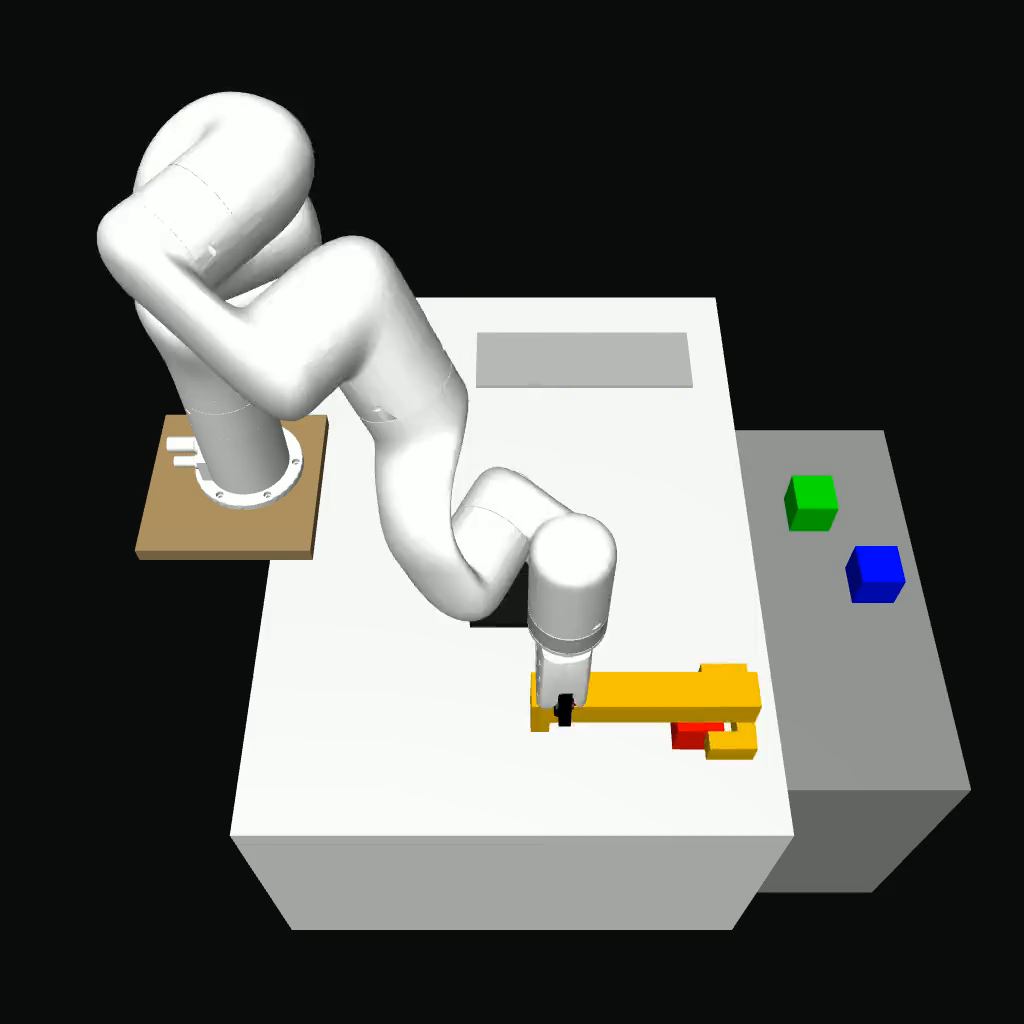}} \\
            \numberedImg{4}{\includegraphics[width=0.4\linewidth, trim=210 90 60 200, clip]{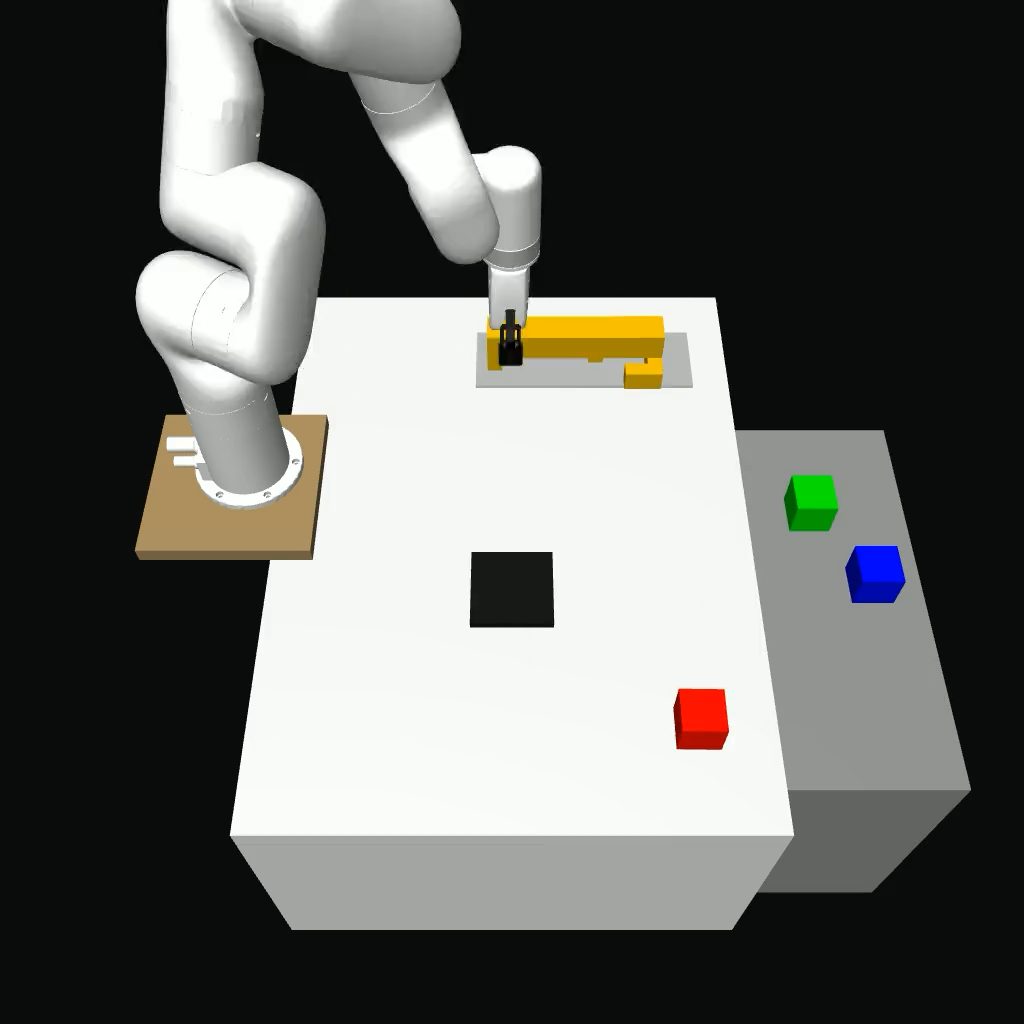}} &
            \numberedImg{5}{\includegraphics[width=0.4\linewidth, trim=210 90 60 200, clip]{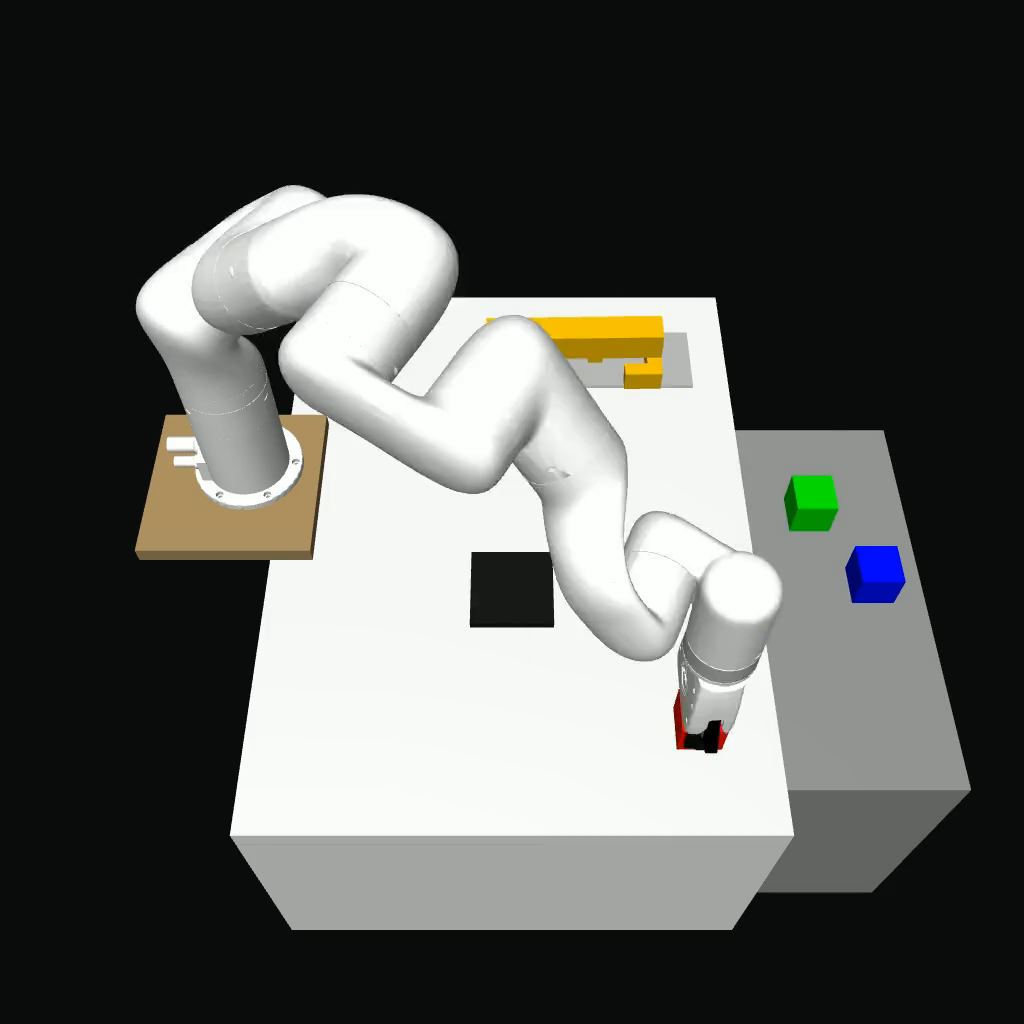}} &
            \numberedImg{6}{\includegraphics[width=0.4\linewidth, trim=210 90 60 200, clip]{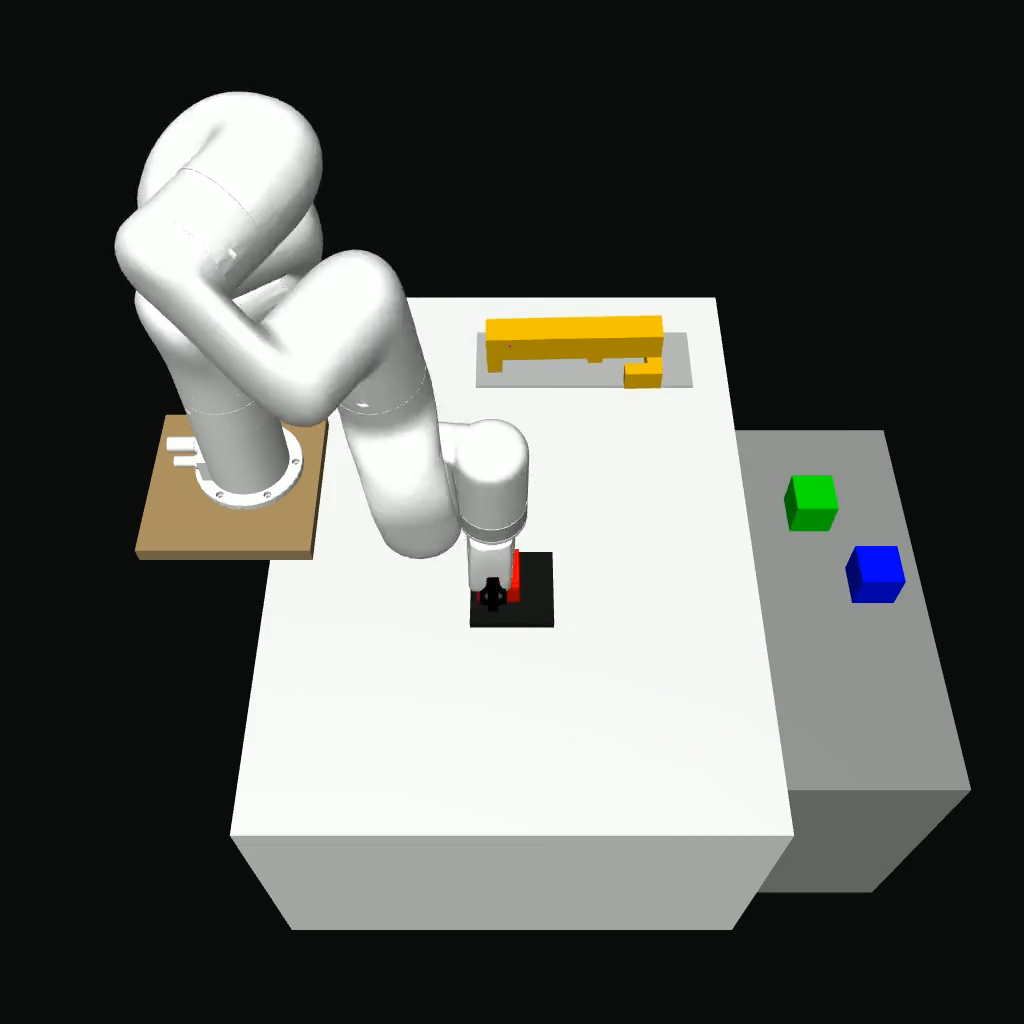}}
        \end{tabular}
        \caption{Tool-Use Task}
    \end{subfigure}

    \caption{\textbf{Robotic Manipulation Tasks.} Across benchmarks, we explicitly test the planner's ability to form long-horizon motion plans.}
    \label{fig:sim_tasks}
\end{figure*}

\paragraph{Architecture} We base our architecture on the GPT-style transformer architecture used by Chi et al.~\cite{chiDiffusionPolicyVisuomotor2023b}, where we modify the architecture to also take in the discrete plan along with its diffusion step. We embed the discrete plan using a linear layer, then concatenate it with the embedded continuous plan, and pass the result to the decoder. There, it is processed along with the encoded observations and projected back to logits with a linear layer. We also adopt the observation conditioning from Diffusion Policy, enabling the use of low-dimensional state observations or images. The architecture is illustrated in Fig. \ref{fig:architecture}. Due to the Transformer architecture, the only increase in model size is the embedding network for the discrete tokens and the diffusion step $k_d$. In our experiments, this amounts to an increase in parameter count by \(\approx 1\%\).

\section{Experimental Evaluation}
\label{sec:result}

In our experiments, we measure the performance of \model on long-horizon decision-making tasks and compare it to baselines. Firstly, we present the long-horizon robotic benchmarking suite in Subsection \ref{sec:task_suite}. We then introduce a set of baselines in Subsection \ref{sec:baselines} and show in Section \ref{sec:results} that \model excels in complex long-horizon planning, adheres to fine-grained conditioning at test time, and adapts easily to successful image conditioning. In Section \ref{sec:real}, we demonstrate the benefits of using \model on real robotic tasks.

\subsection{Simulated Task Suite}\label{sec:task_suite}

To assess the capabilities of each method, we develop a set of simulated robotic tasks to benchmark planners for long-horizon tasks that require both precision and complex decision-making. To benchmark planning performance, we form the tasks as open-loop planning problems, giving only the initial observation of the environment, and roll out the generated plan in full before checking for successful completion. The task suite stands in contrast to existing benchmarks for robotic imitation learning~\cite{chiDiffusionPolicyVisuomotor2023b}, which often focus on evaluating policies for relatively short-horizon tasks in a closed-loop environment. Benchmarks focusing on sequential task execution, such as CALVIN~\cite{meesCALVINBenchmarkLanguageConditioned2022} and Franka Kitchen~\cite{gupta_relay_2019}, test whether the policy can perform an arbitrary sequence of skills. However, in this setup, the policy is given a predetermined task sequence, which differs from our setup, where the planner itself must infer a sequence of actions to reach a goal state. All tasks utilize an X-Arm robotic manipulator, and the planner is tasked with controlling the end-effector position and gripper opening. See Fig. \ref{fig:sim_tasks} for an illustration of the tasks.

\paragraph{X-Arm Sorting} Blocks are initialized in slots on the table in random order, and the task is completed when all the blocks are sorted in different slots. The reward is based on the number of blocks in their correct place, with 100\% indicating a fully sorted state. 
We collect 200 demonstrations using a scripted planner based on an in-place sorting algorithm, Cycle sort, where the demonstrator swaps two blocks using an auxiliary slot as temporary storage. We concurrently log the corresponding discrete sequences using a vocabulary consisting of \textit{Block Identifiers:} \( \{A, B, C\} \), \textit{Actions:} \( \{\text{Pick up}, \text{Place}\} \), and \textit{Slot IDs:} \( \{\text{Slot }1,\dots \text{Slot }3\}\). In addition to the challenge of determining a valid symbolic plan, the initial position of the blocks is randomized within each slot, and models must further construct a kinematically feasible continuous plan. 


\paragraph{Arrange Blocks} Three blocks are placed randomly anywhere at the table, and the planner is tasked to place the blocks in each slot. The demonstrations are collected using a scripted planner that randomly matches blocks to slots, creating highly multimodal demonstrations. Full reward is given when all slots are filled with a block.

\paragraph{Tool-use experiment} To further illustrate the multi-step performance of \model, we construct a task where the robot must use a ``hook'' tool to drag blocks into its workspace, and then stack them. This task extends beyond simple pick-and-place operations, as the robot must utilize the tool to pull the blocks into its workspace. The scripted expert exhibits multimodal behaviour by randomly selecting to the order in which to drag and stack the blocks. The reward is proportional to the number of blocks stacked in the target area, with $\frac{1}{3}$ given for each block successfully stacked.

For all benchmarks, we report the average reward and standard deviation over three seeds over the last 10 checkpoints for each method, following Chi et al.~\cite{chiDiffusionPolicyVisuomotor2023b}. For all benchmarks, we encode the symbols as one-hot vectors.

\begin{figure*}[ht!]
    \vspace{10pt}
    \hspace{-5pt}
    \begin{minipage}[t]{0.60\textwidth}
        \centering
        \includegraphics[width=0.9\linewidth]{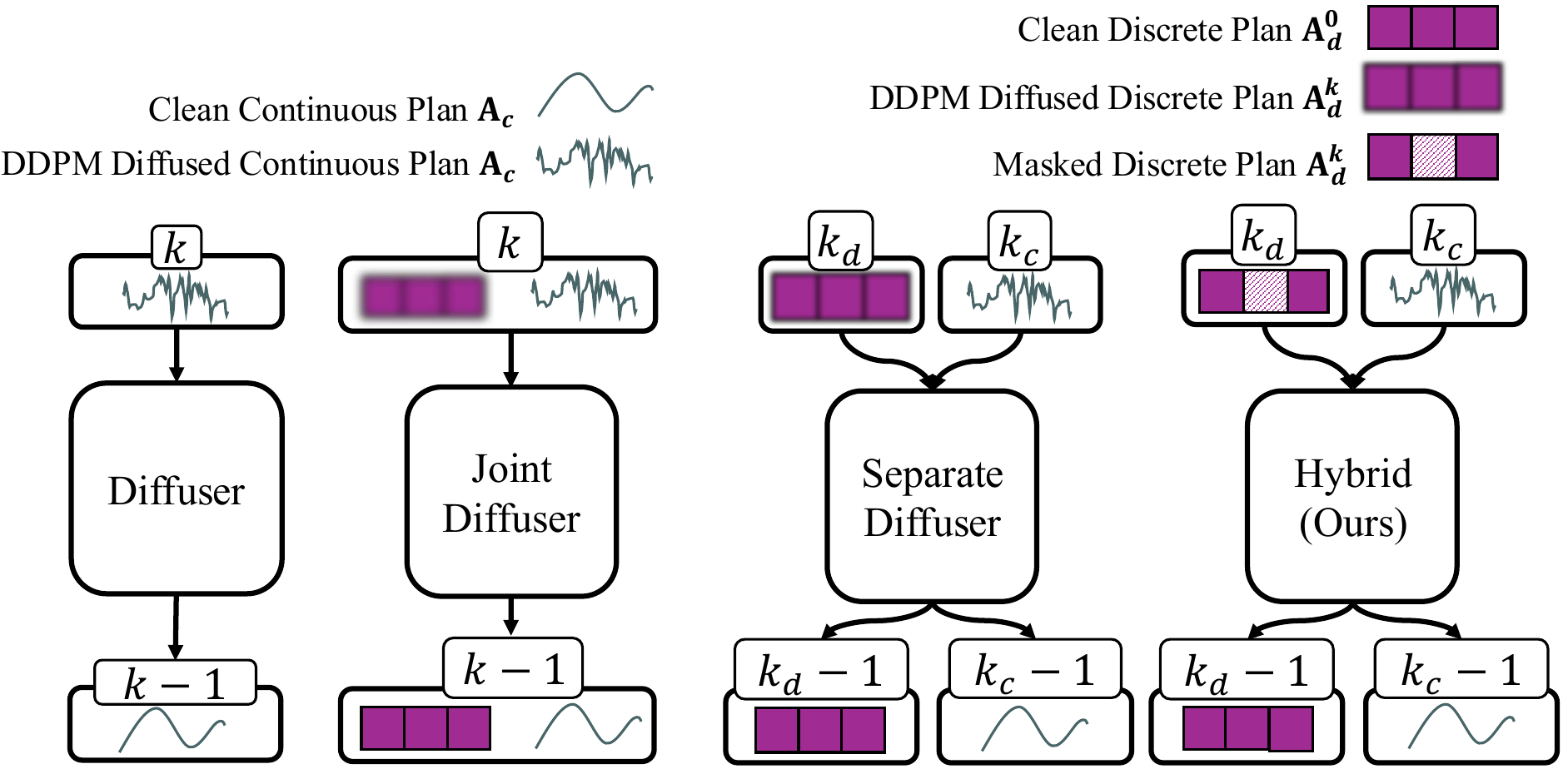}
        \caption{\textbf{Denoiser designs.} In contrast to baselines, we model the discrete and continuous plan jointly, using masked diffusion for the symbolic plan.}
        \label{fig:method_comparison}
    \end{minipage}
    \hspace{-0pt}
    \begin{minipage}[t]{0.39\textwidth}
        
        \centering
        \includegraphics[width=0.9\linewidth]{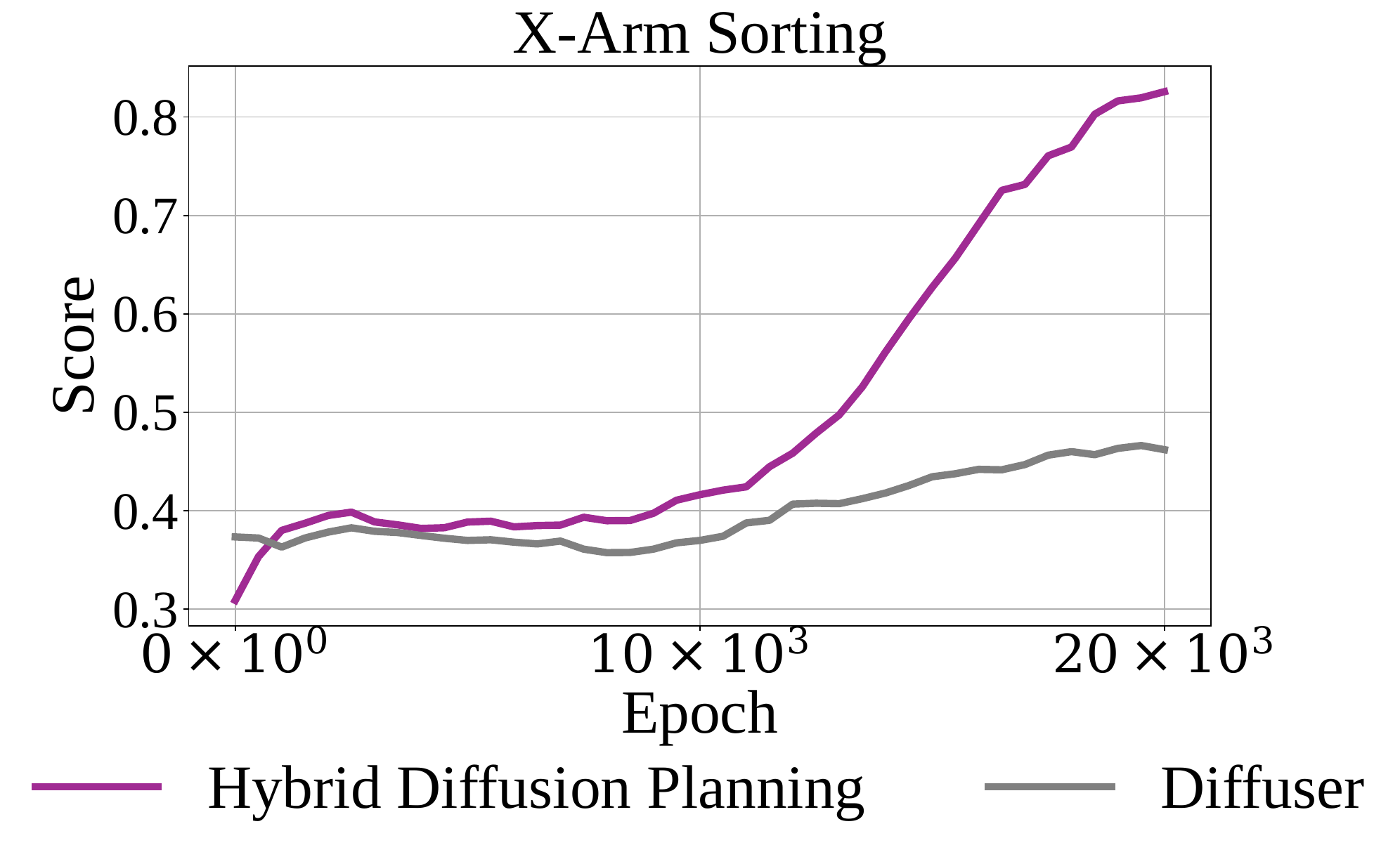}
        \caption{\textbf{Performance over time.} \model quickly learns the planning task compared to Diffuser.}
        \label{fig:score_during_training}
    \end{minipage}%
    \vspace{-15pt}
\end{figure*}

\subsection{Baselines and ablations}
\label{sec:baselines}

We benchmark our method against Diffuser~\cite{jannerPlanningDiffusionFlexible2022a}, which only models the continuous trajectory. We also compare against Hierarchical Diffuser (HD) by Chen et al.~\cite{chen2024simple} and Coupled Hierarchical Diffuser (CHD)~\cite{hao2025chd}. We modify all baselines to accept observation conditioning in the same way as \model, to be applicable to the benchmarking tasks. The Hierarchical Diffusion baselines are allowed access to the full state trajectory at training time, while the other baselines only require $\mathbf{O}$ and the symbolic and continuous actions.

To highlight the effect of key design choices of \model, we additionally form two baselines that model both plans:
\begin{itemize}[leftmargin=1em]
    \item \textbf{Joint Diffuser} Represents the symbolic plan as a continuous sequence and concatenates it to the motion plan, resulting in a single variable. These are always joined throughout the DDPM diffusion process, meaning that the architecture accepts only a single diffusion step, $k$. This represents an incremental change from the Diffuser baseline by providing the symbolic plan during training.
    \item \textbf{Separate Diffuser} Constructed to measure the effect of modeling the continuous and symbolic plan with two separate DDPM diffusion processes. This results in the symbolic and continuous plans differing in the level of noise corruption from each other during training.
\end{itemize}

\noindent An overview of all considered methods is shown in Fig.~\ref{fig:method_comparison}. All baselines use the same Transformer diffusion architecture from Chi et al.~\cite{chiDiffusionPolicyVisuomotor2023b} without causal masking, observation conditioning, and training setup as our models. Observations include the robot's initial configuration and the object poses. The planning horizons $T$ and $T_d$ are set to match the maximum plan length encountered in demonstrations.

\vspace{-5pt}
\subsection{Results}\label{sec:results}

\looseness=-1
\paragraph{\model excels at long-horizon planning} We show the results for the simulated benchmark in Table \ref{tab:sim_results}. Our method outperforms all baselines in all our experiments, achieving a relative improvement of $37\%$ over the second-best performing method, Diffuser. The training curve in Figure \ref{fig:score_during_training} shows that  \model converges significantly faster than Diffuser during training. The results also show that naively incorporating the symbolic plan by concatenation (Joint Diffuser) or a separate diffusion process (Separate Diffuser) does not result in improved performance, and we observe similar failure modes. Thus, modeling the symbolic plan through masked diffusion is critical for achieving the performance increase of \model. The underperformance of hierarchical baselines suggests that decomposing the planning horizon into segments is insufficient for capturing the complex temporal dependencies required for these tasks.

\begin{figure}
\vspace{-2pt}
\centering

\includegraphics[width=0.9\linewidth]{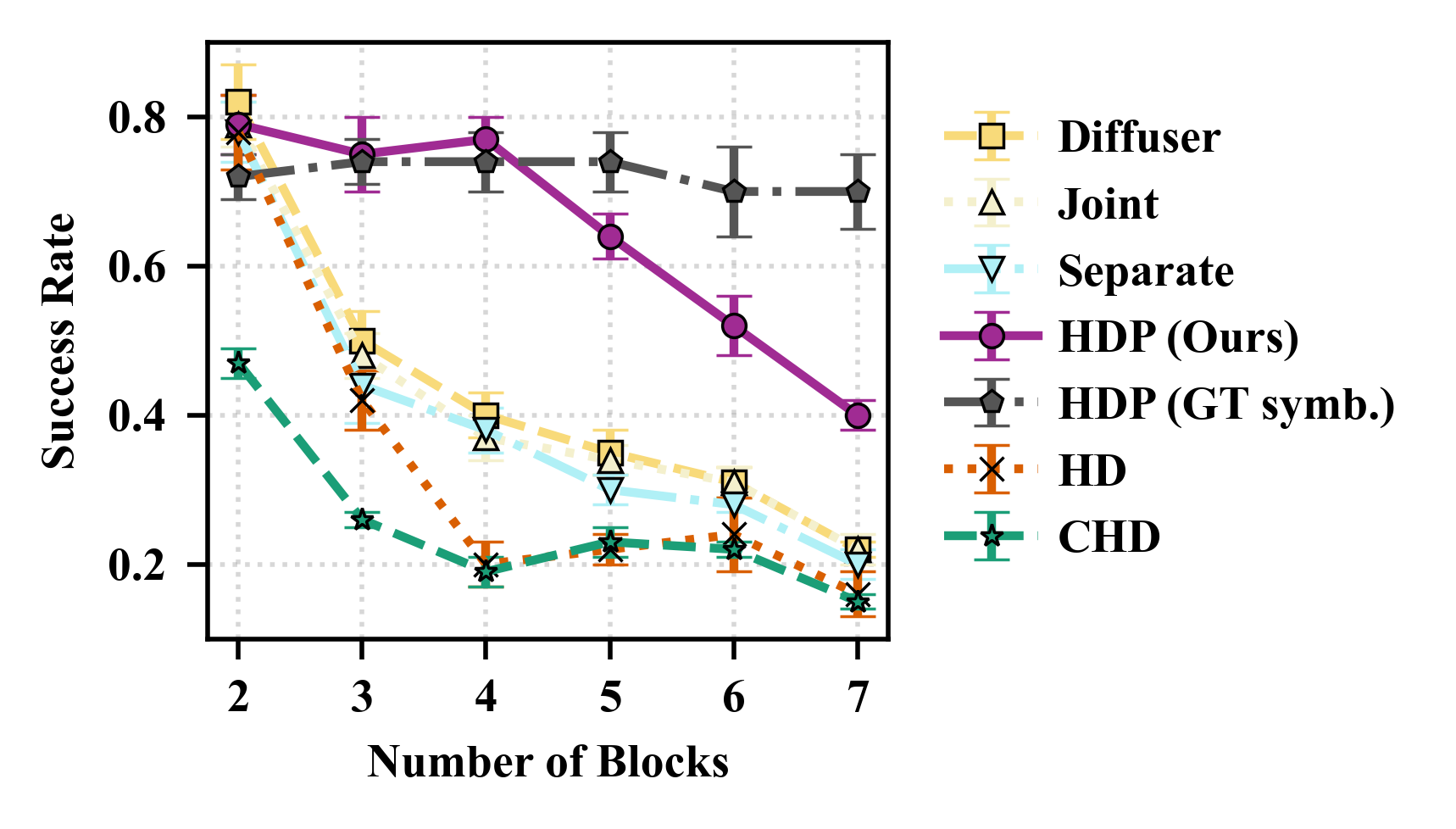}
\vspace{-9pt}
\caption{\textbf{Performance on a simulated} sorting task with increasing number of blocks. We report average results and standard deviation over 3 seeds, over the last 10 checkpoints, on 50 evaluations.}
\label{fig:increasing_num_blocks_sorting_bar_plot}
\end{figure}

\looseness=-1
We also see the highest relative performance increase ($80\%$) on the Sorting task, whereas the increase for the \textit{Arrange Blocks} task is not significant compared to the baselines. The most striking difference between these tasks is the plan temporal length $T$, which is 25 actions for \textit{Arrange Blocks} and 145 for \textit{X-Arm Sorting}. This hints that \model is significantly more robust to task complexity, and to explicitly test this robustness, we evaluate all methods on a 2D sorting task with an increasing number of blocks, using 200 demonstrations collected similarly to \textit{X-Arm Sorting}. We report the results in Figure \ref{fig:increasing_num_blocks_sorting_bar_plot}. This confirms our theory: we observe that while the baseline performance decreases monotonically with increasing task complexity, \model remains remarkably robust. In terms of prediction latency, \model has a $\sim 17\%$ increase compared to the Diffuser baseline. In comparison, Hierarchical Diffusion incurs an increase of $\sim 93\%$.

To further elucidate the working principle of HDP, we conduct experiments for providing HDP with a valid plan at inference, see Fig. \ref{fig:increasing_num_blocks_sorting_bar_plot}. We see that for higher number of blocks, this performs significantly better than unconditional HDP. This shows that, while HDP is dropping in performance due to the exponential increase of initial configurations, it still has effectively learned the coupling between the symbolic and continuous plan. This shows that HDP works because of two principles \textit{(i)} it learns the coupling between the two plan modalities, and \textit{(ii)} by generating them jointly, it is able to self-condition on a symbolic plan and avoid multimodal confusion.

\begin{table}
\centering
\begin{tabular}{ccccc}
\toprule
    Method & \makecell{X-Arm\\Sorting} & \makecell{Arrange\\Blocks} & \makecell{Tool\\Use} & Average \\
    \midrule
    Diffuser & 46\% (6\%) & 66\% (7\%) & 60\% (8\%) & 57\% \\
    Joint & 42\% (5\%) & 61\% (6\%) & 64\% (7\%) & 56\% \\
    Separate & 39\% (5\%) & 62\% (4\%) & 58\% (4\%) & 53\% \\
    HD & 34\% (5\%) & 29\% (3\%) & 4\% (1\%) & 22\% \\
    CHD & 37\% (4\%) & 23\% (2\%) & 3\% (1\%) & 21\% \\
    \model (Ours) & \textbf{86\% (5\%)} & \textbf{74\% (10\%)} & \textbf{78\% (4\%)} & \textbf{79\%} \\
    \bottomrule
\end{tabular}
\vspace{-3pt}
\caption{\textbf{Robotic simulation benchmark performance.}}
\label{tab:sim_results}
\vspace{-10pt}
\end{table}

\begin{figure}
    \centering
    \small
    \begin{tabular}{@{}p{0.25\textwidth}@{\hspace{0.01\textwidth}}p{0.2\textwidth}@{}}
    
        \begin{tabular}{cc}
        \toprule
        Method & Plan Adherence \\
        \midrule
        Diffuser Joint & $32\%$ (27\%) \\
        Diffuser Separate & $32\%$ (22\%)\\
        \model (Ours) & $\mathbf{93\%}$ \textbf{(6\%)} \\
        \bottomrule
        \end{tabular}
        &
        \hspace{3em}
        \raisebox{-0.45\height}{\centering
        \includegraphics[width=0.1\textwidth]{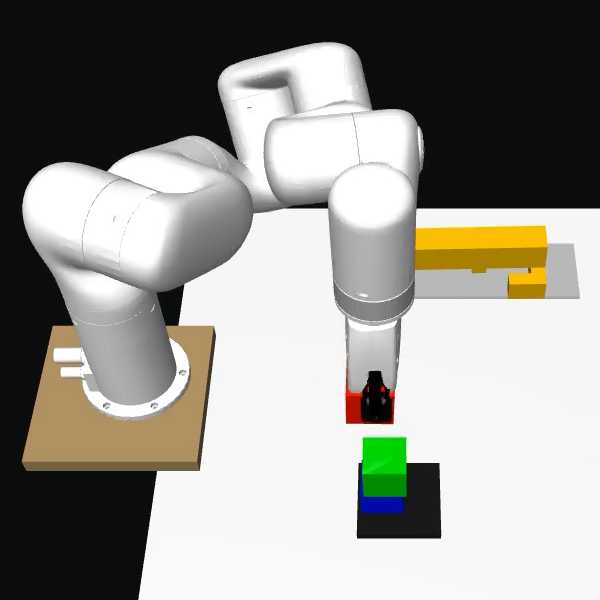}}
    \end{tabular}
    \caption{\textbf{Conditional plan generation.} We can condition \model at test-time by specifying parts of the symbolic plan, such as requiring it to place the red block on top when stacking the blocks.}
    \label{fig:red_on_top}
\end{figure}

\begin{figure*}[t]
    \centering    
    \vspace{10pt}
    \begin{minipage}{0.9\linewidth}
        \centering
        \includegraphics[width=0.15\textwidth]{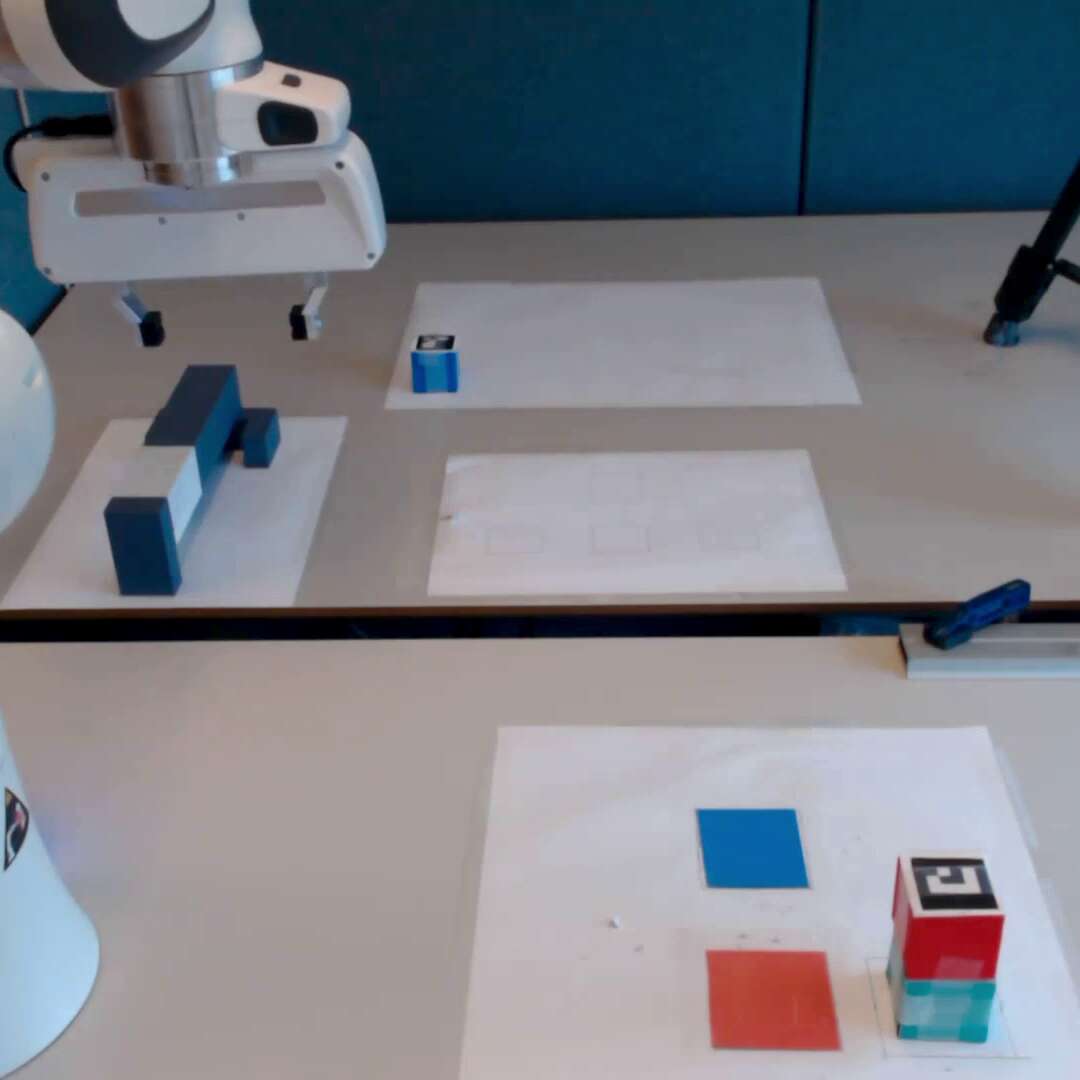}
        \hspace{-0.0em}
        \includegraphics[width=0.15\textwidth]{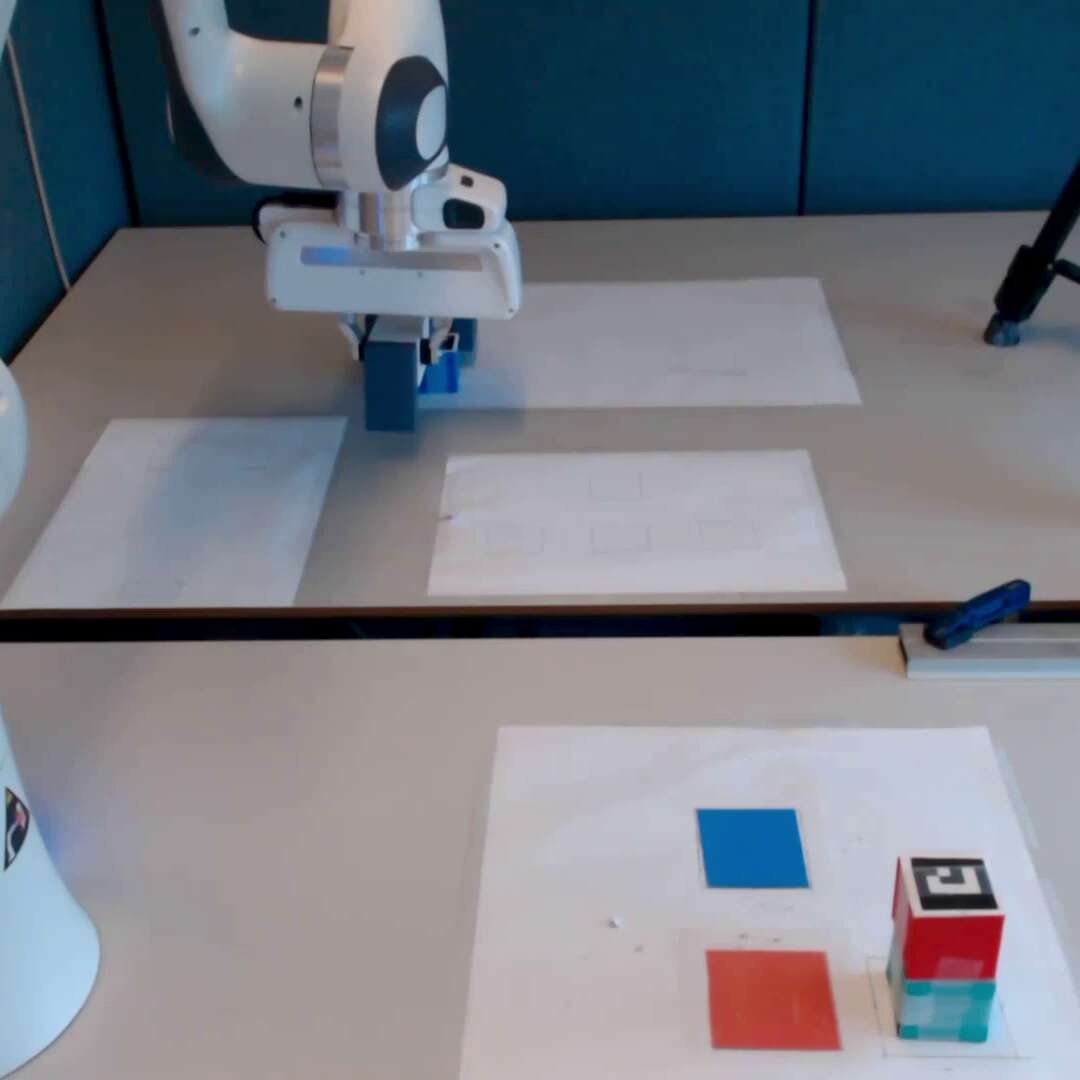}
        \hspace{0.0em}
        \includegraphics[width=0.15\textwidth]{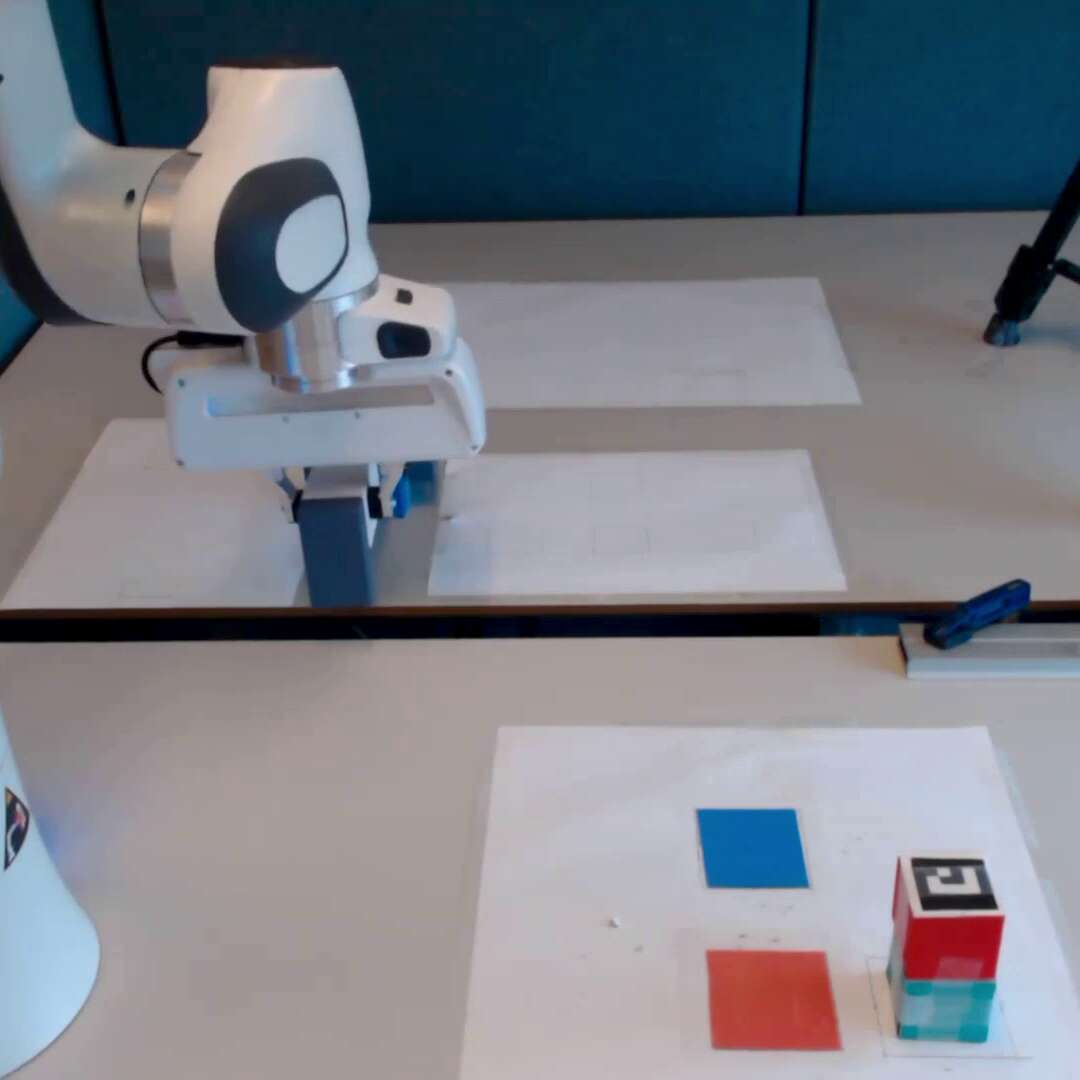}
        \hspace{0.0em}
        \includegraphics[width=0.15\textwidth]{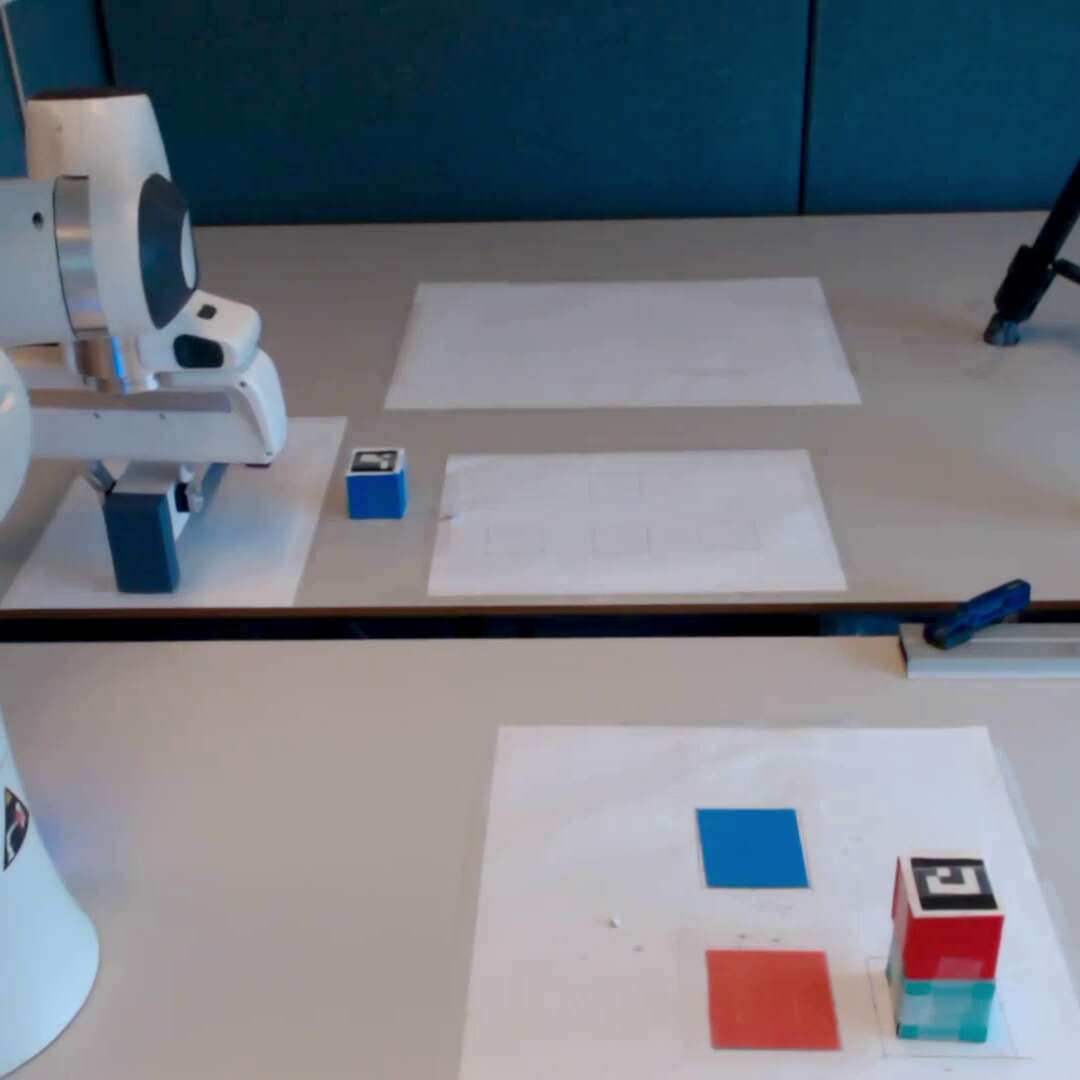}
        \hspace{0.0em}
        \includegraphics[width=0.15\textwidth]{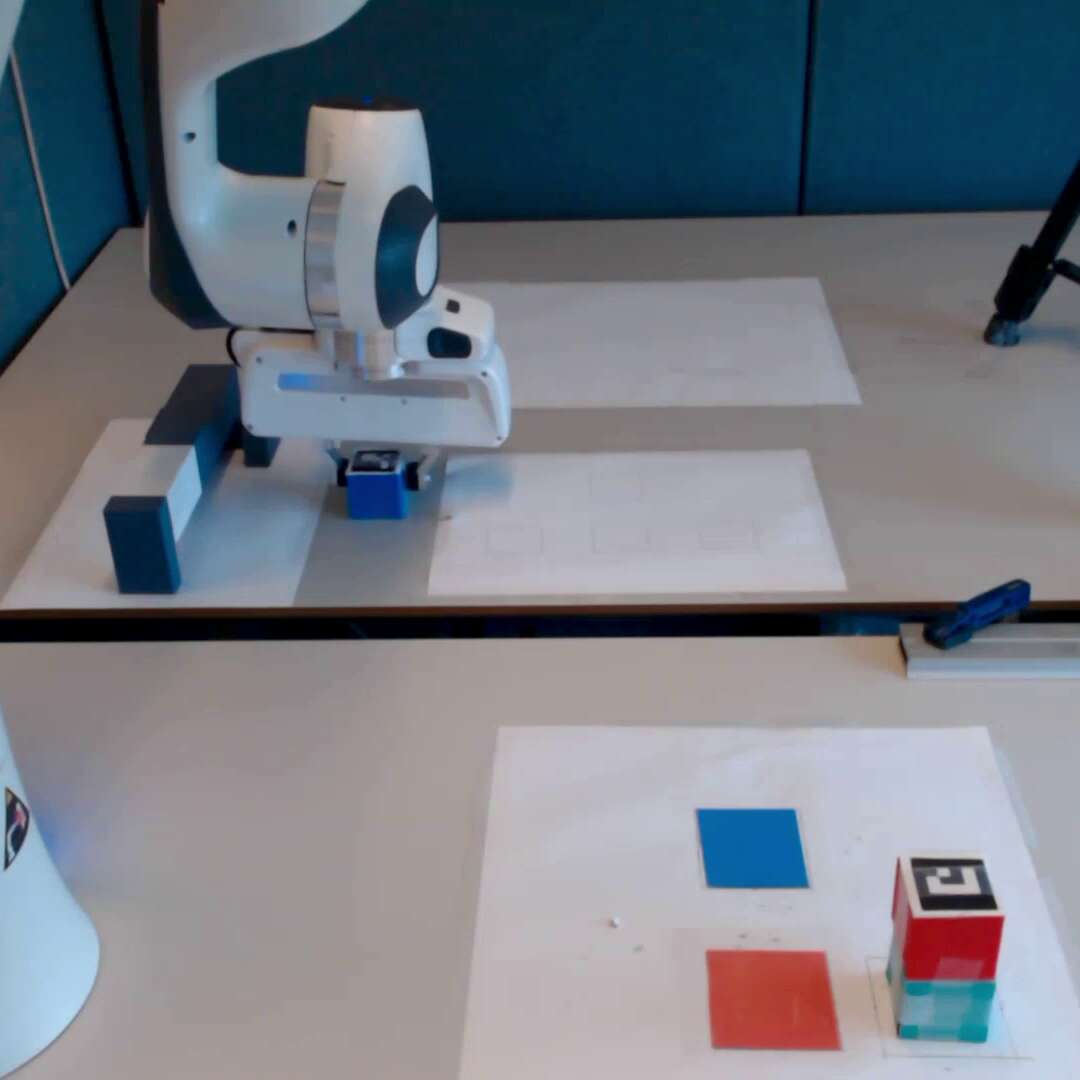}
        \hspace{0.0em}
        \includegraphics[width=0.15\textwidth]{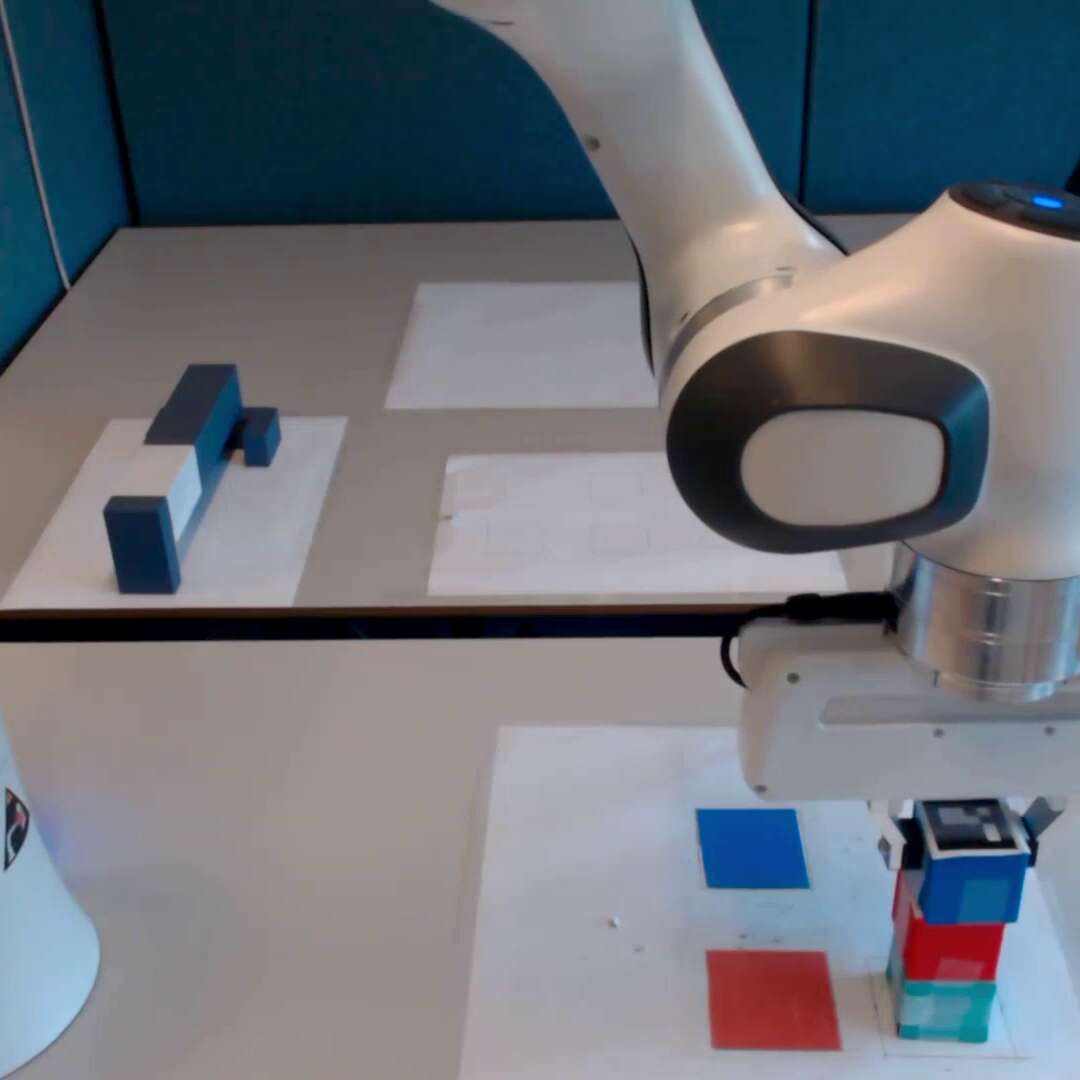}
    \end{minipage}
    
    \caption{\textbf{Real-world rollout.} Rollout sequence showing \model completing the tool use task.}
    \label{fig:real_hook_rollout}
\end{figure*}

\begin{figure}[t]
    \centering
    \vspace{-8pt}
    \begin{tabular}{@{\hspace{0.03\textwidth}}p{0.20\textwidth}@{\hspace{0.01\textwidth}}p{0.2\textwidth}@{}}
    \small
        \begin{tabular}{cc}
        \toprule
        Method & Score \\
        \midrule
        Diffuser & $62\%$ $(7\%)$ \\
        Diffuser Joint & $54\%$ $(8\%)$ \\
        Diffuser Separate & $56\%$ $(9\%)$\\
        \model (Ours) & $\mathbf{83\%}$ $\mathbf{(15 \%)}$ \\
        \bottomrule
        \end{tabular}
        &
        \hspace{2em}
        \raisebox{-0.40\height}{\centering
        \includegraphics[width=0.138\textwidth]{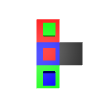}   }
    \end{tabular}
    \caption{\textbf{Image conditioning.} We can easily have the planner accept image observations \textit{(right)} by training an image encoder.}
    \label{fig:image_cond}
\end{figure}

\paragraph{\model allows for, and respects conditioning}\label{sec:cond_sampling} In addition to the dramatic performance gain of our method, the inclusion of modeling symbolic plans allows for many practical ways of conditional sampling. For example, the discrete variable diffusion process enables us to specify only parts of the symbolic plan while keeping the rest masked, as the model is trained to encounter partially masked symbolic plans during training. This allows for a more fine-grained conditioning functionality than traditional language-conditioning, which accepts only complete sentences. 

\looseness=-1
However, this conditioning is effective only if the planner respects it and exhibits high adherence to it. We therefore evaluate all methods on whether they respect the given conditioning and find that the design choices in \model are crucial for allowing successful conditioning. To systematically test the plan adherence of \model compared to baselines, we construct a variant of the \textit{Tool Use} task. For all methods, we condition on a partially unmasked symbolic sequence ending with stacking the red block as the last block. If the planner adheres to this, the red block will be on top of the tower. To explicitly check for adherence, we modify the reward function to return $1$ if all the blocks are stacked \textit{and} the red block is on the top of the others. This allows us to measure the probability of the red block being on top if all blocks are stacked, a measure we call \textit{conditional plan coherence}. For instance, using unconditional sampling, experiments show that all methods put the red on top with $30$-$35\%$ probability, given that they succeed in stacking all blocks. Figure \ref{fig:red_on_top} shows the result of this experiment. The conditional coherence for all methods shows that \model significantly outperforms the baselines when tasked to respect the conditioning. Note that Diffuser is left out since it does not allow for any symbolic conditioning. This experiment demonstrates that our design choices are crucial for respecting the conditioning with a high probability, as baselines naively incorporating the symbolic plan do not adhere to it at all, rendering conditioning ineffective.


\paragraph{\model accepts image observations} A significant benefit of diffusion-based planners is that they accept diverse observation modalities, such as images and proprioceptive sensing~\cite{chiDiffusionPolicyVisuomotor2023b}. To showcase this, we evaluate all methods on the \textit{X-Arm Sorting} task where the planner is only given access to an image observation of the initial state. To enable the models to accept image observations, we utilize a pre-trained ResNet encoder, which is further trained concurrently with the planner, following Chi et al.~\cite{chiDiffusionPolicyVisuomotor2023b}. The result of this experiment is shown in Fig.~\ref{fig:image_cond}. We observe that the robustness of \model transfer to the image-conditioning task: it outperforms the baselines in this setting, too. However, as expected, the higher-dimensional observation modality makes the problem more challenging, resulting in a performance drop across the board. 


\section{Real-world Evaluation}\label{sec:real}

\looseness=-1
In addition to our simulated experiments, we conduct an evaluation on two real-life tasks. We task the methods to solve physical versions of the \textit{Sorting} and \textit{Tool-use} tasks using a Franka Emika manipulator.

\paragraph{Real-world Sorting} As in the simulated \textit{X-arm Sorting} task, the robot is tasked to sort the blocks in-place using one auxiliary slot. A full reward is given if all blocks are sorted into their slot with the corresponding color, and none otherwise. To ensure reproducibility and fairness, we use a random generator to instruct the evaluator where to place the blocks at the beginning of each episode. The initial block permutations are generated using the same seeds for all methods, ensuring a fair comparison. 

\paragraph{Real-world Tool-Use Task} Similarly to the simulated variant, the robot is tasked to drag blocks from outside its workspace and stack them. To reach the blocks, the robot uses a 3D-printed replica of the tool to pull them into the workspace, as shown in Fig. \ref{fig:real_hook_rollout}. Similar to the sorting task, a script illustrates where to place the blocks at the beginning of the episode. The reward is calculated in the same way as for the simulated task.

\looseness=-1
Demonstrations are provided by scripted experts, similar to the simulated tasks. To provide the state observations to the planners in both tasks, the blocks are labeled with ArUco markers, and their positions are tracked using RGB-D images from a RealSense camera. For rolling out the sampled plans, we utilize the Deoxys~\cite{zhu_viola_2023} framework, which incorporates an operational-space controller, resulting in compliant behavior during collisions. We evaluate each method for 10 runs, and report the average success rate. A run will result in zero score if the operator has to intervene, which happens before collisions.

\begin{table}[t]
    \centering
    \begin{tabular}{ccc}
    \toprule
    Method & Sorting & Tool Use\\
    \midrule
    Diffuser &  $20\%$ & $6.7\%$\\
    Joint Diffuser & $10\%$ & $10\%$\\
    Separate Diffuser & $0\%$ & $6.7\%$\\
    \model (Ours)& $\mathbf{70\%}$ & $\mathbf{60\%}$ \\
    \bottomrule
    \end{tabular}
    \caption{\textbf{Real-world task performance.} Success rates of different methods averaged over 10 trials.}
    \label{tab:real_results}
\end{table}

The results are presented in Tab. \ref{tab:real_results}, and we see that \model shows superior performance in real-life tasks as well. It yields similar results to its simulation-based evaluation, while the baselines experience a significant performance drop. To a larger extent than the simulated versions, these real-life tasks require high precision.

In the Tool-use task, the most significant failure mode are collisions and misplacements, due to its long horizon and required precision. In both tasks, we see that \model exhibits more precise manipulation. In the sorting task, however, a majority of baseline failures are not due to collisions, but rather to inconsistent plans, as demonstrated in Fig. \ref{fig:teaser}. These are plans where the robot doesn't fail due to collisions or failing to grasp the blocks, but rather that the overall plan is invalid. This results in the robot manipulating the blocks, but leaving them in an unsorted state. \model, however, only fails due to imprecise motion, such as missing blocks, but the overall plan remains consistent. 

\section{Conclusion}

\label{sec:conclusion}
We present \model, a hybrid diffusion planning algorithm that learns to plan consistent, long-horizon plans and corresponding symbolic plans from demonstrations. Through simulated and real-world experiments, we demonstrate how such a system enhances long-horizon planning performance and further enables flexible and controllable planning. A limitation of the method is the need for symbolic sequences, such as text or one-hot sequences, to describe the demonstrations. While this requires additional resources, it can be efficiently provided verbally during demonstration. Incorporating full-fledged language instead of one-hot sequences is an exciting direction for future research.

\bibliography{IEEEabrv,example_compact}

\begin{thebibliography}{10}
\providecommand{\url}[1]{#1}
\csname url@rmstyle\endcsname
\providecommand{\newblock}{\relax}
\providecommand{\bibinfo}[2]{#2}
\providecommand\BIBentrySTDinterwordspacing{\spaceskip=0pt\relax}
\providecommand\BIBentryALTinterwordstretchfactor{4}
\providecommand\BIBentryALTinterwordspacing{\spaceskip=\fontdimen2\font plus
\BIBentryALTinterwordstretchfactor\fontdimen3\font minus \fontdimen4\font\relax}
\providecommand\BIBforeignlanguage[2]{{%
\expandafter\ifx\csname l@#1\endcsname\relax
\typeout{** WARNING: IEEEtran.bst: No hyphenation pattern has been}%
\typeout{** loaded for the language `#1'. Using the pattern for}%
\typeout{** the default language instead.}%
\else
\language=\csname l@#1\endcsname
\fi
#2}}

\bibitem{jannerPlanningDiffusionFlexible2022a}
M.~Janner, Y.~Du, J.~Tenenbaum, and S.~Levine, ``\BIBforeignlanguage{en}{Planning with {Diffusion} for {Flexible} {Behavior} {Synthesis}},'' in \emph{\BIBforeignlanguage{en}{Proc. of the 39th {Int.} {Conf.} on {Machine} {Learning}}}.\hskip 1em plus 0.5em minus 0.4em\relax PMLR, June 2022, pp. 9902--9915.

\bibitem{hoDenoisingDiffusionProbabilistic2020a}
J.~Ho, A.~Jain, and P.~Abbeel, ``Denoising {Diffusion} {Probabilistic} {Models},'' in \emph{Advances in {Neural} {Information} {Processing} {Systems}}, vol.~33.\hskip 1em plus 0.5em minus 0.4em\relax Curran Associates, Inc., 2020, pp. 6840--6851.

\bibitem{chiDiffusionPolicyVisuomotor2023b}
C.~Chi, S.~Feng, Y.~Du, Z.~Xu, E.~Cousineau, B.~Burchfiel, and S.~Song, ``Diffusion {Policy}: {Visuomotor} {Policy} {Learning} via {Action} {Diffusion},'' in \emph{Robotics: {Science} and {Systems} {XIX}}.\hskip 1em plus 0.5em minus 0.4em\relax Robotics: Science and Systems Foundation, July 2023.

\bibitem{ajayConditionalGenerativeModeling2023}
A.~Ajay, Y.~Du, A.~Gupta, J.~B. Tenenbaum, T.~S. Jaakkola, and P.~Agrawal, ``Is conditional generative modeling all you need for decision making?'' in \emph{The Eleventh Int. Conf. on Learning Representations}, 2023.

\bibitem{luo_potential_2024}
Y.~Luo, C.~Sun, J.~B. Tenenbaum, and Y.~Du, ``Potential based diffusion motion planning,'' in \emph{Proc. of the 41st Int. Conf. on Machine Learning}, ser. Proc. of Machine Learning Research, R.~Salakhutdinov, Z.~Kolter, K.~Heller, A.~Weller, N.~Oliver, J.~Scarlett, and F.~Berkenkamp, Eds., vol. 235.\hskip 1em plus 0.5em minus 0.4em\relax PMLR, 21--27 Jul 2024, pp. 33\,486--33\,510.

\bibitem{carvalhoMotionPlanningDiffusion2023}
J.~Carvalho, A.~Le, M.~Baierl, D.~Koert, and J.~Peters, ``Motion planning diffusion: Learning and planning of robot motions with diffusion models,'' in \emph{IEEE/RSJ Int. Conf. on Intelligent Robots and Systems (IROS)}, 2023.

\bibitem{dong_diffuserlite_2024}
Z.~Dong, J.~HAO, Y.~Yuan, F.~Ni, Y.~Wang, P.~Li, and Y.~ZHENG, ``Diffuserlite: Towards real-time diffusion planning,'' in \emph{The Thirty-eighth Annual Conf. on Neural Information Processing Systems}, 2024.

\bibitem{reussGoalConditionedImitationLearning2023a}
M.~Reuss, M.~Li, X.~Jia, and R.~Lioutikov, ``Goal-{Conditioned} {Imitation} {Learning} using {Score}-based {Diffusion} {Policies},'' in \emph{Robotics: {Science} and {Systems} {XIX}}.\hskip 1em plus 0.5em minus 0.4em\relax Robotics: Science and Systems Foundation, July 2023.

\bibitem{xian_chaineddiffuser_2023}
Z.~Xian, N.~Gkanatsios, T.~Gervet, T.-W. Ke, and K.~Fragkiadaki, ``\BIBforeignlanguage{en}{{ChainedDiffuser}: {Unifying} {Trajectory} {Diffusion} and {Keypose} {Prediction} for {Robotic} {Manipulation}},'' in \emph{\BIBforeignlanguage{en}{Proc. of {The} 7th {Conf.} on {Robot} {Learning}}}.\hskip 1em plus 0.5em minus 0.4em\relax PMLR, Dec. 2023, pp. 2323--2339.

\bibitem{mishra_generative_2023}
U.~A. Mishra, S.~Xue, Y.~Chen, and D.~Xu, ``Generative skill chaining: Long-horizon skill planning with diffusion models,'' in \emph{7th Annual Conf. on Robot Learning}, 2023.

\bibitem{luo_generative_2025}
Y.~Luo, U.~A. Mishra, Y.~Du, and D.~Xu, ``Generative {Trajectory} {Stitching} through {Diffusion} {Composition},'' Mar. 2025, arXiv:2503.05153 [cs].

\bibitem{garrett_integrated_2020}
C.~R. Garrett, R.~Chitnis, R.~Holladay, B.~Kim, T.~Silver, L.~P. Kaelbling, and T.~Lozano-P{\'e}rez, ``Integrated task and motion planning,'' \emph{Annual Review of Control, Robotics, and Autonomous Systems}, vol.~4, no. Volume 4, 2021, pp. 265--293, 2021.

\bibitem{garrett_pddlstream_2020}
C.~R. Garrett, T.~Lozano-Perez, and L.~P. Kaelbling, ``Pddlstream: Integrating symbolic planners and blackbox samplers via optimistic adaptive planning,'' in \emph{Int. Conf. on Automated Planning and Scheduling}, 2018.

\bibitem{shi_simplified_2024}
J.~Shi, K.~Han, Z.~Wang, A.~Doucet, and M.~Titsias, ``Simplified and generalized masked diffusion for discrete data,'' in \emph{The Thirty-eighth Annual Conf. on Neural Information Processing Systems}, 2024.

\bibitem{meesCALVINBenchmarkLanguageConditioned2022}
O.~Mees, L.~Hermann, E.~Rosete-Beas, and W.~Burgard, ``Calvin: A benchmark for language-conditioned policy learning for long-horizon robot manipulation tasks,'' \emph{IEEE Robotics and Automation Letters (RA-L)}, vol.~7, no.~3, pp. 7327--7334, 2022.

\bibitem{gupta_relay_2019}
A.~Gupta, V.~Kumar, C.~Lynch, S.~Levine, and K.~Hausman, ``Relay policy learning: Solving long-horizon tasks via imitation and reinforcement learning,'' in \emph{Proc. of the Conf. on Robot Learning}, ser. Proc. of Machine Learning Research, L.~P. Kaelbling, D.~Kragic, and K.~Sugiura, Eds., vol. 100.\hskip 1em plus 0.5em minus 0.4em\relax PMLR, 30 Oct--01 Nov 2020.

\bibitem{pearceImitatingHumanBehaviour2023}
T.~Pearce, T.~Rashid, A.~Kanervisto, D.~Bignell, M.~Sun, R.~Georgescu, S.~V. Macua, S.~Z. Tan, I.~Momennejad, K.~Hofmann, and S.~Devlin, ``Imitating human behaviour with diffusion models,'' in \emph{The Eleventh Int. Conf. on Learning Representations}, 2023.

\bibitem{saha_edmp_2023}
K.~Saha, V.~Mandadi, J.~Reddy, A.~Srikanth, A.~Agarwal, B.~Sen, A.~Singh, and M.~Krishna, ``Edmp: Ensemble-of-costs-guided diffusion for motion planning,'' in \emph{2024 IEEE Int. Conf. on Robotics and Automation (ICRA)}, 2024, pp. 10\,351--10\,358.

\bibitem{ubukata_diffusion_2024}
T.~Ubukata, J.~Li, and K.~Tei, ``Diffusion {Model} for {Planning}: {A} {Systematic} {Literature} {Review},'' Aug. 2024, arXiv:2408.10266.

\bibitem{pmlr-v202-li23ad}
W.~Li, X.~Wang, B.~Jin, and H.~Zha, ``Hierarchical diffusion for offline decision making,'' in \emph{Proc. of the 40th Int. Conf. on Machine Learning}, ser. Proc. of Machine Learning Research, vol. 202.\hskip 1em plus 0.5em minus 0.4em\relax PMLR, 23--29 Jul 2023, pp. 20\,035--20\,064.

\bibitem{chen2024simple}
C.~Chen, F.~Deng, K.~Kawaguchi, C.~Gulcehre, and S.~Ahn, ``Simple hierarchical planning with diffusion,'' in \emph{Int. Conf. on Representation Learning}, 2024.

\bibitem{hao2025chd}
C.~Hao, A.~Xiao, Z.~Xue, and H.~Soh, ``{CHD}: Coupled hierarchical diffusion for long-horizon tasks,'' in \emph{9th Annual Conference on Robot Learning}, 2025.

\bibitem{garrett_online_2020}
C.~R. Garrett, C.~Paxton, T.~Lozano-Perez, L.~P. Kaelbling, and D.~Fox, ``\BIBforeignlanguage{en}{Online {Replanning} in {Belief} {Space} for {Partially} {Observable} {Task} and {Motion} {Problems}},'' in \emph{\BIBforeignlanguage{en}{2020 {IEEE} {Int.} {Conf.} on {Robotics} and {Automation} ({ICRA})}}.\hskip 1em plus 0.5em minus 0.4em\relax Paris, France: IEEE, May 2020, pp. 5678--5684.

\bibitem{fang_dimsam_2024}
X.~Fang, C.~R. Garrett, C.~Eppner, T.~Lozano-Pérez, L.~P. Kaelbling, and D.~Fox, ``{DiMSam}: {Diffusion} {Models} as {Samplers} for {Task} and {Motion} {Planning} under {Partial} {Observability},'' in \emph{2024 {IEEE}/{RSJ} {Int.} {Conf.} on {Intelligent} {Robots} and {Systems} ({IROS})}, Oct. 2024, pp. 1412--1419.

\bibitem{mandlekar_human---loop_2023}
A.~Mandlekar, C.~R. Garrett, D.~Xu, and D.~Fox, ``Human-in-the-loop task and motion planning for imitation learning,'' in \emph{Proc. of The 7th Conf. on Robot Learning}, ser. Proc. of Machine Learning Research, J.~Tan, M.~Toussaint, and K.~Darvish, Eds., vol. 229.\hskip 1em plus 0.5em minus 0.4em\relax PMLR, 06--09 Nov 2023, pp. 3030--3060.

\bibitem{zhou_spire_2024}
Z.~Zhou, A.~Garg, D.~Fox, C.~R. Garrett, and A.~Mandlekar, ``Spire: Synergistic planning, imitation, and reinforcement learning for long-horizon manipulation,'' in \emph{Proc. of The 8th Conf. on Robot Learning}, ser. Proceedings of Machine Learning Research, P.~Agrawal, O.~Kroemer, and W.~Burgard, Eds., vol. 270.\hskip 1em plus 0.5em minus 0.4em\relax PMLR, 06--09 Nov 2025, pp. 2347--2371.

\bibitem{silver_predicate_2025}
T.~Silver, R.~Chitnis, N.~Kumar, W.~McClinton, T.~Lozano-Pérez, L.~Kaelbling, and J.~B. Tenenbaum, ``Predicate invention for bilevel planning,'' \emph{Proc. of the AAAI Conf. on Artificial Intelligence}, vol.~37, no.~10, pp. 12\,120--12\,129, Jun. 2023.

\bibitem{shah_real_2025}
N.~Shah, J.~Nagpal, and S.~Srivastava, ``From {Real} {World} to {Logic} and {Back}: {Learning} {Generalizable} {Relational} {Concepts} {For} {Long} {Horizon} {Robot} {Planning},'' June 2025, arXiv:2402.11871 [cs].

\bibitem{zengTransporterNetworksRearranging2022}
A.~Zeng, P.~Florence, J.~Tompson, S.~Welker, J.~Chien, M.~Attarian, T.~Armstrong, I.~Krasin, D.~Duong, V.~Sindhwani, and J.~Lee, ``Transporter networks: Rearranging the visual world for robotic manipulation,'' in \emph{Proc. of the 2020 Conf. on Robot Learning}, ser. Proc. of Machine Learning Research, J.~Kober, F.~Ramos, and C.~Tomlin, Eds., vol. 155.\hskip 1em plus 0.5em minus 0.4em\relax PMLR, 16--18 Nov 2021, pp. 726--747.

\bibitem{furrutter2024quantum}
F.~F{\"u}rrutter, G.~Mu{\~n}oz-Gil, and H.~J. Briegel, ``Quantum circuit synthesis with diffusion models,'' \emph{Nature Machine Intelligence}, vol.~6, no.~5, pp. 515--524, 2024.

\bibitem{chen2024diffusion}
B.~Chen, D.~Mart{\'\i}~Mons{\'o}, Y.~Du, M.~Simchowitz, R.~Tedrake, and V.~Sitzmann, ``Diffusion forcing: Next-token prediction meets full-sequence diffusion,'' \emph{Advances in Neural Information Processing Systems}, vol.~37, pp. 24\,081--24\,125, 2024.

\bibitem{zhu_viola_2023}
Y.~Zhu, A.~Joshi, P.~Stone, and Y.~Zhu, ``\BIBforeignlanguage{en}{{VIOLA}: {Imitation} {Learning} for {Vision}-{Based} {Manipulation} with {Object} {Proposal} {Priors}},'' in \emph{\BIBforeignlanguage{en}{Proc. of {The} 6th {Conf.} on {Robot} {Learning}}}.\hskip 1em plus 0.5em minus 0.4em\relax PMLR, Mar. 2023, pp. 1199--1210.

\bibitem{kingmaVariationalDiffusionModels2023}
D.~P. Kingma, T.~Salimans, B.~Poole, and J.~Ho, ``Variational diffusion models,'' in \emph{Advances in Neural Information Processing Systems}, A.~Beygelzimer, Y.~Dauphin, P.~Liang, and J.~W. Vaughan, Eds., 2021.

\bibitem{chang_maskgit_2022}
H.~Chang, H.~Zhang, L.~Jiang, C.~Liu, and W.~T. Freeman, ``Maskgit: Masked generative image transformer,'' in \emph{The IEEE Conf. on Computer Vision and Pattern Recognition (CVPR)}, June 2022.

\end{thebibliography}

\clearpage

\begin{appendices}
\section{Experimental details}

\subsection{Training and Evaluation Details}
For the simulated results, we follow the evaluation procedure of  Chi et al. \cite{chiDiffusionPolicyVisuomotor2023b}, and train the models for three seeds in parallel. They are evaluated with 50 environment initializations at regular intervals during training, and the average score is calculated over the last 10 checkpoints over the three seeds, effectively capturing performance over 1500 initializations. Training is done using NVIDIA A100 GPUs, with training times ranging from 30 minutes for \textit{Real-world Sorting} to 15 hours for \textit{Rearrange Blocks}.

We provide hyperparameters for each experimental setup in Table \ref{tab:hyperparams_all} and Table \ref{tab:hyperparams_img}. We use a batch size of 64 for all experiments, and set the architecture with hyperparameters in Table \ref{tab:transformer_arch}.

\begin{table}
    \centering
    \setlength{\tabcolsep}{4pt} 
    \begin{tabular}{lcccccc}
        \toprule
        Experiment & Epochs & \# Demos & $H_c$ & $H_d$ & $D_{ac}$ & Vocab. Size \\
        \midrule
        X-Arm Sorting & 5000 & 200 & 145 & 108 & 4 & 10 \\
        Arrange Blocks & 5000 & 4000 & 25 & 15 & 4 & 9 \\
        Tool-Use & 10000 & 4000 & 61 & 15 & 4 & 6 \\
        Real-world Sorting & 5000 & 200 & 145 & 108 & 4 & 10 \\
        2 Block Sorting & 6000 & 200 & 73 & 72 & 3 & 8 \\
        3 Block Sorting & 6000 & 200 & 109 & 108 & 3 & 10 \\
        4 Block Sorting & 6000 & 200 & 145 & 144 & 3 & 12 \\
        \bottomrule
    \end{tabular}
    \vspace{1em}
    \caption{\textbf{Hyperparameters for all tasks.} $H_c$ and $H_d$ are the continuous and discrete planning horizons, respectively. $D_{ac}$ denotes the dimensionality of the continuous action space.}
    \label{tab:hyperparams_all}
\end{table}

\begin{table}
    \centering
    \setlength{\tabcolsep}{4pt} 
    \begin{tabular}{lcccccc}
        \toprule
        Experiment & Epochs & \# Demos & $H_c$ & $H_d$ & $D_{ac}$ & Vocab. Size \\
        \midrule
        X-Arm Sorting & 20000 & 200 & 145 & 108 & 4 & 10 \\
        \bottomrule
    \end{tabular}
    \vspace{1em}
    \caption{\textbf{Hyperparameters for Image-conditioned tasks.}}
    \label{tab:hyperparams_img}
\end{table}

\subsection{Architecture hyperparameters}
See the Diffusion Policy codebase~\cite{chiDiffusionPolicyVisuomotor2023b} for details on the original architecture. The discrete diffusion level $k_d$ is processed with a learned embedding consisting of two linear layers with a Mish non-linearity before concatenating with the embedding of continuous variable diffusion step $k_c$ and observation $\mathbf{O}$.

\begin{table}[h!]
    \centering
    \begin{tabular}{lc}
        \toprule
        Parameter & Value \\
        \midrule
        Num layers & 8 \\
        Num heads & 4 \\
        Emb. dim. & 256 \\
        Drop emb. prob. & 0.0 \\
        Drop atten. prob. & 0.3 \\
        Causal Attention & Disabled \\
        \bottomrule
    \end{tabular}
    \caption{\textbf{Transformer architecture hyperparameters.}}
    \label{tab:transformer_arch}
\end{table}

\section{Algorithmic Details}

This section outlines details on combining MD4~\cite{shi_simplified_2024} with DDPM~\cite{hoDenoisingDiffusionProbabilistic2020a} during training and sampling. 

During training, the diffusion steps for the discrete variable diffusion are sampled from a continuous uniform distribution $k_d \sim \mathcal{U}[0,1]$. We sample the continuously distributed diffusion step $k_d$ with the low-discrepancy sampling~\cite{kingmaVariationalDiffusionModels2023} following Shi et al. \cite{shi_simplified_2024}. For the continuous variable diffusion, DDPM expects a sample from a categorical distribution over the set of all training levels $k_c \sim \mathcal{U}\{0, \dots, N-1\}$.

At sampling time, the continuous diffusion iterates through the diffusion levels $\{N-1, \dots, 0\}$, while the discrete variable diffusion applies a cosine masking schedule~\cite{chang_maskgit_2022}. The variable $i$ iterates from $0$ to $N-1$, and 
\begin{align*}
t &= \cos\left( \frac{\pi i}{2N} \right)\\
s &
   = \cos\left( \frac{\pi (i + 1)}{2N} \right),
\end{align*}
where $t$ is passed through the model (as $k_d$), and both $s$ and $t$ are used for denoising the sample. See Shi et al. \cite{shi_simplified_2024} for further details on MD4 sampling.

\section{Rollout visualizations}
Figure \ref{fig:real_rollouts} shows the rollout of all methods on the \textit{Real-world Sorting} task, all with the same permutation. All methods except Separate Diffuser plan a non-collision sequence. However, \model is the only method that solves the task.

\newcommand{\imgwidth}{0.144\textwidth}

\begin{figure*}
    \centering

    \textbf{Hybrid} \\
    \begin{tabular}{@{}c@{}c@{}c@{}c@{}c@{}c@{}c@{}}
        \includegraphics[width=\imgwidth]{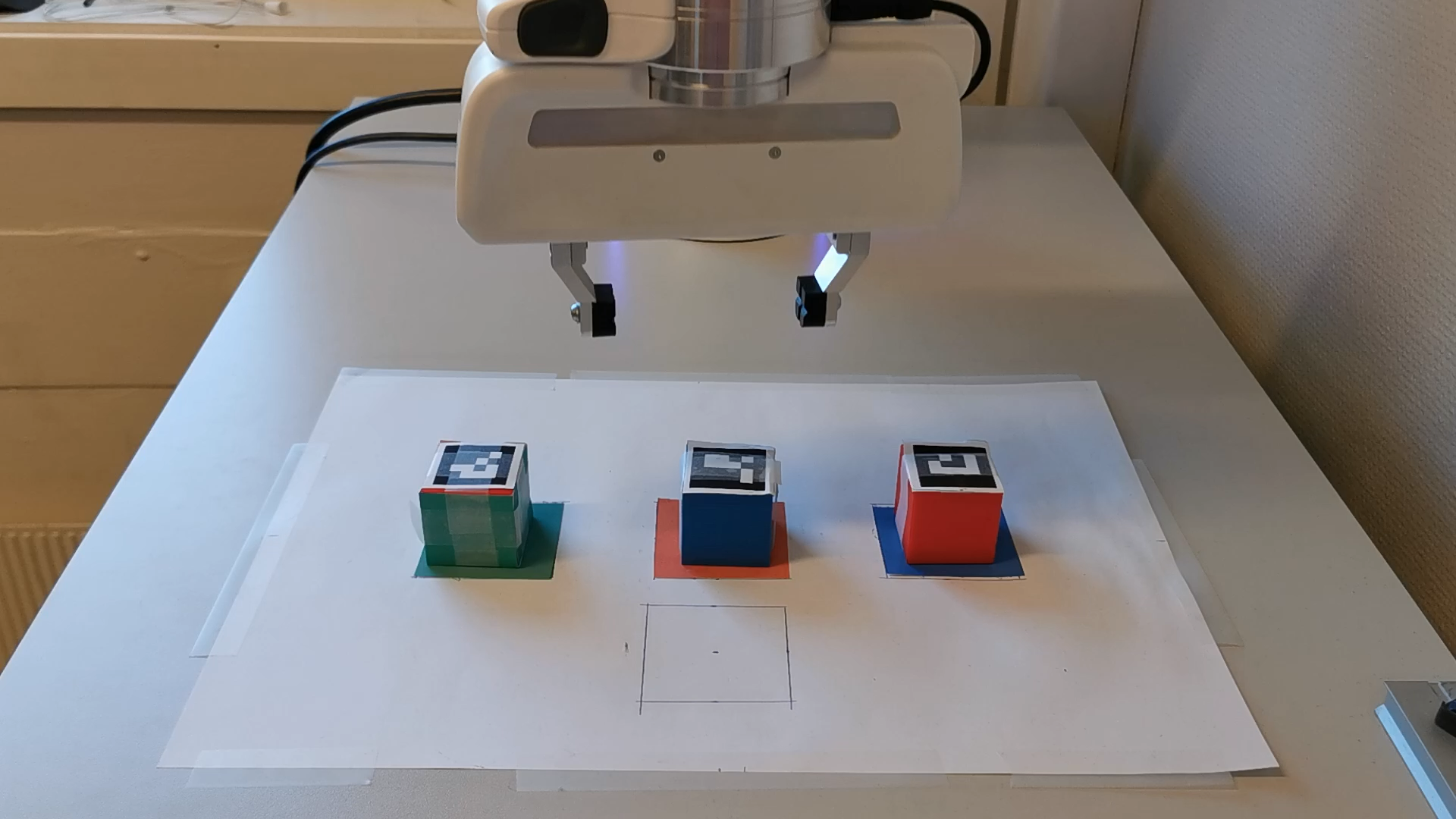} &
        \includegraphics[width=\imgwidth]{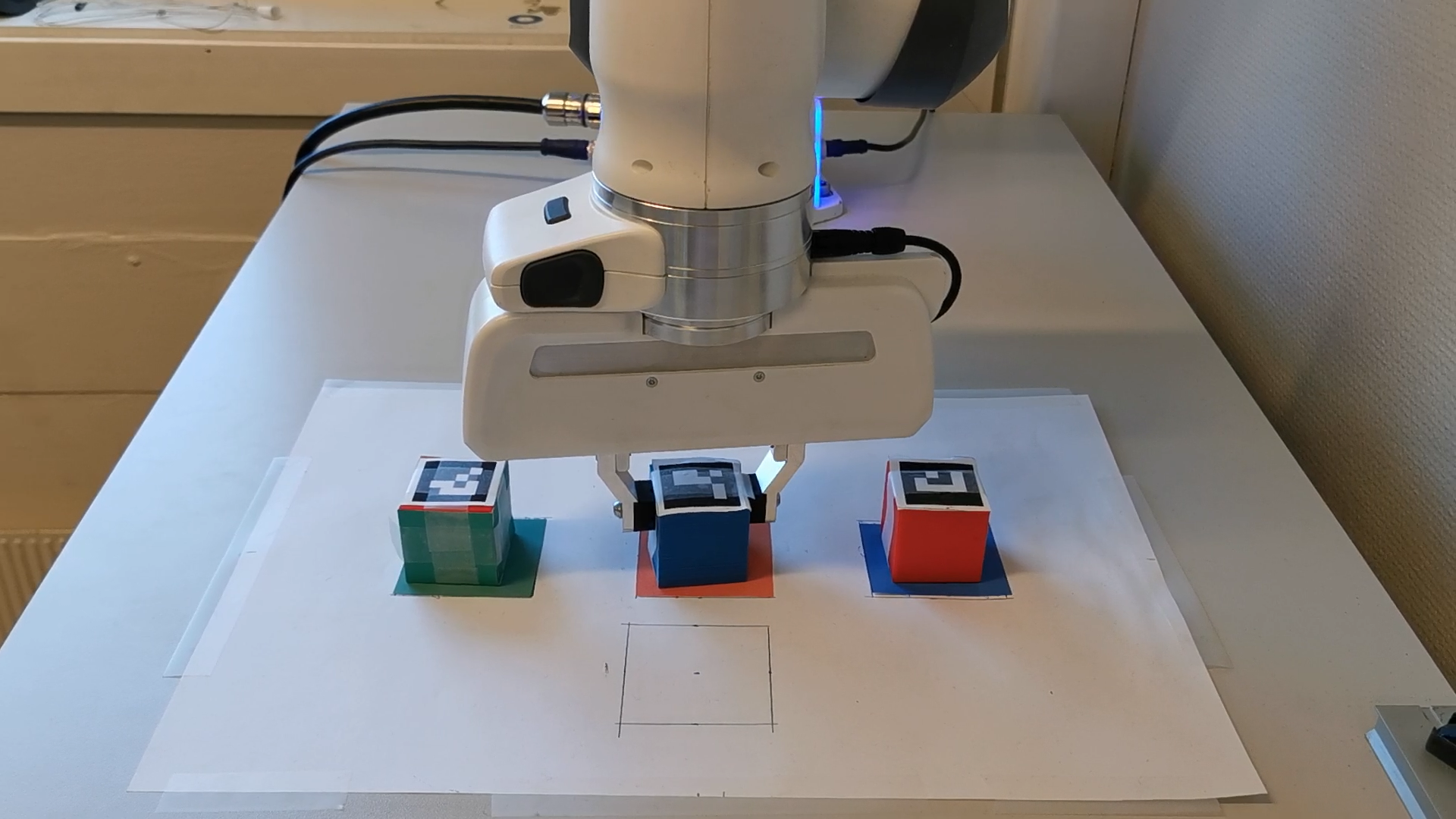} &
        \includegraphics[width=\imgwidth]{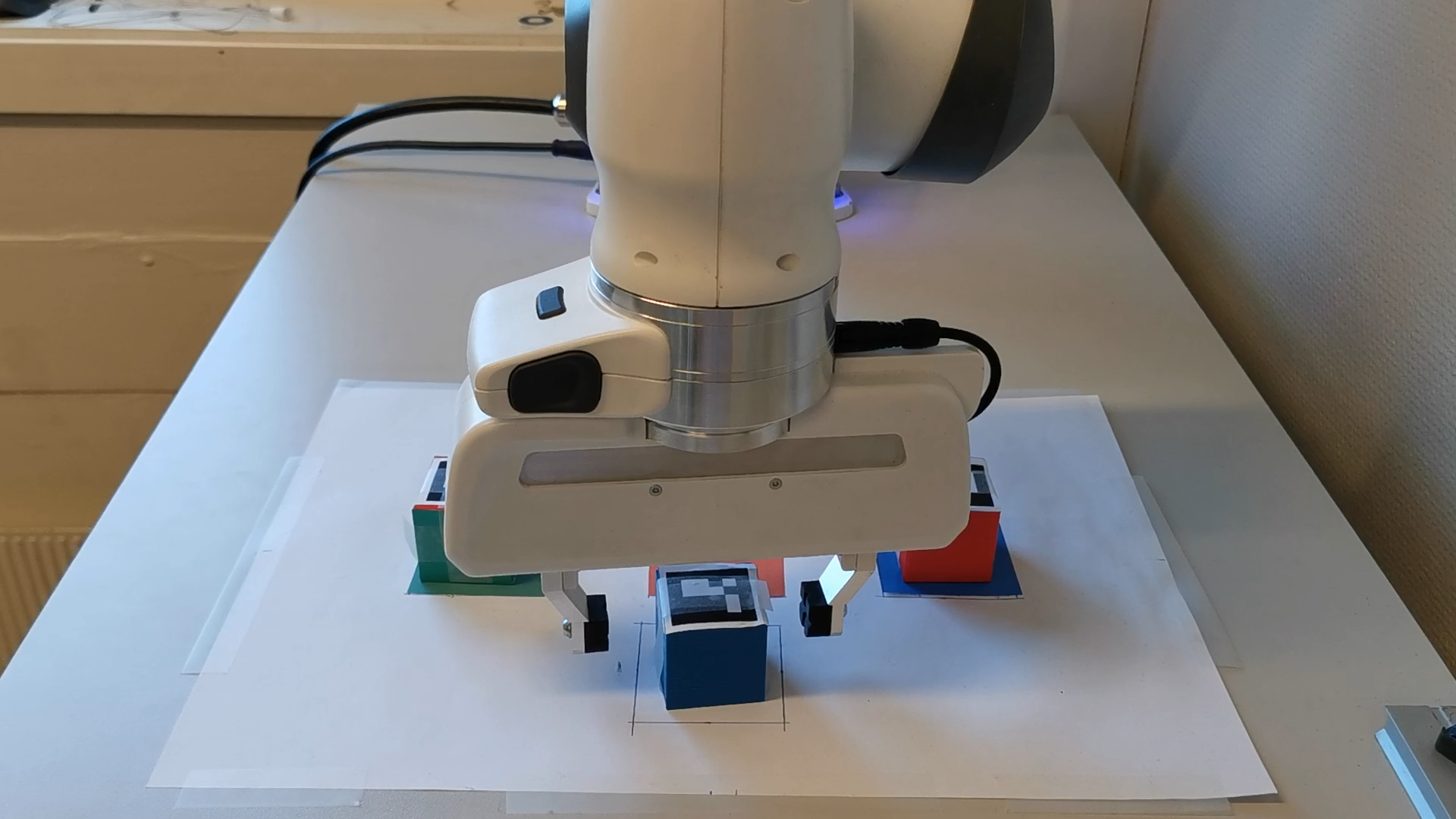} &
        \includegraphics[width=\imgwidth]{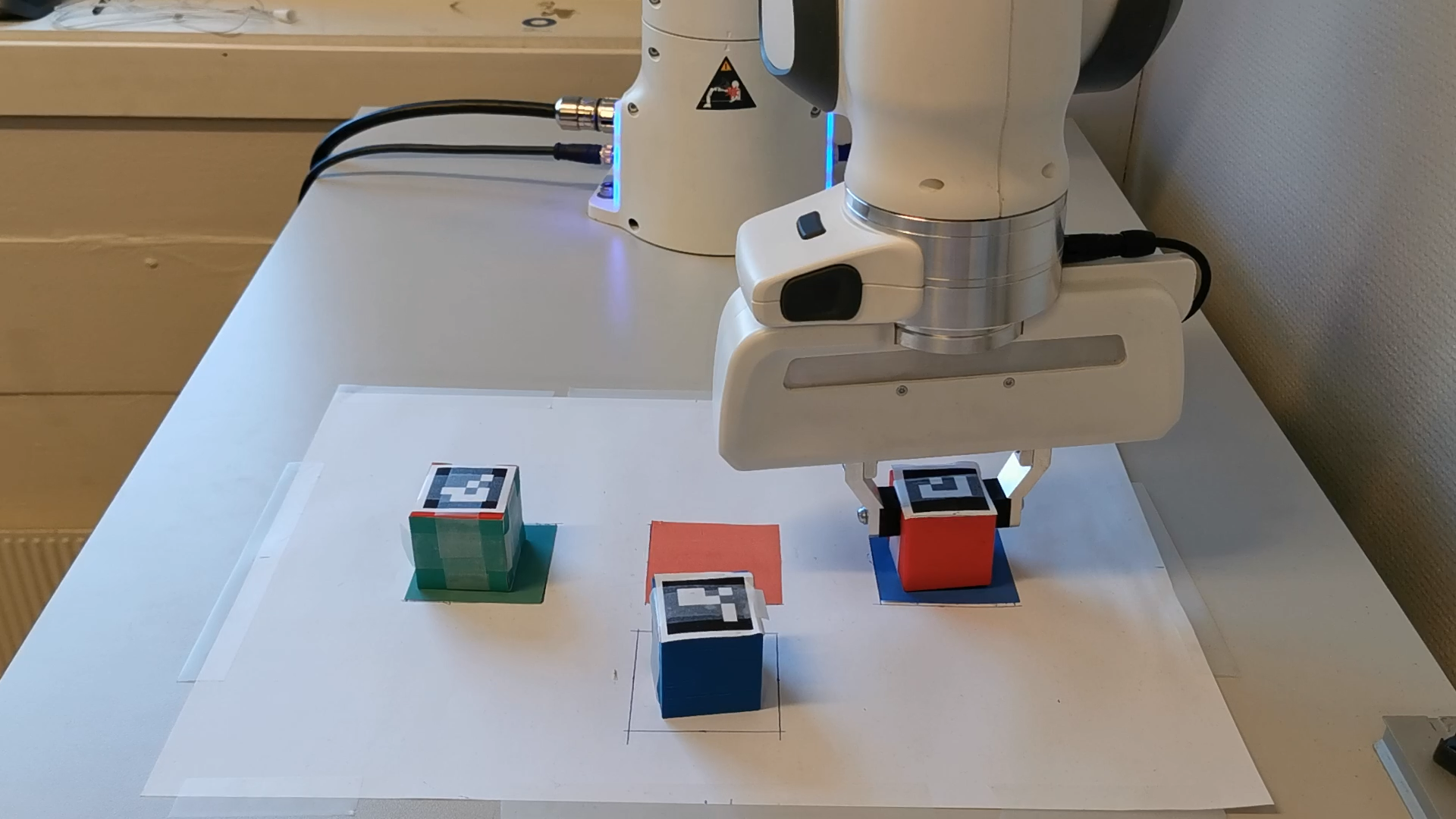} &
        \includegraphics[width=\imgwidth]{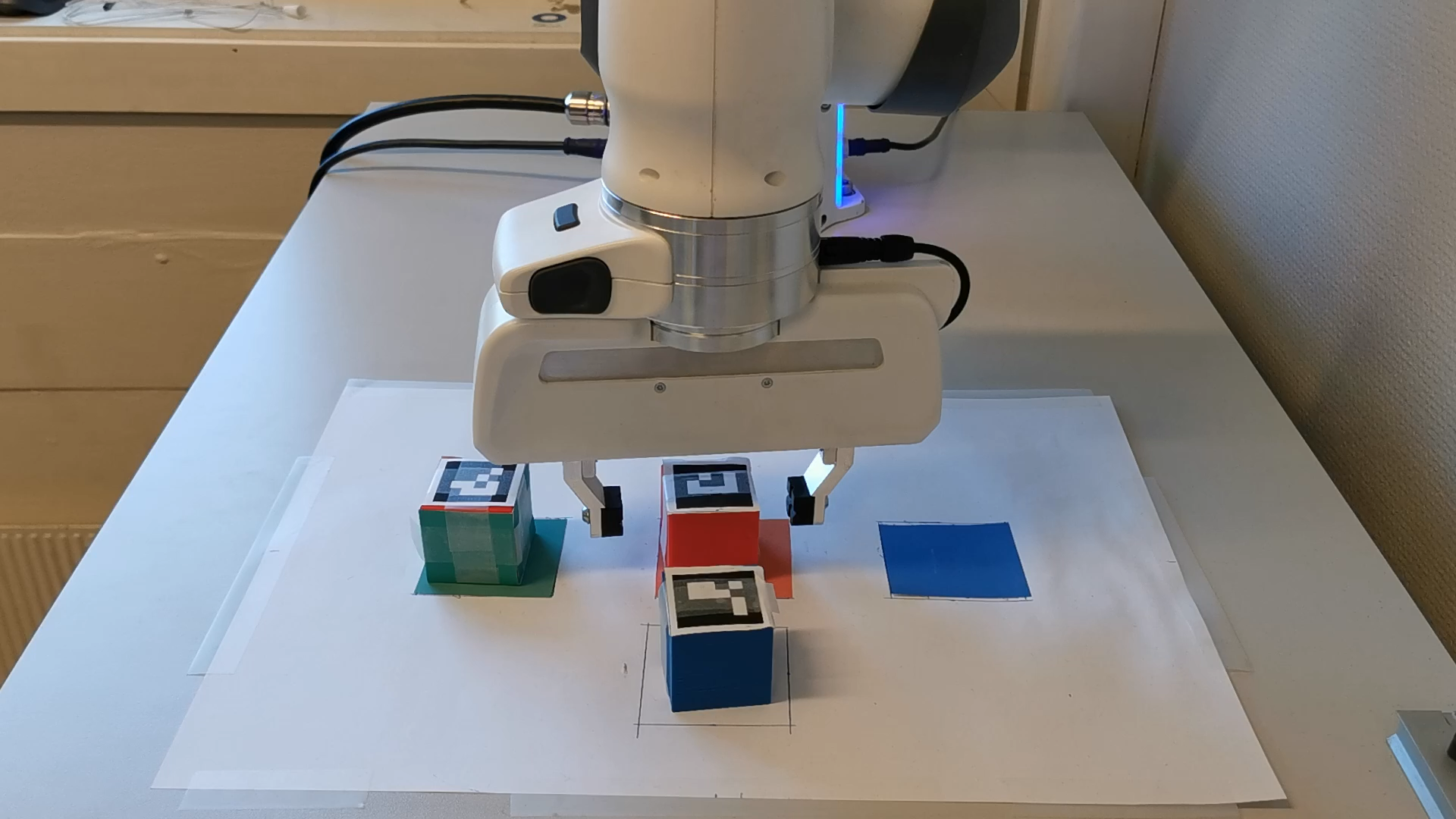} &
        \includegraphics[width=\imgwidth]{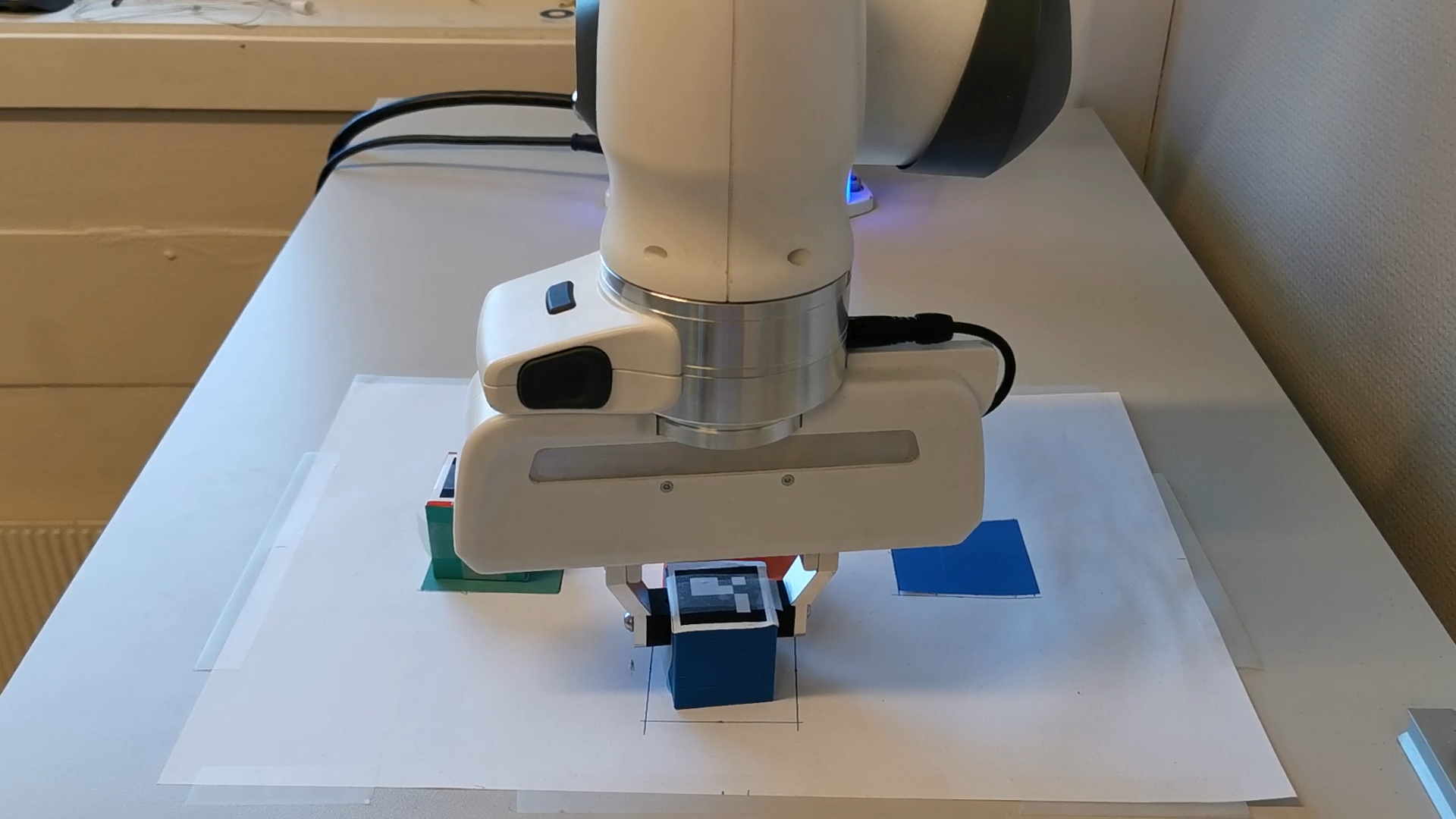} &
        \includegraphics[width=\imgwidth]{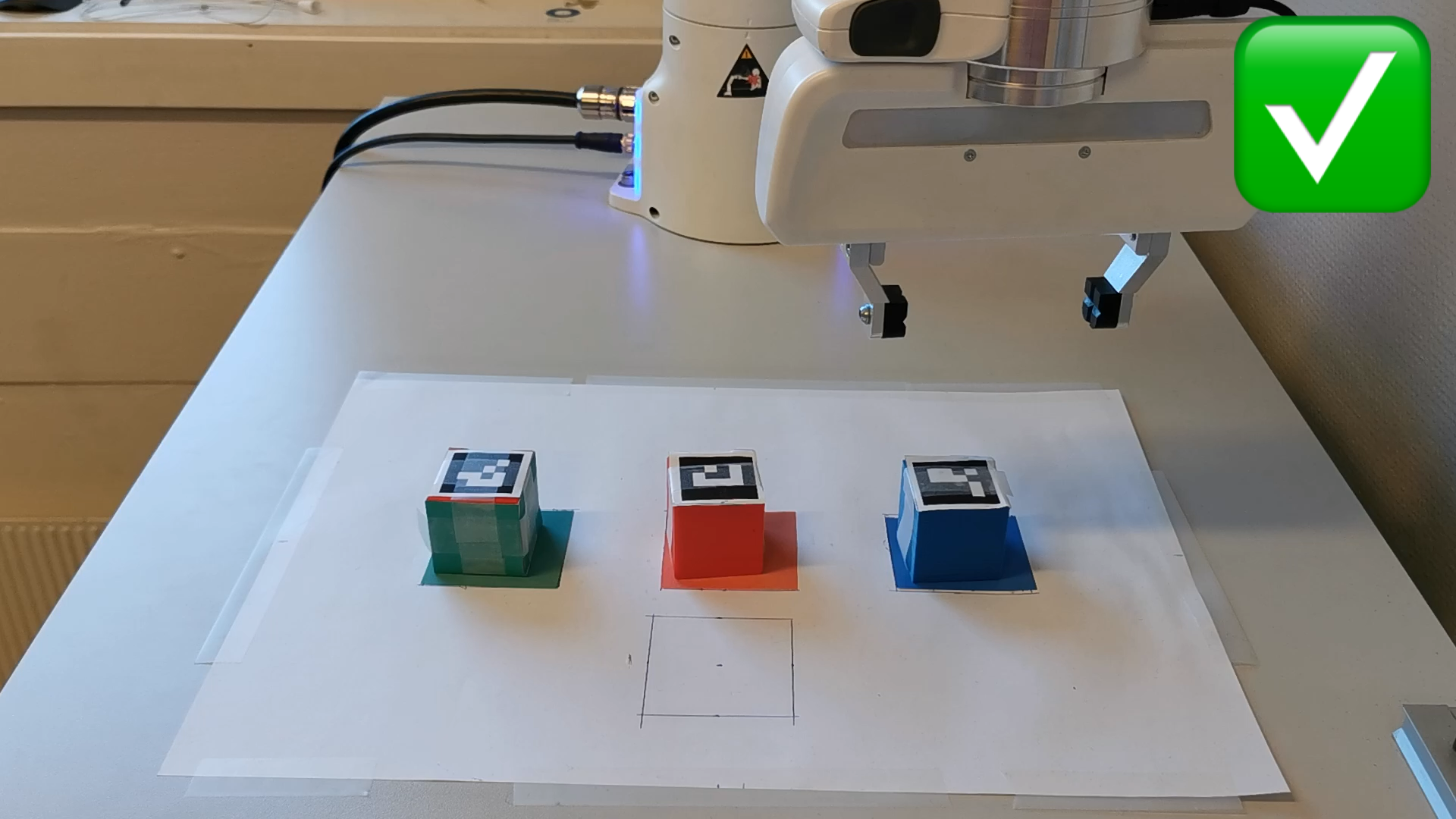}
    \end{tabular}

    \vspace{0.5em}

    \textbf{Diffuser} \\
    \begin{tabular}{@{}c@{}c@{}c@{}c@{}c@{}c@{}c@{}}
        \includegraphics[width=\imgwidth]{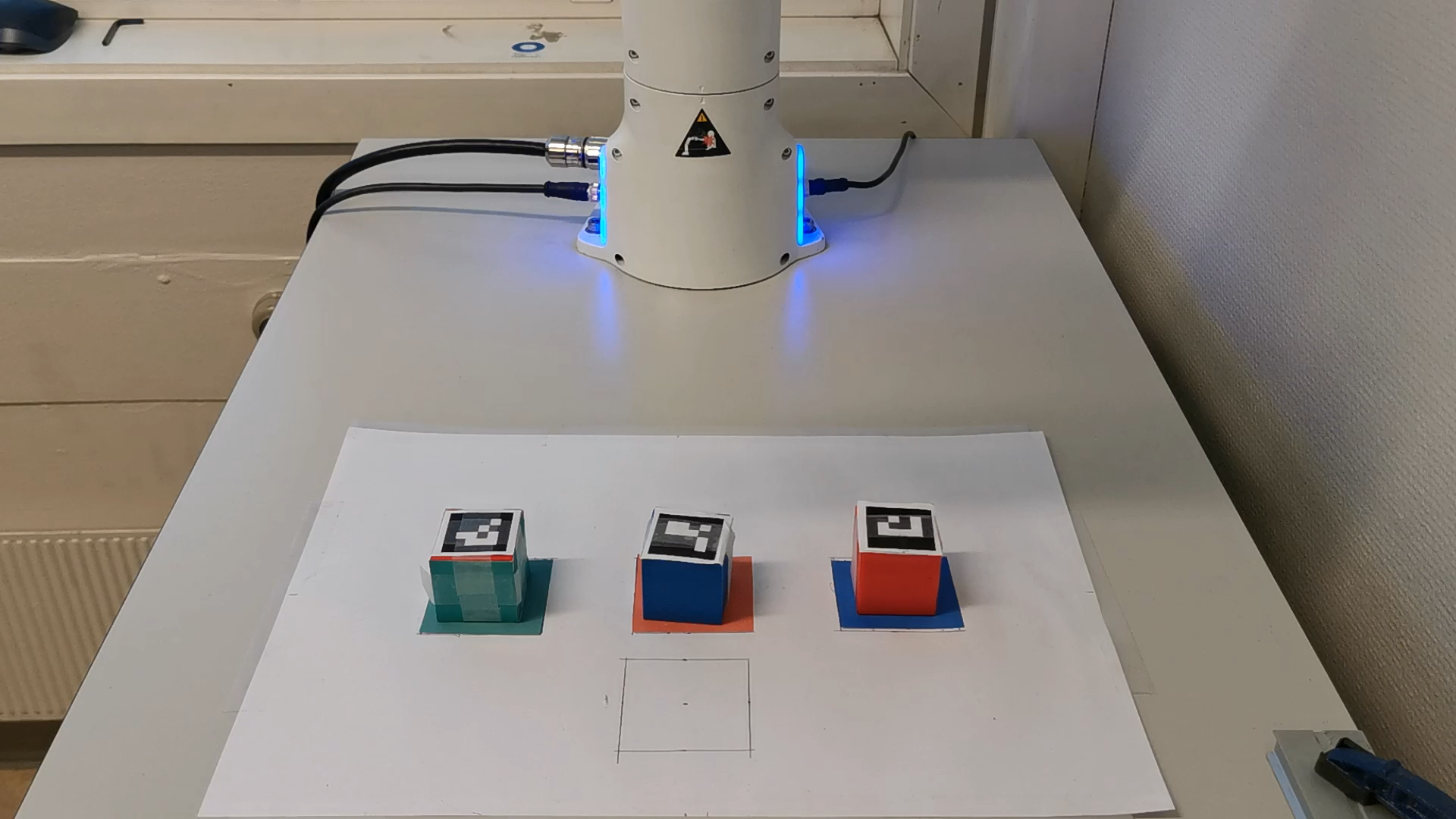} &
        \includegraphics[width=\imgwidth]{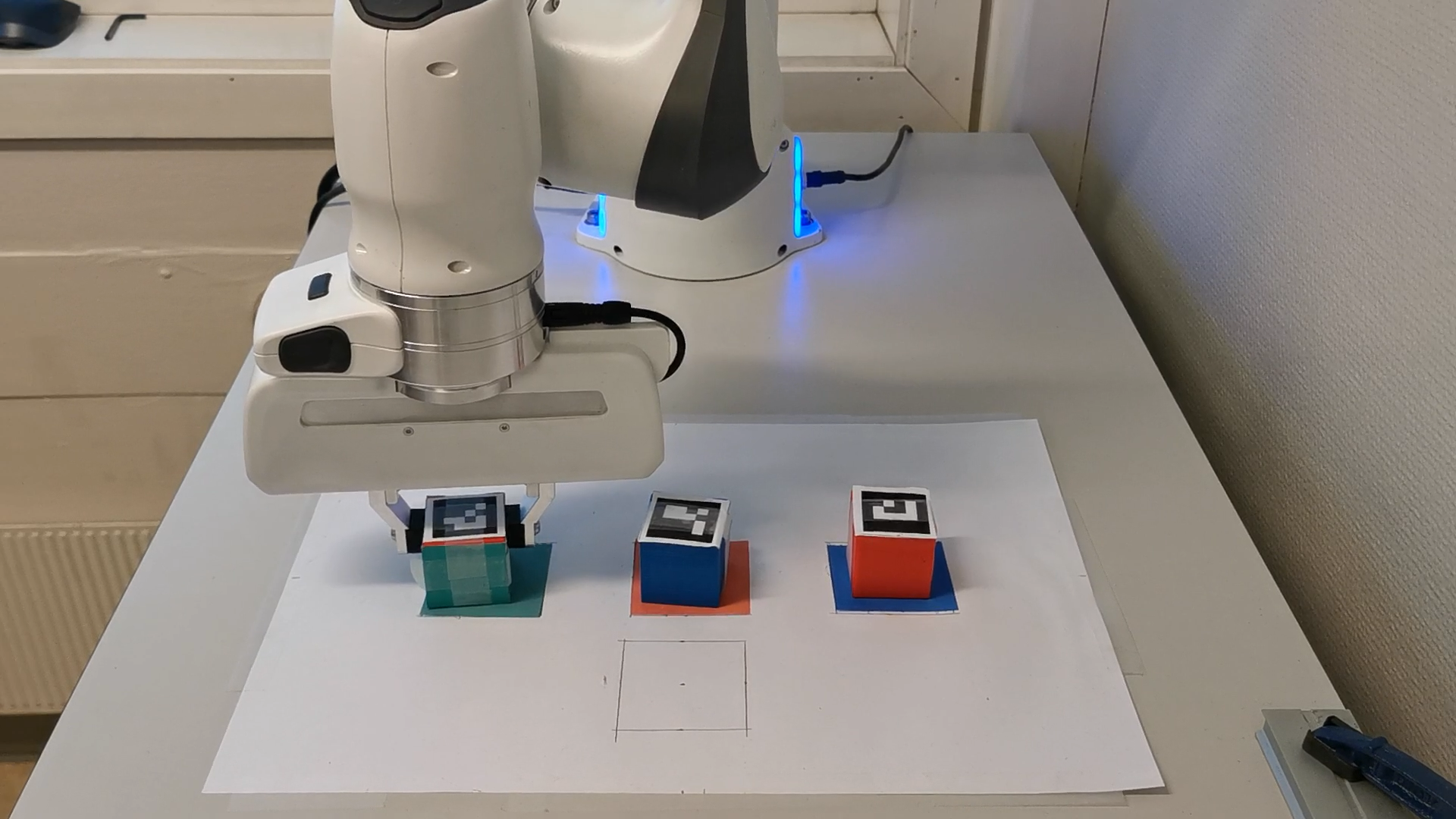} &
        \includegraphics[width=\imgwidth]{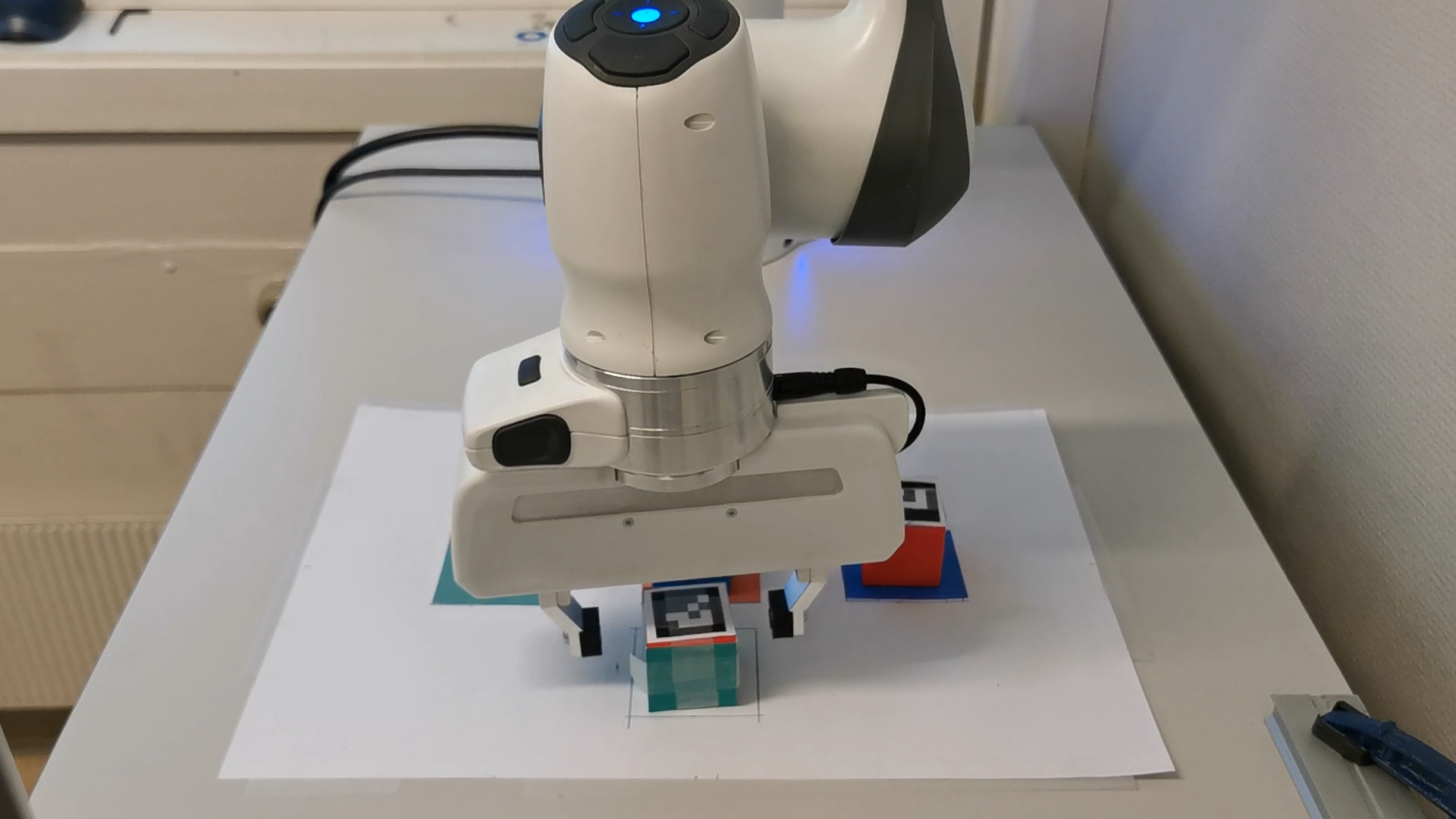} &
        \includegraphics[width=\imgwidth]{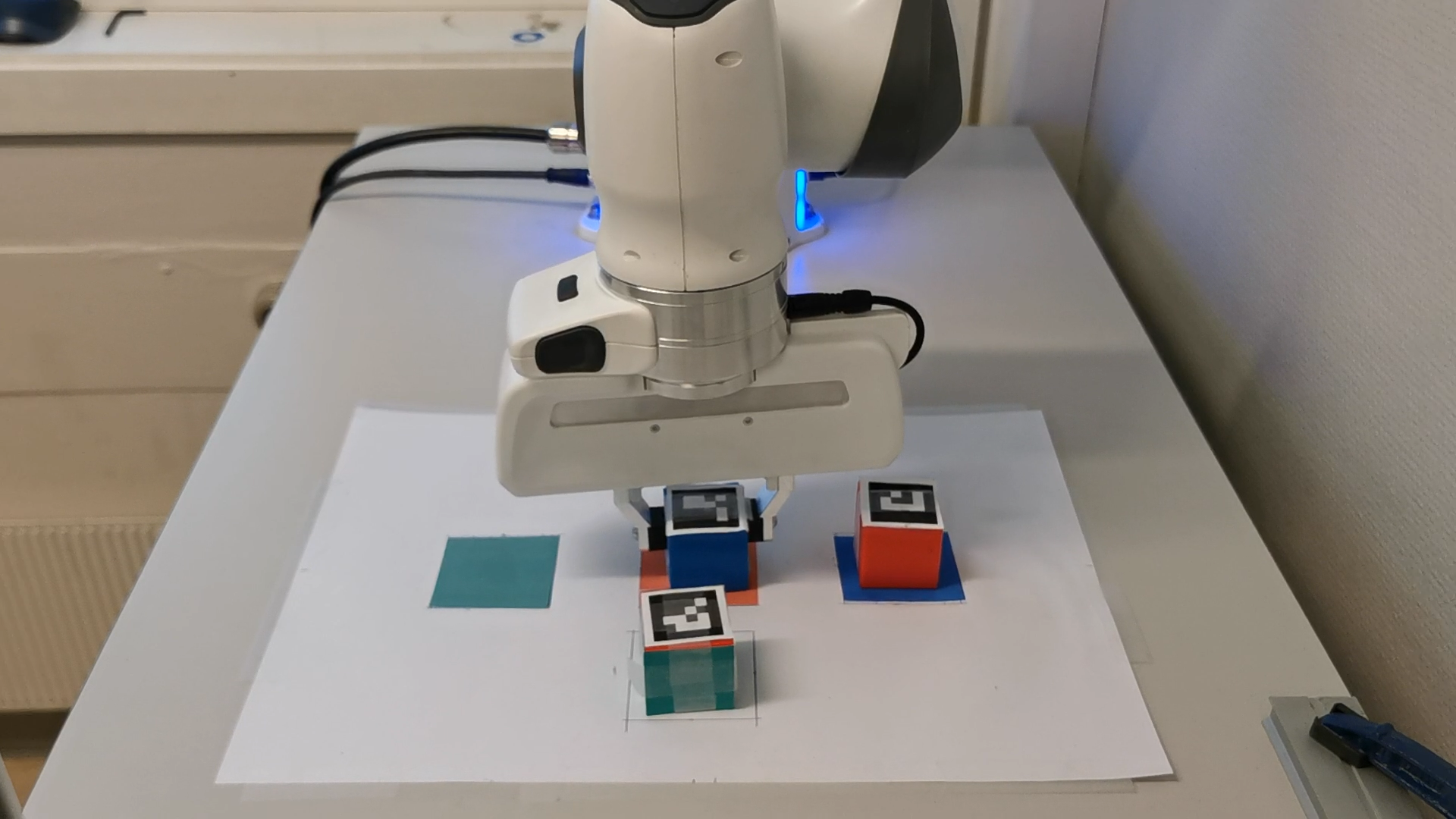} &
        \includegraphics[width=\imgwidth]{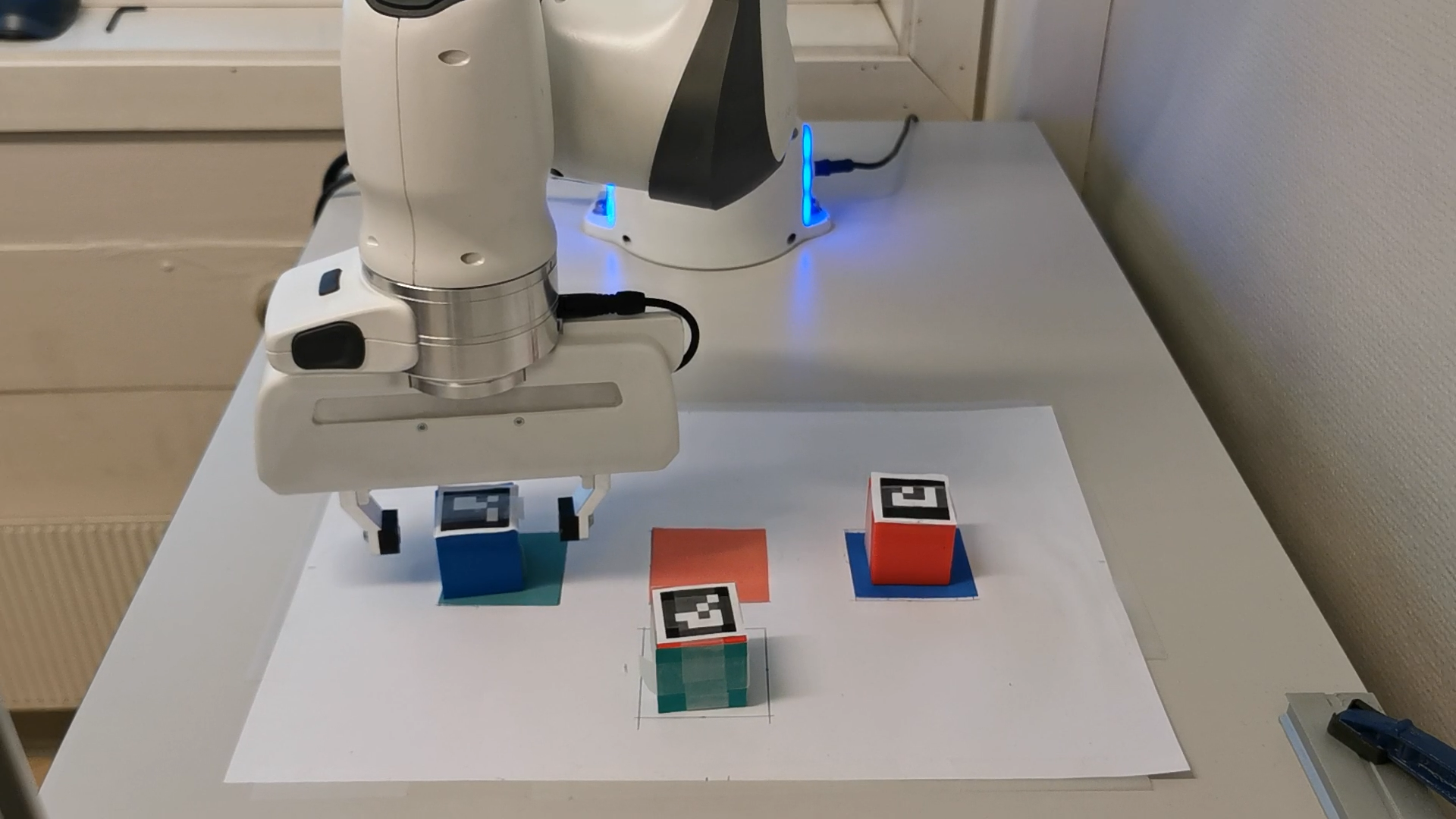} &
        \includegraphics[width=\imgwidth]{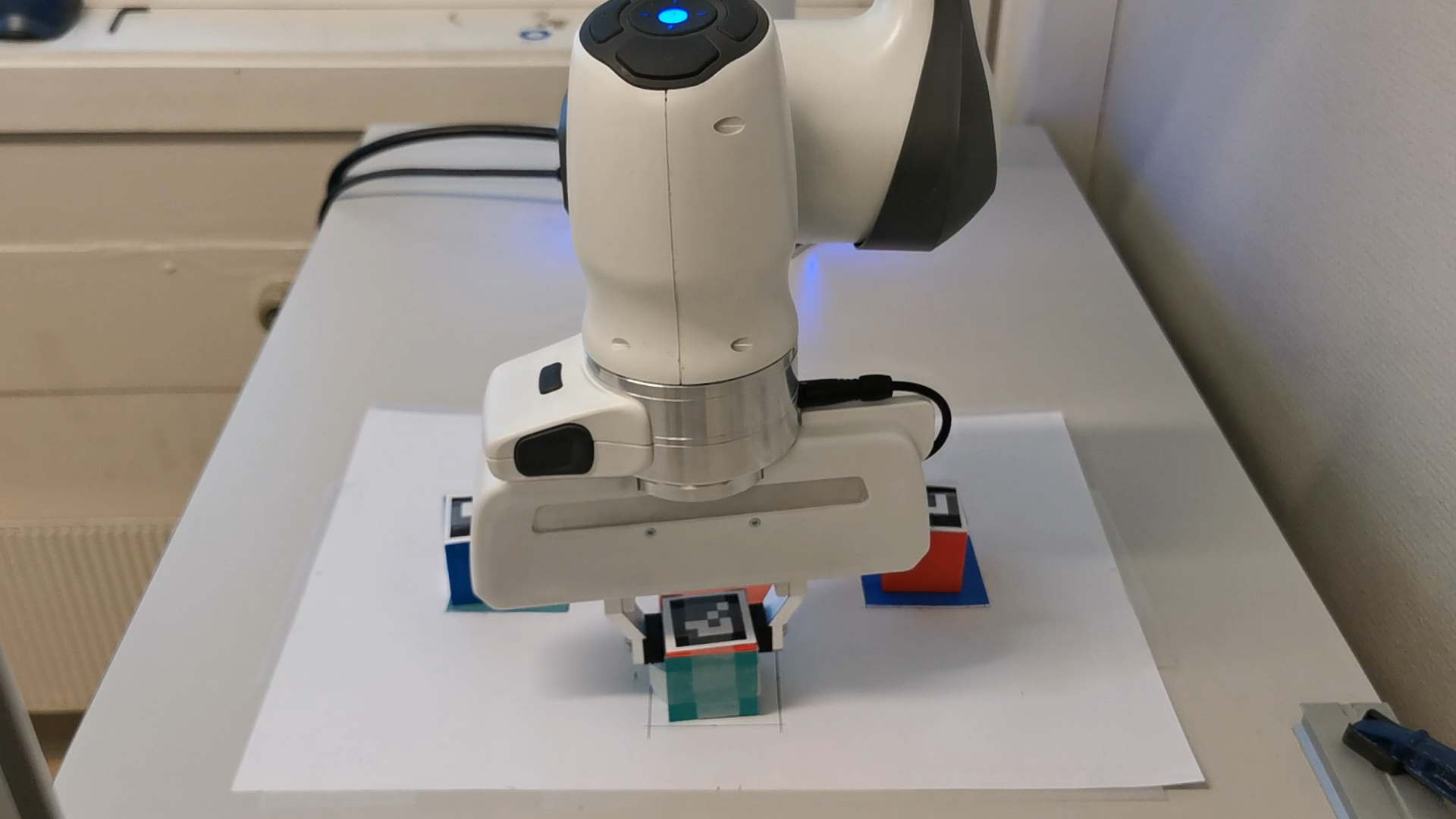} &
        \includegraphics[width=\imgwidth]{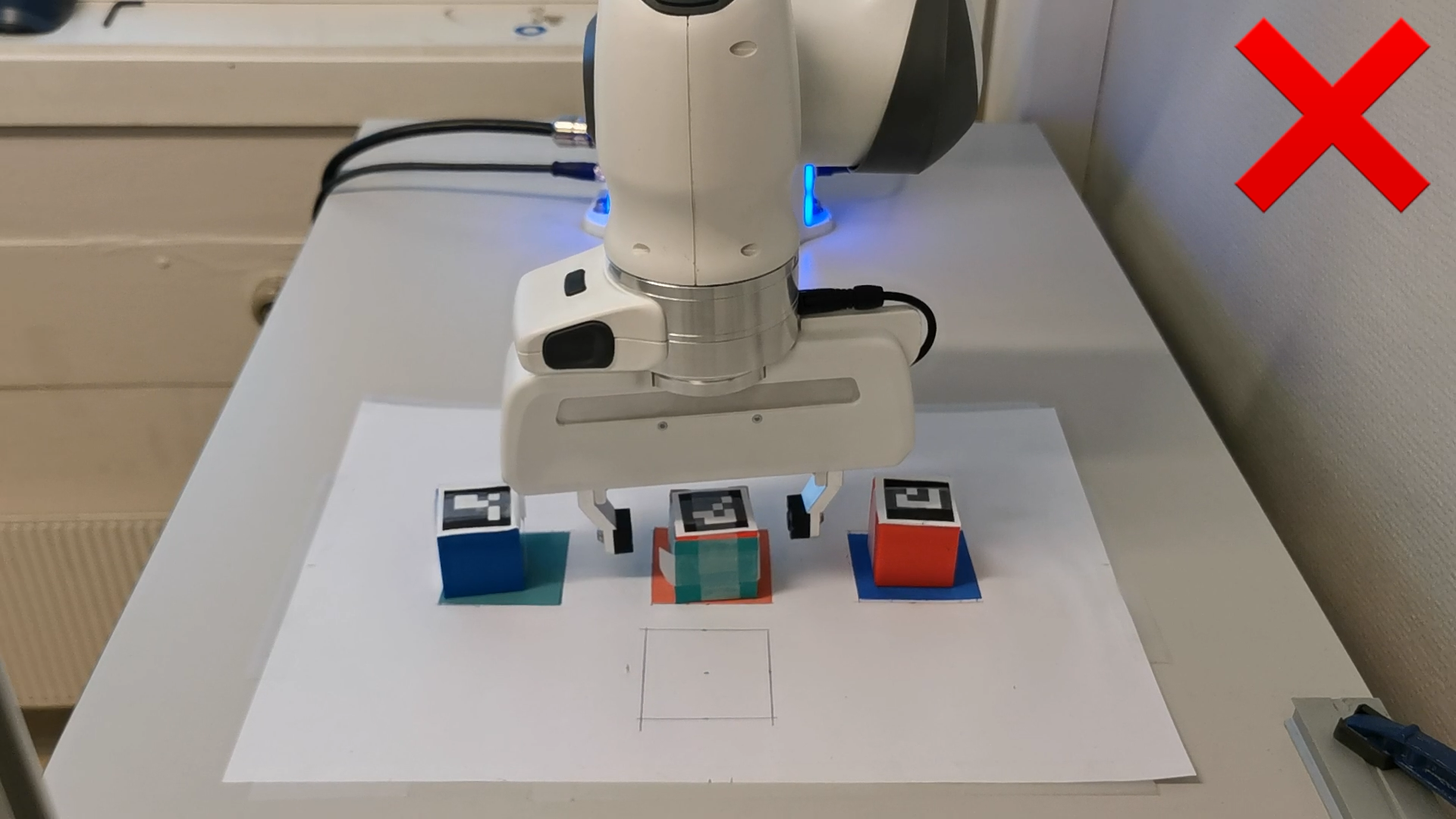}
    \end{tabular}

    \vspace{0.5em}

    \textbf{Joint Diffuser} \\
    \begin{tabular}{@{}c@{}c@{}c@{}c@{}c@{}c@{}c@{}}
        \includegraphics[width=\imgwidth]{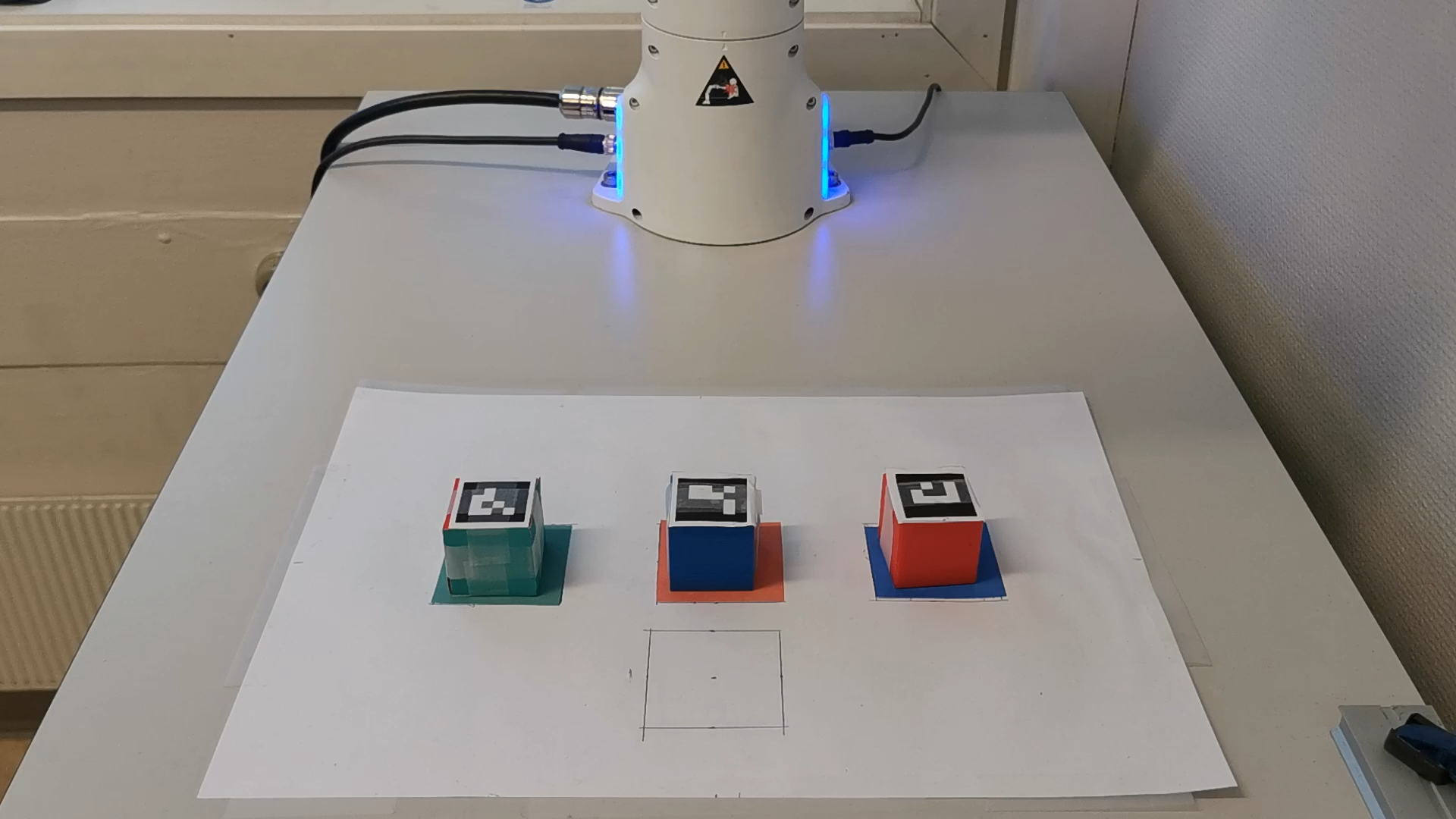} &
        \includegraphics[width=\imgwidth]{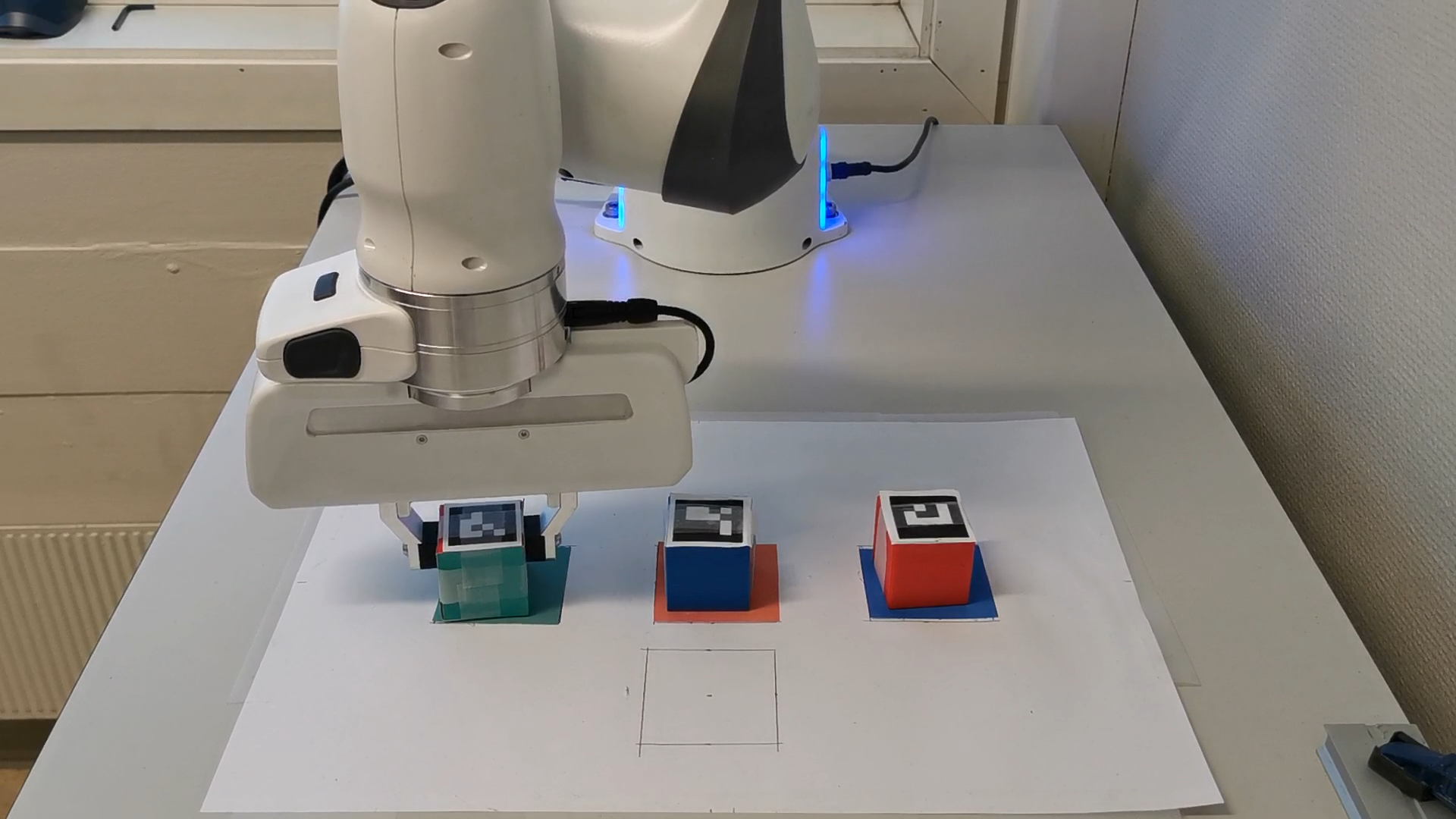} &
        \includegraphics[width=\imgwidth]{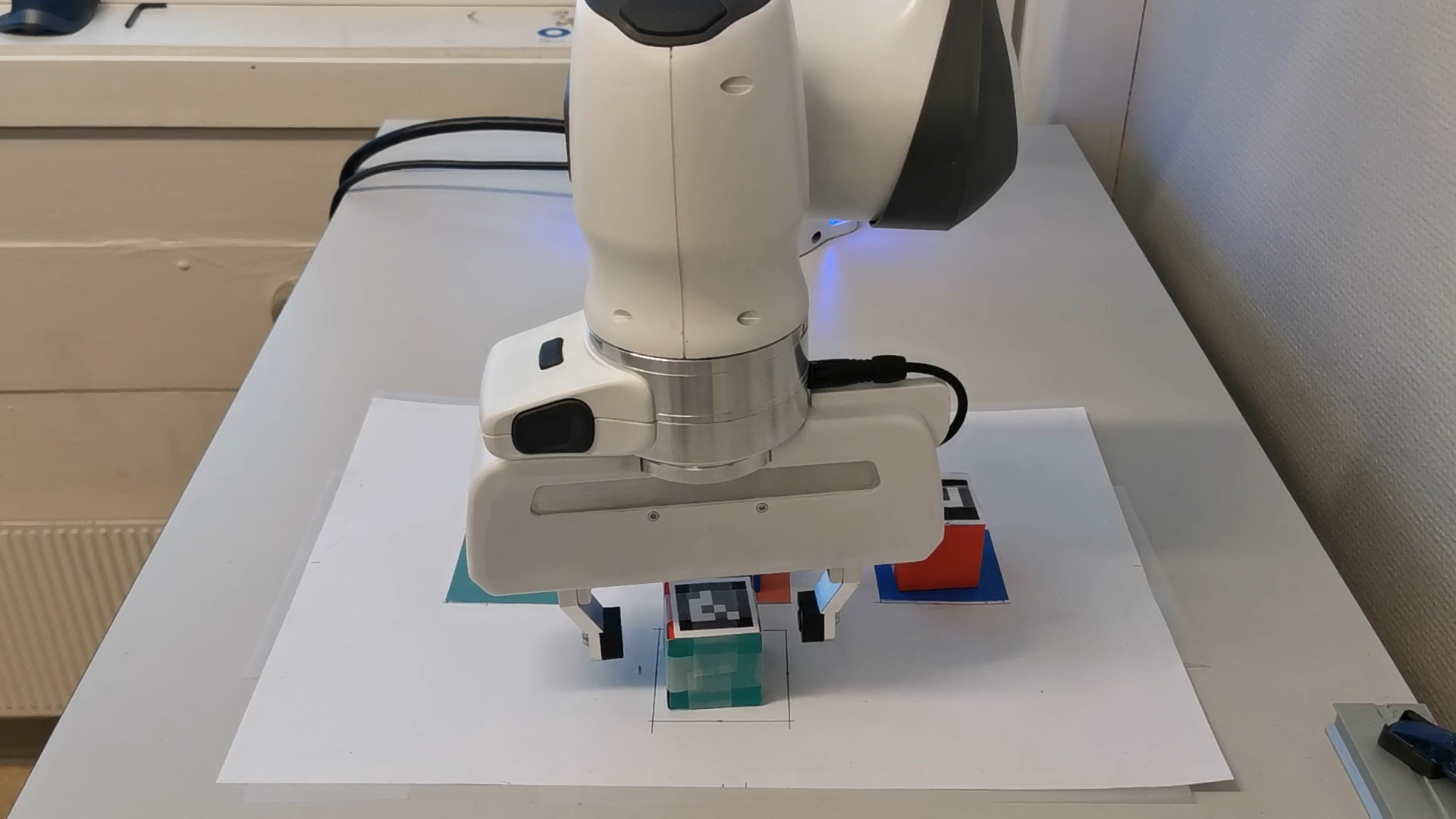} &
        \includegraphics[width=\imgwidth]{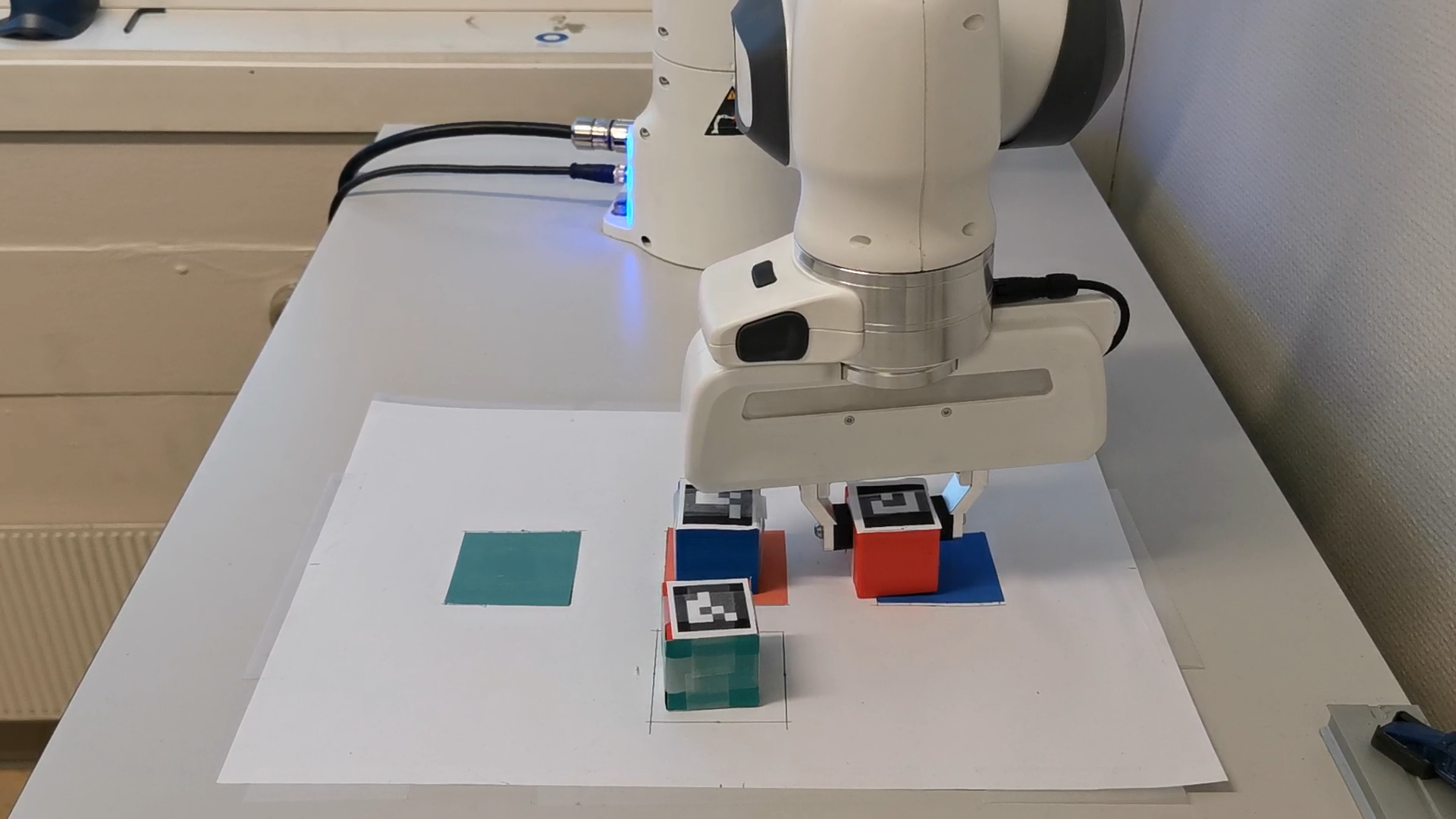} &
        \includegraphics[width=\imgwidth]{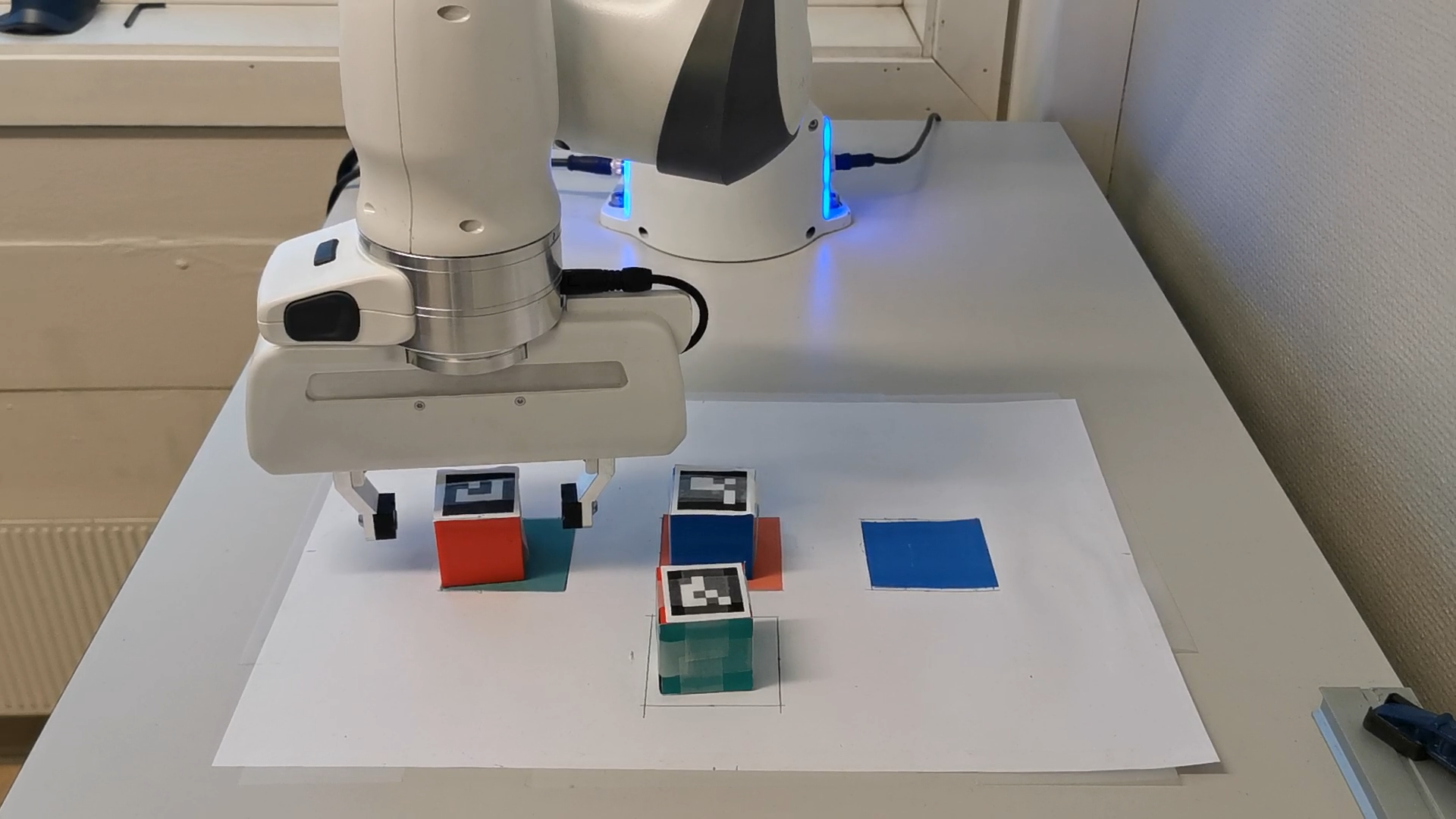} &
        \includegraphics[width=\imgwidth]{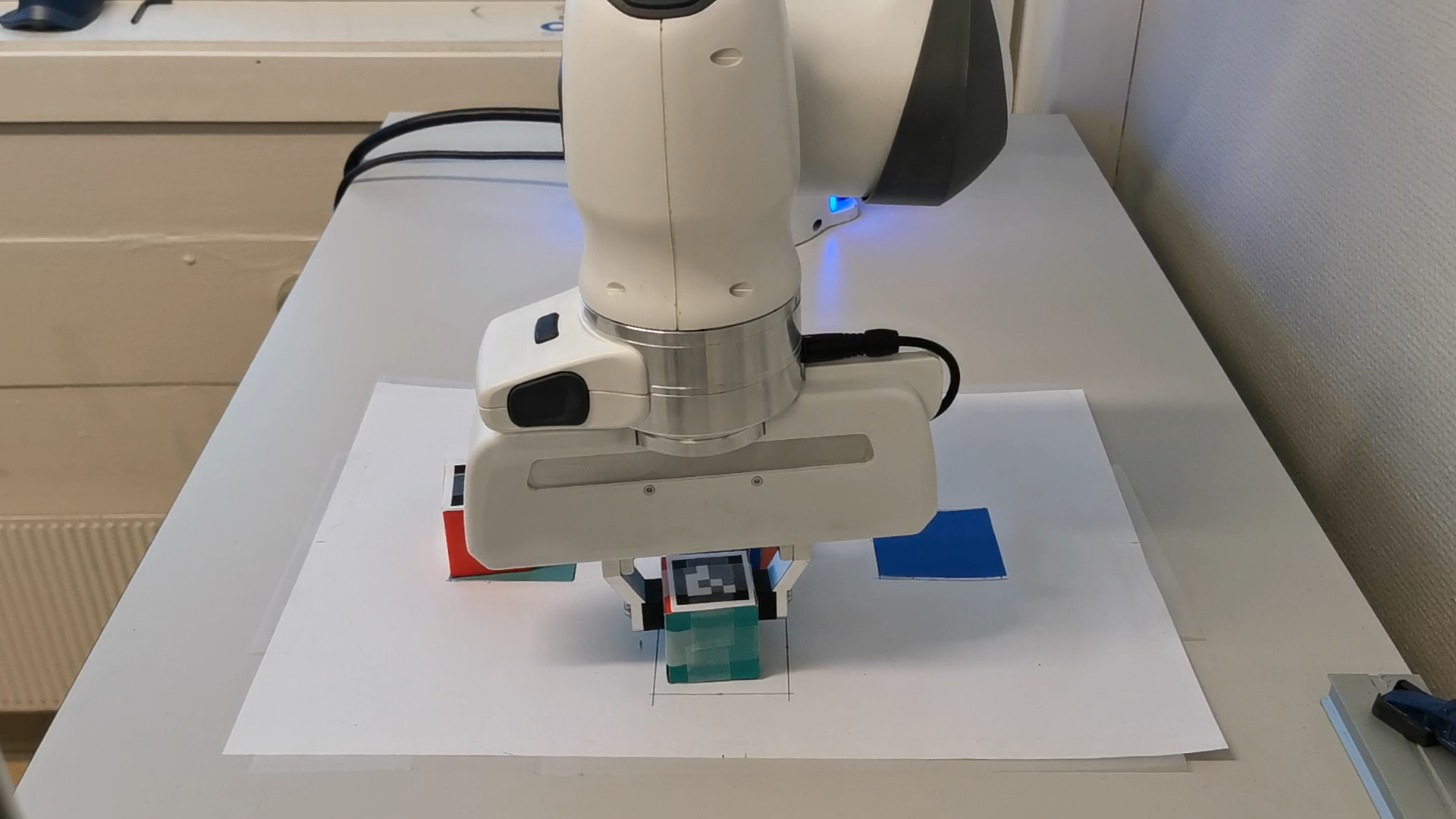} &
        \includegraphics[width=\imgwidth]{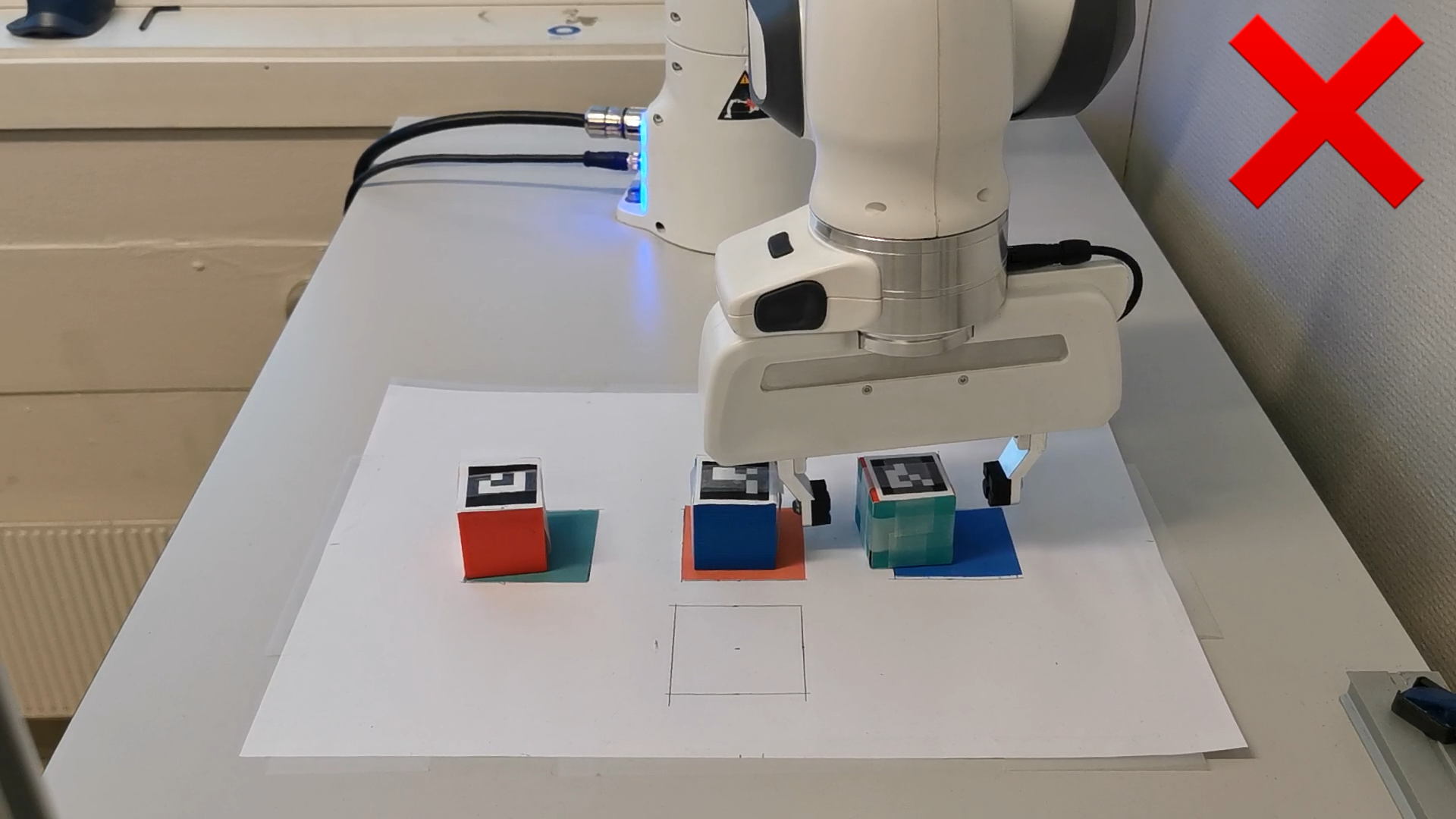}
    \end{tabular}

    \vspace{0.5em}

    \textbf{Separate Diffuser} \\
    \begin{tabular}{@{}c@{}c@{}c@{}c@{}c@{}c@{}c@{}}
        \hspace{-22em}
        \includegraphics[width=\imgwidth]{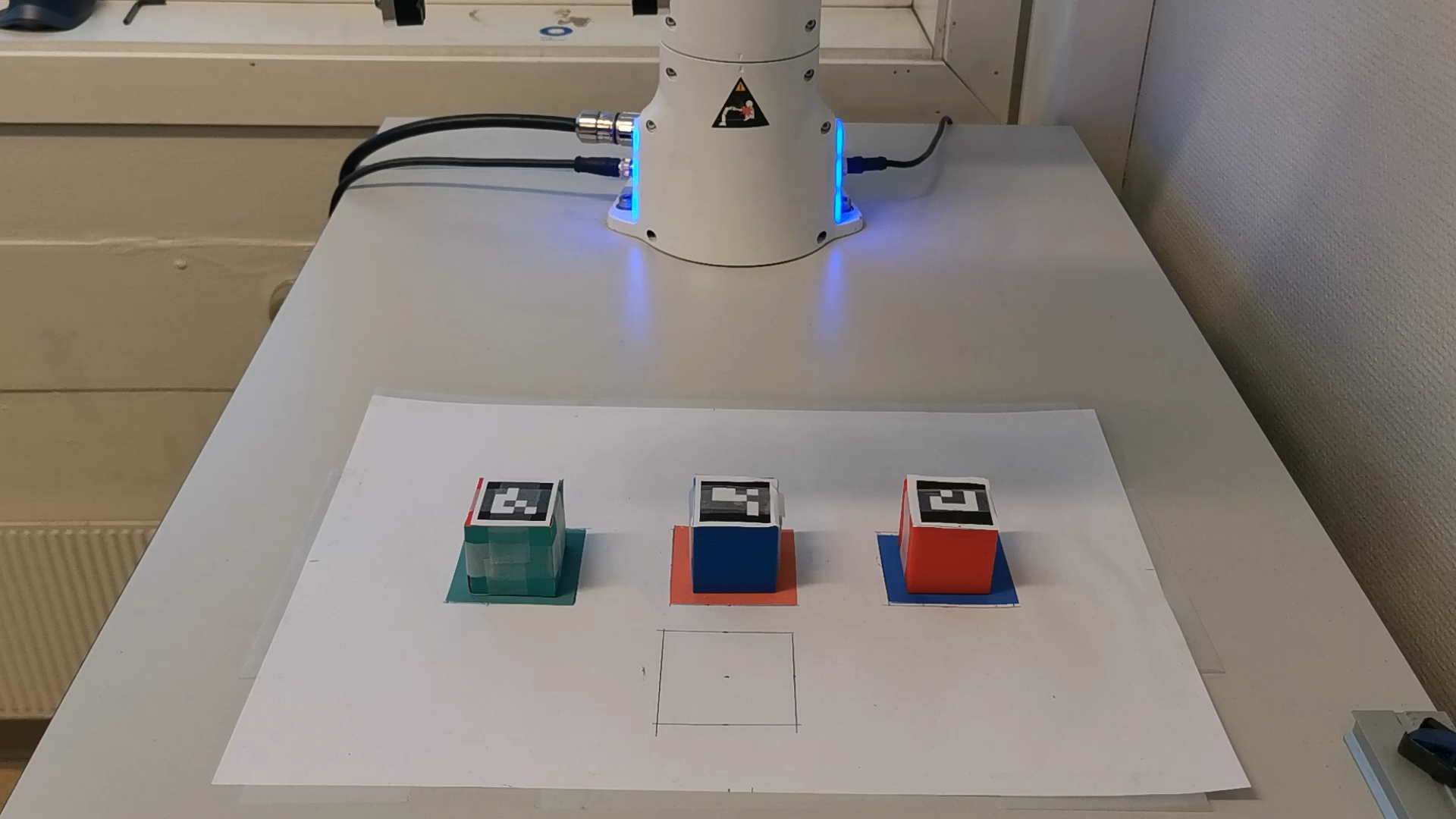} &
        \includegraphics[width=\imgwidth]{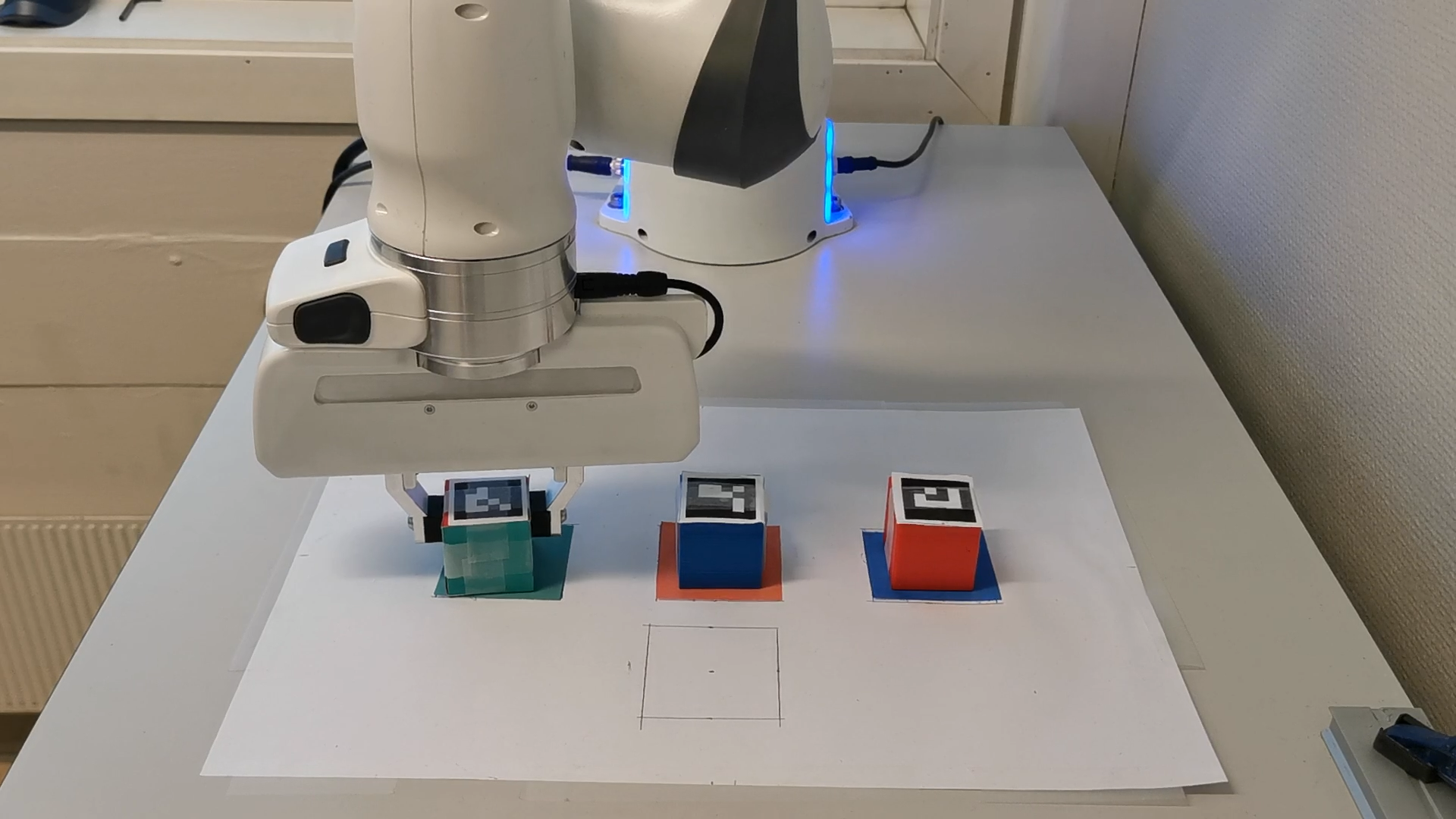} &
        \includegraphics[width=\imgwidth]{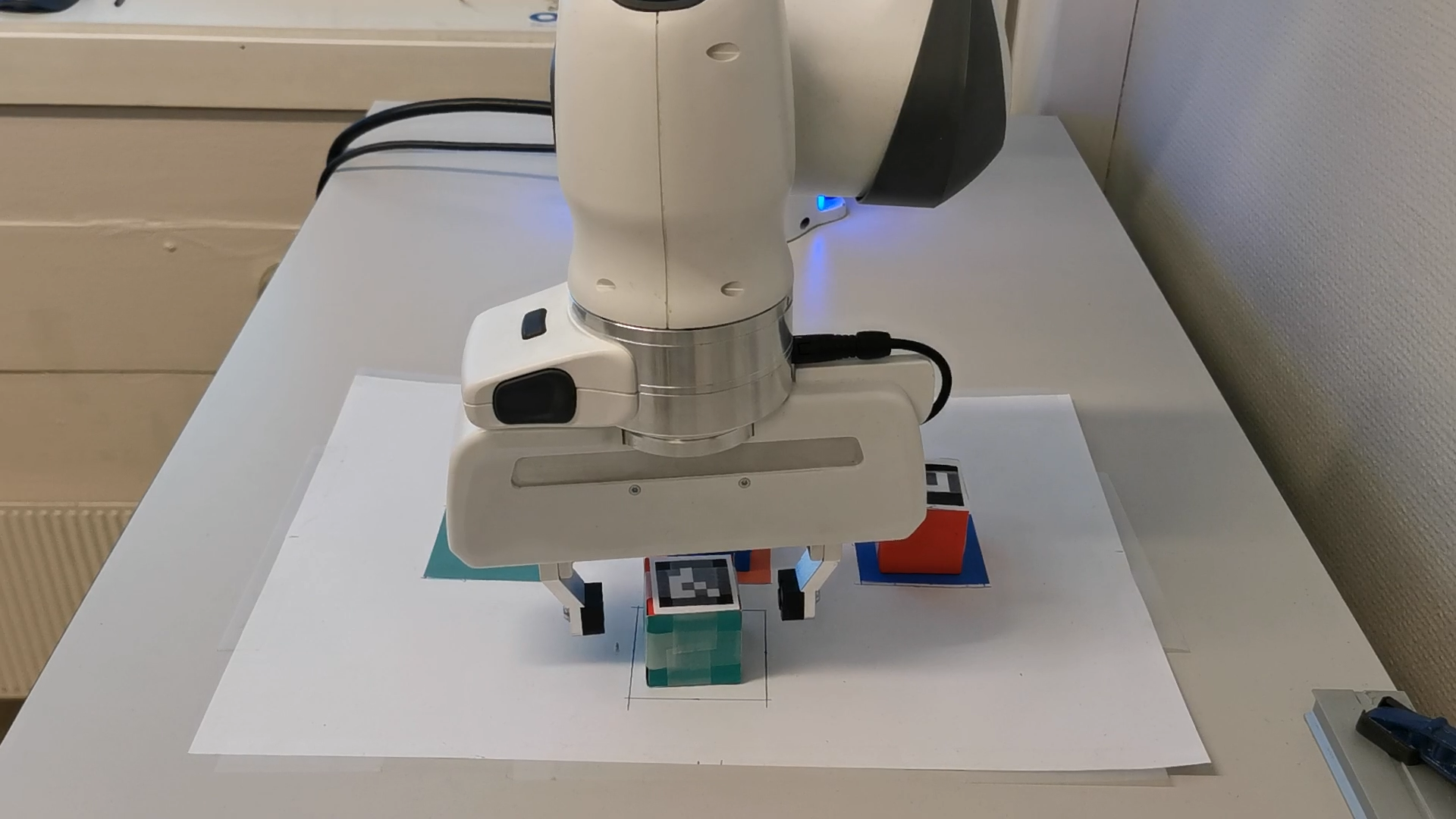} &
        \includegraphics[width=\imgwidth]{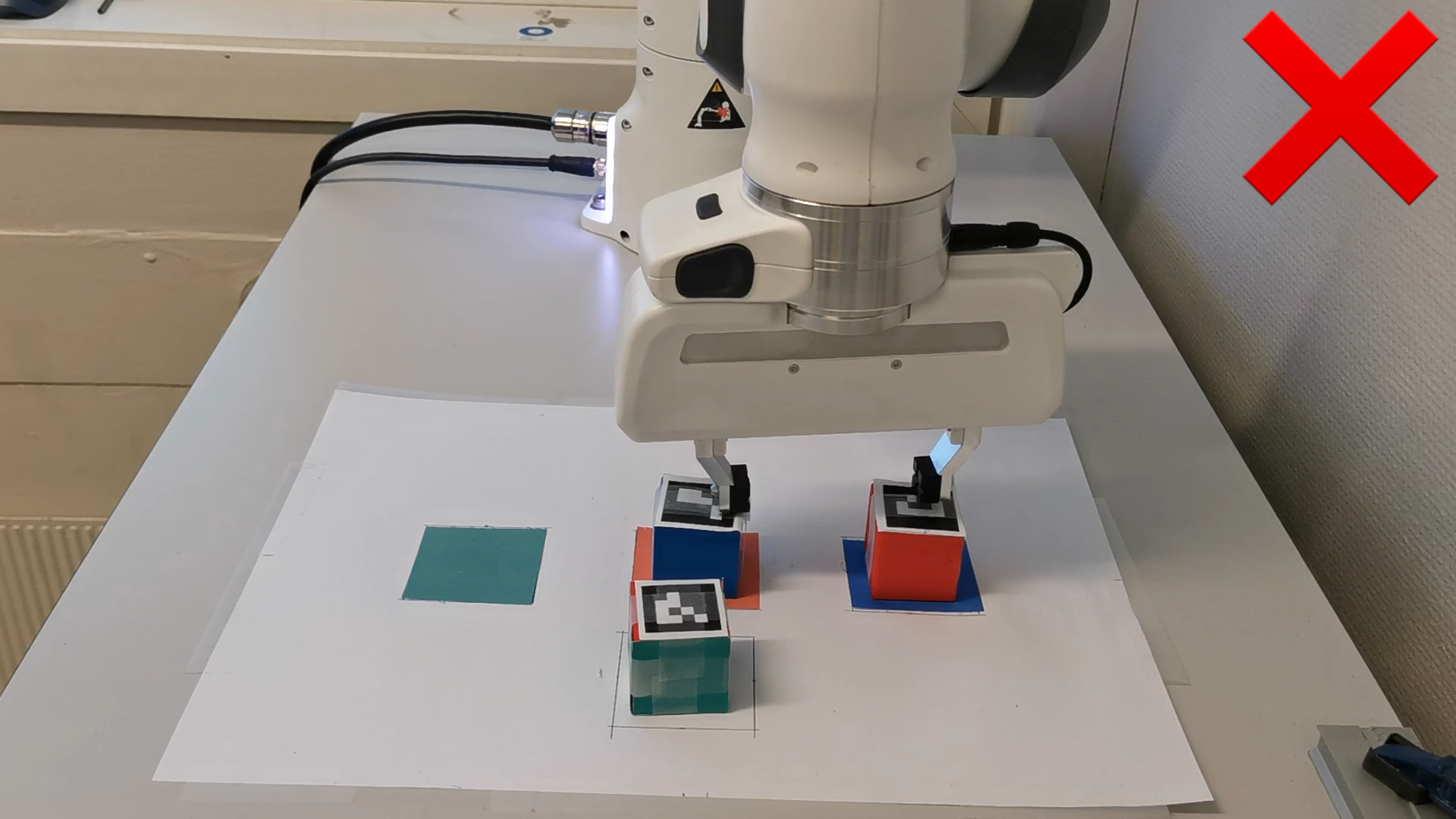} 
    \end{tabular}

    \caption{\textbf{Real-world sorting rollouts} for all methods.}
    \label{fig:real_rollouts}
\end{figure*}

\end{appendices}

\end{document}